%% file: main.tex
\documentclass[runningheads]{llncs}
\usepackage{graphicx}
\usepackage{tikz}
\usepackage{comment}
\usepackage{amsmath,amssymb}
\usepackage{color}
\usepackage{multirow}
\usepackage{graphicx}
\usepackage{booktabs}
\usepackage{multirow}
\usepackage[abs]{overpic}
\usepackage{microtype}
\usepackage{bbm}
\usepackage{algpseudocode}
\usepackage[ruled,vlined]{algorithm2e}
\usepackage[accsupp]{axessibility}
\usepackage[width=122mm,left=12mm,paperwidth=146mm,height=193mm,top=12mm,paperheight=217mm]{geometry}
\usepackage{tikz,pgfplots}

\input{macros}

\usepackage[pagebackref=true,breaklinks=true,colorlinks,bookmarks=false]{hyperref}
\definecolor{darkgreen}{RGB}{59,155,85}
\definecolor{darkred}{RGB}{190,37,3}
\hypersetup{citecolor=darkgreen}

\usepackage{xr}
\makeatletter
\newcommand*{\addFileDependency}[1]{
  \typeout{(#1)}
  \@addtofilelist{#1}
  \IfFileExists{#1}{}{\typeout{No file #1.}}
}
\makeatother

\newcommand*{\myexternaldocument}[1]{%
    \externaldocument{#1}%
    \addFileDependency{#1.tex}%
    \addFileDependency{#1.aux}%
}

\myexternaldocument{sections/7.2_standalone_supp}

\begin{document}
\pagestyle{headings}
\mainmatter

\title{Box2Mask: Weakly Supervised 3D Semantic Instance Segmentation Using Bounding Boxes
\vspace{-5mm}}

\titlerunning{3D SIS Through Bounding Box Supervision}

\author{
Julian Chibane\inst{1,2}
\and
Francis Engelmann\inst{3}
\and
Tuan Anh Tran\inst{2}\index{Tran, Tuan Anh}
\and\\
Gerard Pons-Moll\inst{1,2}
}
\authorrunning{J. Chibane et al.}
\institute{University of Tübingen, Germany\\
\and
Max Planck Institute for Informatics, Saarland Informatics Campus, Germany\\
\and
ETH Zürich, AI Center, Switzerland\\
}
\maketitle

\input{notation}

\input{sections/0_abstract}
\input{sections/1_introduction}
\input{sections/2_related_work}

\input{sections/3_method}
\input{sections/4_experiments}
\input{sections/5_conclusion}
\input{sections/6_acknowledgements}

\bibliographystyle{splncs04}
\bibliography{abbrev, egbib}

\newpage
\clearpage
\appendix
\input{sections/7_supplementary}

\end{document}

%% file: macros.tex
\usepackage{booktabs}
\usepackage{balance}
\usepackage{multirow}
\usepackage{cite}
\usepackage[font=small,labelfont=bf]{caption}
\usepackage{color}
\usepackage{xcolor}
\usepackage{pifont}
\usepackage{float}
\usepackage{tikz}
\usepackage{wrapfig}

\usepackage{colortbl}
\usepackage{subfigure}

\newcommand{\refsec}[1]{Sec.\,\ref{sec:#1}}

\newcommand{\reffig}[1]{Fig.\,\ref{fig:#1}}
\newcommand{\reftab}[1]{Tab.\,\ref{tab:#1}}

\newcommand{\eg}{\emph{e.g.}}
\newcommand{\ie}{\emph{i.e.}}
\newcommand{\etal}{\emph{et al.}}
\newcommand{\vs}{\emph{v.s.}}

\definecolor{m_green}{RGB}{212, 251, 121}
\definecolor{m_orange}{RGB}{255, 212, 121}
\definecolor{m_red}{RGB}{255, 126, 121}
\definecolor{m_violet}{RGB}{215, 131, 255}
\definecolor{m_blue}{RGB}{118, 214, 255}

\newcommand{\parag}[1]{\vskip8pt \noindent \textbf{#1}}

\newcommand{\name}{Box2Mask}

\definecolor{darkgreen}{RGB}{0,255,0}
\definecolor{linkgreen}{RGB}{52,130,48}

\newcommand{\ColorMapCircle}{\ding{108}}

\newif\ifmynotes

\mynotestrue 

\definecolor{notetext}{rgb}{0.7,0,0}

%% file: notation.tex
\newcommand{\point}[0]{\mathbf{p}}
\newcommand{\points}[0]{\mathcal{P}}

\newcommand{\bbox}[0]{\mathbf{b}}
\newcommand{\boxes}[0]{\mathcal{B}}

\newcommand{\fg}[0]{\mathcal{F}}
\newcommand{\decided}[0]{\mathcal{D}}
\newcommand{\asso}[0]{a}

\newcommand{\bboffsets}[0]{o}
\newcommand{\bbsize}[0]{s}
\newcommand{\iou}[0]{iou}
\newcommand{\sem}[0]{sem}

\makeatletter
\newcommand*{\defeq}{\mathrel{\rlap{%
                     \raisebox{0.3ex}{$\m@th\cdot$}}%
                     \raisebox{-0.3ex}{$\m@th\cdot$}}%
                     =}
\makeatother

%% file: sections/0_abstract.tex
\vspace{-13px}
\input{figures_tex/teaser}
\begin{abstract}
    Current 3D segmentation methods heavily rely on large-scale point-cloud datasets, which are notoriously laborious to annotate.
    Few attempts have been made to circumvent the need for dense per-point annotations.
    In this work, we look at weakly-supervised 3D semantic instance segmentation.
    The key idea is to leverage 3D bounding box labels which are easier and faster to annotate.
    Indeed, we show that it is possible to train dense segmentation models using only  bounding box labels.
    At the core of our method, \name{}, lies a deep model, inspired by classical Hough voting, that directly votes for bounding box parameters, and a clustering method specifically tailored to bounding box votes.
    This goes beyond commonly used center votes, which would not fully exploit the bounding box annotations.
    On ScanNet test, our weakly supervised model attains leading performance among other weakly supervised approaches (+18\,mAP$_{50}$).
    Remarkably, it also achieves $97\%$ of the mAP$_{50}$ score of current fully supervised models.
    To further illustrate the practicality of our work,
    we train \name{} on the recently released ARKitScenes dataset which is annotated with 3D bounding boxes only, and show, for the first time, compelling 3D instance segmentation masks.
\keywords{\small
3D Semantic Instance Segmentation,
Weakly Supervised Learning,
3D Scene Understanding
}
\end{abstract}

%% file: figures_tex/teaser.tex
\definecolor{forestgreen}{rgb}{0.13, 0.55, 0.13}

\begin{center}
\vspace{2mm}
\url{virtualhumans.mpi-inf.mpg.de/box2mask}
\end{center}
\begin{figure}[h]
    \centering
    \vspace{-4mm}
    \includegraphics[width=\textwidth, trim=0 0px 0 0, clip]{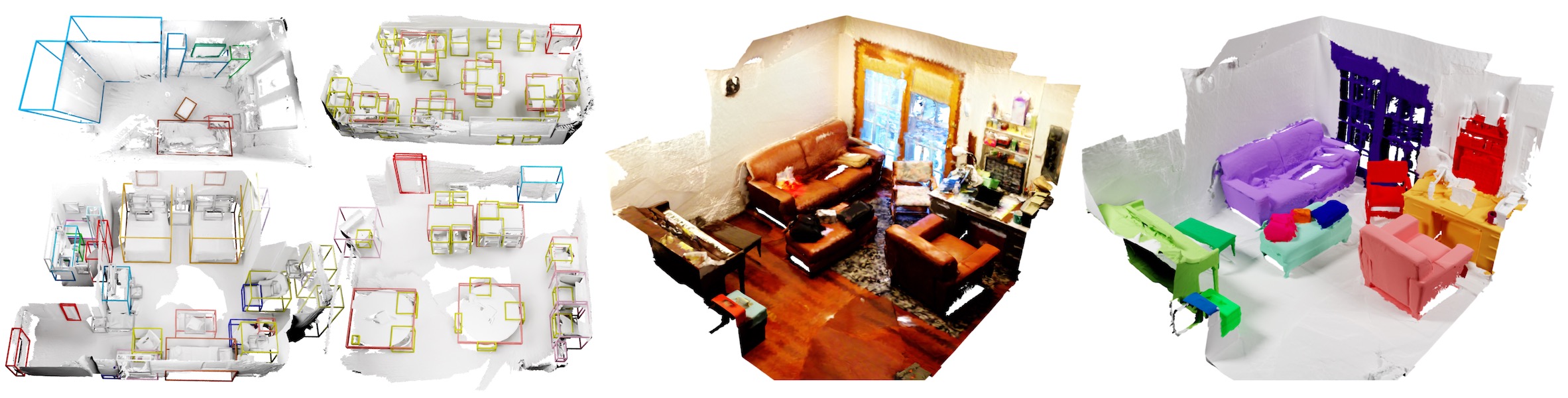}
    \vspace{-1.5mm}
    \begin{footnotesize}
    \centering
    \begin{tabular}{ccc}
         Training Data      & Input & Prediction \\
          3D Bounding Box Annotations   & 3D Test Scene &  Instance Segmentation\\
               \textcolor{red}{\textbf{sparse, object-level }}                        &              &  \textcolor{forestgreen}{\textbf{dense, point-level}} \\
         \hspace{0.34\textwidth} & \hspace{0.355\textwidth} & \hspace{0.2\textwidth}
     \end{tabular}
    \end{footnotesize}
    \vspace{-9px}
    \caption{\small
    Can we use 3D bounding box annotations alone to train dense 3D semantic instance segmentation models? We find that this is the case and propose a Deep Hough Voting based method that fully exploits bounding box annotations.}
    \label{fig:teaser}
    \vspace{-9mm}
\end{figure}

%% file: sections/1_introduction.tex
\input{figures/annotationtypes/annotationtypes}

\section{Introduction}
Semantic instance segmentation of 3D scenes is one of the fundamental challenges in computer vision and robotics.
The goal is to predict a foreground-background mask and a semantic class (\eg,~`chair', `fireplace') for each object in a 3D scene (point cloud or mesh).
Over the last years, the research community has contributed numerous methods 
\cite{Yang19NIPS, Jiang20CVPR, He21CVPR, Han20CVPR, Liang21CVPR, Engelmann20CVPR, Chen21ICCV, Lahoud19ICCV, Elich19GCPR}.
This rapid development was made possible, not only by substantial advances in 3D deep learning backbones \cite{Qi17CVPR, Qi17NIPS, Schult20CVPR, Thomas19ICCV, Choy19CVPR, Graham18CVPR}, but also by large-scale 3D datasets \cite{Armeni16CVPR, Dai17CVPR, Song17CVPR, Mo19CVPR} crucial to train data-hungry deep models.
While the acquisition of large datasets has become easier with commodity 3D scanners \cite{Baruch21NIPS}, per-point annotations (Fig. \ref{fig:annotationtypes}, left) -- largely required by current methods -- are still very labour-intensive. For example, labeling an average scene in ScanNet takes $\sim$22.3 minutes \cite{Dai17CVPR}. It is therefore highly desirable to alleviate the need for dense point labels.
Only few works have addressed this challenge.
Hou \etal \cite{Hou21CVPR} build on self-supervised pre-training~\cite{Xie20ECCV} and propose contrastive learning techniques using sparse point annotations (Fig. \ref{fig:annotationtypes}, middle) which,
however, depend on carefully selected points during the annotation process.

The key idea of this work is to use 3D bounding box annotations as weak supervision signal for dense 3D semantic instance segmentation (Fig. \ref{fig:annotationtypes}, right).
Despite promising results on image understanding tasks \cite{khoreva17CPVR},
bounding boxes have so far been overlooked for dense 3D instance segmentation. 
We find that obtaining dense segmentation masks from object detection models (predicting one box per object) is non-trivial and leads to unsatisfying results (see Sec.\,\ref{sec:analysis}).
We present \name{}, the first method for dense instance segmentation trained solely on coarse bounding box annotations. 
The main result of this paper is that our weakly supervised method
 outperforms previous weakly supervised works~\cite{Hu21CVPR,Xie20ECCV} by a large margin, and 
is even competitive with fully-supervised state-of-the-art methods \cite{Liang21CVPR, Chen21ICCV}.
To achieve this goal, we face two key challenges. First, we lack dense per-point instance annotation.
Second, there is no obvious representation for instance segmentation, because in contrast to semantic segmentation we cannot assign a single categorical label to points.

To address these challenges, we represent instances with the six parameters of an axis-aligned bounding box, fully exploiting the given labels.
Since bounding boxes cover the full extent of the object, they are naturally a richer instance representation than the commonly used centers.
Moreover, leveraging this representation leads to novel algorithms for voting, instance clustering, and a new training strategy to cope with weak labels.
Specifically, we train a model where each point in the scene votes for the bounding box to which it belongs.
We devise a new  algorithm to cluster votes based on box volumetric overlap.  
Such overlap can be back-projected to the original scene points to obtain a probabilistic instance mask. 
Since we lack dense labels, we propose a training strategy where point to instance associations are approximated on the fly based on bounding boxes. An overview of Box2Mask is presented in Fig. \ref{fig:method_overview}.
We evaluate our approach on three challenging indoor 3D datasets: ScanNet\cite{Dai17CVPR}, S3DIS\cite{Armeni16CVPR} and ARKitScenes~\cite{Baruch21NIPS}.

In summary, our contributions are as follows:
\begin{itemize}
\item We propose a principled method leveraging bounding boxes both as a representation and to guide the training scheme.
This comprises a novel method for voting, instance clustering and training with weak labels.
Code, models and annotations are available at \url{virtualhumans.mpi-inf.mpg.de/box2mask}.
\item We present the first dense 3D instance semantic segmentation method trained with only bounding boxes. It is competitive with the best fully supervised baselines (97\% of HAIS\cite{Chen21ICCV} on ScanNet test, mAP$_{50}$) and it largely outperforms weakly supervised alternatives (+18\,mAP$_{50}$ compared to CSC\cite{Hou21CVPR}). On the largest scene dataset ARKitScenes annotated only with 3D bounding boxes, we obtain for the first time compelling 3D instance segmentation results. 
\end{itemize}

%% file: figures/annotationtypes/annotationtypes.tex
\begin{figure}[t]
    \centering
\begin{overpic}[width=\textwidth,,tics=10]
{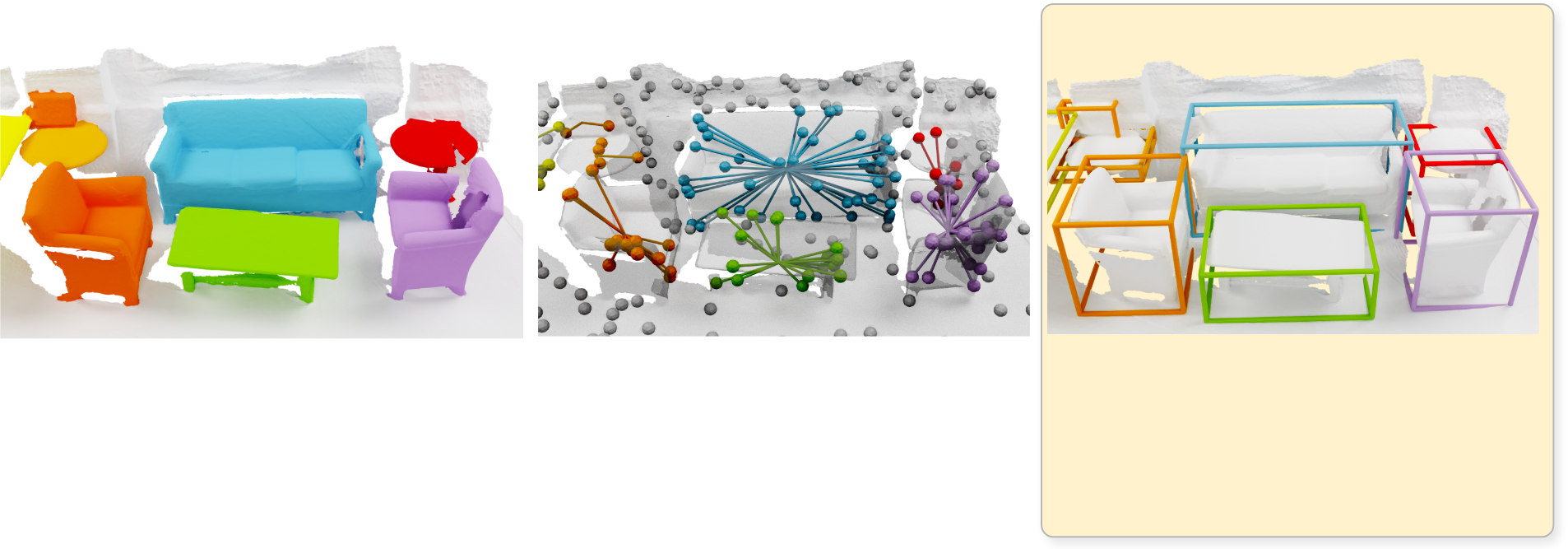}
\put(30,112){\footnotesize Dense Points}
\put(145,112){\footnotesize Sparse Points}
\put(255,112){\footnotesize Bounding Boxes}
\put(1,33){\footnotesize Predictions:}
\put(1,20.5){\footnotesize Labor:}
\put(1,8){\footnotesize Availability:}

\put(55,33){\footnotesize {\color{darkgreen} \ding{51}} Accurate}
\put(55,20.5){\footnotesize {\color{darkred} \ding{55}} Costly}
\put(55,8){\footnotesize {\color{darkred} \ding{55}} Few Datasets}

\put(137,33){\footnotesize {\color{darkred} \ding{55}} Less Accurate}
\put(137,20.5){\footnotesize {\color{darkgreen} \ding{51}} Cheaper}
\put(137,8){\footnotesize {\color{darkred} \ding{55}} Single Dataset}

\put(255,33){\footnotesize {\color{darkgreen} \ding{51}} Accurate}
\put(255,20.5){\footnotesize {\color{darkgreen} \ding{51}} Cheaper}
\put(255,8){\footnotesize {\color{darkgreen} \ding{51}} Many Datasets}

\end{overpic}
\vspace{-7mm}
    \caption{ 
    \textbf{Annotation Types.}
    Our key finding is that bounding box annotations serve as a surprisingly valuable annotation type for learning dense 3D instance masks.
    Prior work either requires per-point annotations \emph{(left)} with instance ids and semantic classes for millions of points, or initial weak supervision methods~\cite{Hou21CVPR} use sparse point annotations \emph{(middle)}, where a subset of points is annotated with instance centers and semantic classes.
    We propose to use bounding box annotations \emph{(right)}, where each object is annotated with its tight fitting box and a semantic label.
    We find boxes combine desirable properties: they allow for results on par with full supervision,
    reduce annotation effort to the object-level and are readily available in several large-scale 3D datasets \cite{Baruch21NIPS,Caesar20CVPR,Liang21CVPR,Hahlert20cityscapes}.  
}
    \label{fig:annotationtypes}
    \vspace{-5mm}
\end{figure}
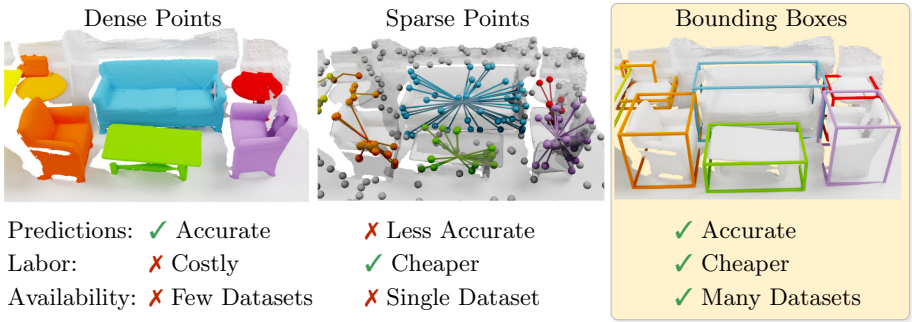

%% file: sections/2_related_work.tex
\section{Related Work}
\label{sec:related}
\vspace{-4mm}
\parag{Densely Supervised 3D Instance Segmentation.}
The first deep models for 3D instance segmentation] (SGPN\cite{Wang18CVPR}, 3D-BEVIS\cite{Elich19GCPR}, ASIS\cite{Wang19CVPR}) estimated instances by grouping learned point features in an abstract embedding space.
Extending this, MTML\cite{Lahoud19ICCV} proposes an additional learned directional embedding space.
All these methods require non-learned, computationally expensive point-feature clustering.
Similar to the popular MaskRCNN\cite{He17CVPR} for 2D instance segmentation, 3D-SIS\cite{Hou19CVPR} extracts bounding box proposals and extracts the per-voxel masks via a 3D-RoI layer.
An interesting alternative is proposed in 3D-BoNet\cite{Yang19NIPS} which, from a single global scene descriptor, directly predicts all object bounding boxes which are then segmented into foreground and background.
Both previous methods are sensitive to missed object detections since they cannot be recovered at later stages in the model.
More recently, several works group points based on predicted semantics and object centers \cite{Jiang20CVPR,Engelmann20CVPR,Chen21ICCV,Liang21CVPR,Han20CVPR}.
3D-MPA \cite{Engelmann20CVPR} first combines points into sets of points and then groups these into objects based on features learned via a graph convolutional network.
PointGroup\cite{Jiang20CVPR} combines sets of points based on both the original point positions and learned center positions.
OccuSeg\cite{Han20CVPR} additionally predicts instance occupancy as a proxy for the physical size of an instance.
Similar in spirit to \cite{Engelmann20CVPR}, HAIS\cite{Chen21ICCV} groups points into sets of points, and performs additional set refinement steps as well as scoring as in \cite{Jiang20CVPR}.
SSTNet\cite{Liang21CVPR} generalizes the over-segmentation idea from \cite{Han20CVPR} using a super-point semantic tree network which hierarchically merges segments into object instances. 
Different from the above methods, our \name{} does not estimate object centers but votes for object bounding boxes.
By doing so, the voting mechanism is robust towards varying object sizes (see \reffig{nmc}).

\parag{Weak Supervision for 3D Segmentation.}
Many efforts have been made to reduce the labeling cost of dense annotation for 2D images, with models learning from only weak annotations such as bounding boxes\cite{Dai15ICCV, khoreva17CPVR, Hsu19NIPS}, scribbles\cite{Lin16CVPR, Wang19IJCAI, Zhang2020CVPR}, points\cite{Bearman2016ECCV} and image-level labels\cite{Qi16ECCV, Joon17CVPR, Zhou18CVPR, Ahn18CVPR}.

In 3D, mostly semantic segmentation on point clouds was addressed with weak labels, where only sparsely labeled points are given \cite{Xu20CVPR, Hou21CVPR, Cheng21AAAI, Jiang2021CVPR, Liu21CVPR, Xu20CVPR, Zhang21AAAI, Zhang21ICCV}.
SSPC-Net\cite{Cheng21AAAI} and Liu \etal{}\cite{Liu21CVPR} assign sparse labels to super-points, construct graphs over super-points and learn to propagate semantic labels between nodes in the graph from labeled super-points to the unlabeled super-points.
In this fashion, and with only 10\% of labeled points, Xu \etal{}\cite{Xu20CVPR} achieve a performance close to fully supervised semantic segmentation methods.
Another line of work explores scene-level annotations or subcloud-level annotations for semantic segmentation \cite{Zhou16CVPR,Wei20CVPR,Ren21CVPR}.
Here, only a list of semantic classes contained in a scene (or part of it) are assumed, without precise localization. %
However, since scene and sub-scene labels are the coarsest annotations assumed, results typically lack details.

Only recently, the first work on weak supervision for 3D semantic instance segmentation was proposed, assuming a sparse number of points is annotated with their instance centers\,\cite{Hou21CVPR}.
To reduce data hungriness, unsupervised pre-training on 3D point clouds\cite{gidaris18ICLR, Xie20ECCV, li18CVPR} is used. 
PointContrast\cite{Xie20ECCV} improves supervised down-stream tasks significantly via contrastive pre-training on point clouds. CSC\cite{Hou21CVPR} additionally incorporates point-level correspondences and spatial contexts in a scene.
CSC achieves encouraging initial results, but still leaves a significant gap to fully supervised methods.

%% file: sections/3_method.tex
\input{figures/method_overview}

\section{Hough Voting for Bounding Boxes}
\label{sec:voting}
Our model votes for instances represented as bounding boxes (Fig. \ref{fig:method_overview}).
This is unlike prior work, which represents instances as centers.
Our experiments show that the proposed box representation has several advantages over centers.

\textit{Encoding.} 
Input are scene points $\mathcal{S}\in \mathbb{R}^{N \times F}$, where $N$ is the number of scene points and $F$ the number of per-point input features ($F$\,=\,$9$ in our experiments, for position, color and estimated surface normal).
We use the popular sparse convolutional U-Nets~\cite{Choy19CVPR,Graham18CVPR} as backbone, to obtain per scene point features.  
Those require discretization of point positions into regular grids, but allow for high resolution (we use 2\,cm).
In case multiple points fall into the same grid location, instead of averaging features and associated ground truth annotations leading to a blurry combination, we pick the features of the nearest neighboring scene point.
We retain this mapping to allow reversing the discretization later.

\textit{Decoding.}
Based on the point features, we predict a bounding box, a score, and a semantic label per scene point.
We branch out decoding with separate (3 layers) MLP networks.
We predict axis-aligned bounding boxes, parameterized using their centers, and sizes (width, height, depth), with one MLP for center and one for size. 
Similarly, another MLP predicts a scalar score, estimating the intersection-over-union of the predicted bounding box with the ground truth, and a fourth branch predicts the semantic label (``chair", ``table",  ...). 

\section{Clustering and Back-projection}
\label{sec:clustering}
\input{figures/distancemetric/iou_clustering}

Next, we want to turn our model predictions into instance masks. 
Since, scene points vote for the bounding box of their instance, all points voting for the same box define an instance and points voting for a different box define a different instance.
However, due to noisy predictions, box votes from points of the same instance will not be perfectly aligned which demands a clustering strategy.

In contrast to clustering based on centers with Euclidean distance metrics (as commonly employed in 3D instance segmentation), we make full use of our bounding box votes, by defining a novel 3D clustering method based on volumetric similarity (Fig.\,\ref{fig:nmc} A,\,B). 
Specifically, we define our clustering similarity-metric over two bounding box votes, $\bbox_a$ and $\bbox_b$, as the intersection-over-union (IoU):
\begin{equation}
\label{eq:vote_space_similarity}
\text{vote-space similarity: } \mathrm{IoU}(\bbox_a,\bbox_b) = \frac{\text{area of overlap}}{\text{area of union}}.
\end{equation}
Voting and clustering of bounding boxes has two key benefits:
First, IoU allows to separate two instances when no box overlap is present, which requires careful handcrafted thresholds for center voting (see Fig.~\ref{fig:nmc}).
Second, while center clustering will fail when two instances have the same center (\eg{}, an apple in a bowl), boxes additionally distinguish instance size.
\vspace{-2mm}

\parag{Clustering Volumetrically.} 
First, all bounding box votes are sorted in descending order according to the predicted scores.
Then, the highest-scoring box is picked and serves as the representative, $\bbox_r$, of the first cluster.
All boxes, $\bbox$, that are sufficiently similar to the representative, $\mathrm{IoU}(\bbox,\bbox_r) > \tau$, are assigned to this cluster.
Higher values of $\tau \in (0,1)$ will result in numerous smaller clusters and lower values will result in fewer larger clusters.
The next step is to take the next highest scoring box that has not yet been clustered: 
it will serve as next representative.
This process is repeated until all boxes are assigned to a representative or are chosen as representative themselves.  
We call this clustering \emph{non-maximum clustering} (NMC).
A pseudo-code description is given in the appendix. 

\vspace{-2mm}
\parag{Back-projection to Instance Masks.}
Ultimately, we are interested in clusters in the original point cloud.
Therefore, we \textit{back-project} each clustered bounding box to the point that voted for it. 
All points that voted for boxes within the same cluster form an instance mask. 
For semantic instance segmentation, each instance mask should be accompanied with a semantic class and a score.
Since our model predicts semantics for each scene point, we obtain instance labels by performing a majority-vote per mask.
For the score, we rely on the predicted IoU score of the point that voted for the instance's cluster representative $\bbox_r$.

\section{Training with Weak Bounding Box Labels}
\label{sec:bb_associations}
\vspace{-1mm}
Fully supervised 3D instance segmentation methods rely on densely annotated point clouds and learn to predict at each scene point $\point$ some ground truth value, $gt(\point)$ (\eg{} instance center).
However, this strategy cannot be applied when a scene is annotated only with a set of bounding boxes.
More specifically, boxes do not define instance ground truth on a point-level, such that $gt(\point)$ is unclear.
We address this issue by finding a strategy to approximate point-to-bounding box associations. 
More formally, let $\boxes = \{\bbox_1,...,\bbox_B\}$ , $\bbox_i = [\text{center},\text{size},\text{label}] \in \mathbb{R}^6\times \mathbb{N}$ be the set of annotated boxes in a scene, we define a function
$\asso: \points \rightarrow \boxes $
which maps a 3D scene point $\point \in \points$ to a ground truth bounding box $\asso(\point) \in \boxes$.
Once such a function is found, the model can be trained in a similar fashion as fully supervised models, replacing the exact point-to-point ground truth $gt(\point)$ with our approximate point-to-box ground truth $gt\bigl(\asso(\point)\bigr)$.

\parag{How Should the Mapping Function $\asso$ Be Defined?}
Since ideally, an object bounding box contains all the points of its instance,
the possible box associations of a point are reduced to only those boxes containing it.
In turn, if a point is contained in no bounding box, it can only be part of the background (\eg, wall, ceiling, floor).
This simple observation has an important effect:
with high certainty, we can learn to segment (or discard) non-instance points, a crucial part of instance segmentation.
We can specify our approximate associations further for points contained in only a single box:
all those points will actually belong to the instance of the box, up to points from non-annotated background points.
If a point, however, is located in multiple bounding boxes, we cannot get exact point-to-box association.
These observations can be formulated into our initial approximate association function:
\vspace{-3px}
\begin{equation}
\label{eq:associate_func}
    \asso(\point)= 
    \begin{cases}
        \text{background},& \text{if $\point$ is not contained in any $\bbox \in \boxes$}\\ 
        \bbox,          & \text{if $\point$ is only contained in a single $\bbox \in \boxes$}\\  
        \text{undecided} &  \text{else }\\
    \end{cases}
\end{equation}
and updates the co-domain of $\asso$ to $\boxes \cup \{\mathrm{background,undecided}\}$. These associations are already surprisingly useful for supervising on \textit{decided} points only (i.e. none or single box points): in experiments, this initial strategy achieves 87\% of current fully-supervised methods with dense per-point labels.

A key remaining question however is: can we increase prediction quality by integrating approximate associations for points in multiple boxes?
In our analysis (Tab. \ref{tab:association_strategies}), we find that choosing the smallest of multiple available boxes improves over other strategies.
This makes sense since smaller objects are often fully contained in bounding boxes of larger objects (a pillow on a sofa, a sink in a cabinet).
Using this strategy, and only relying on bounding box annotations, we achieve 97\% of the performance of fully-supervised methods trained with dense per-point labels.

\subsubsection*{Losses.}
\label{sec:losses}
Let $\points$ be the set of scene points and $\boxes$ the set of annotated bounding boxes. Using our association function, $\asso$, we define our losses, only given the box annotations.
We define our instance losses only for points associated with the scene foreground $\fg \defeq \{\point \in \points | \asso(\point) \in \boxes\}$, excluding ``background'' and ``undecided'' points. 
Our instance prediction losses are defined as: 
\begin{equation}
\begin{aligned}
 &\mathcal{L}_\text{offset} \defeq \frac{1}{|\fg|}   \sum  \limits_{\point \in \fg} {\parallel \bboffsets\bigl(\point, \asso(\point)\bigr) - \widehat{\bboffsets}(\point) \parallel}_1\text{,} \\
 &\mathcal{L}_\text{size} \defeq \frac{1}{|\fg|}   \sum  \limits_{\point \in \fg} {\parallel \bbsize\bigl(\asso(\point)\bigr) - \widehat{\bbsize}(\point) \parallel}_1\text{,} \\
  &\mathcal{L}_\text{score} \defeq \frac{1}{|\fg|} \sum \limits_{\point \in \fg} \text{CE}\Bigl( \iou\bigl(\asso(\point)\bigr), \widehat{\iou}(\point)\Bigr)\text{,} \\
\end{aligned}
\end{equation}
where $\bboffsets$ is the offset from $\point$ to the center of its associated bounding box, $\asso(\point)$; $\bbsize$ is the size (width, height, depth) and $\iou$ is the IoU of the predicted bounding box with the associated box, $\asso(\point)$.
We denote the predicted values with a hat and the cross-entropy loss with CE.
Similarly, the dense semantic segmentation is learned from only bounding boxes, relying on their semantic label. In contrast to above instance losses, the semantic loss, $\mathcal{L}_\text{sem}$, includes the points associated with the background $\decided \defeq \{\point \in \points | \asso(\point) \neq \text{undecided}\}$: 
\begin{equation}
 \mathcal{L}_\text{sem} \defeq  \frac{1}{|\decided|}   \sum  \limits_{\point \in \decided}\text{CE}\Bigl( \sem\bigl(\asso(\point)\bigr), \widehat{\sem}(\point)\Bigr).
\end{equation}
where $\sem$ defines the ground truth, including a generic semantic class "background" for all points associated with it:
\begin{equation}
    \sem(\point) \defeq 
\begin{cases}
    \text{background\_class},   & \text{if $\point$ in no box}\\
    \text{label}\bigl(\asso(\point)\bigr),   & \text{else}\\
\end{cases}
\end{equation}
Importantly, this allows us at inference time, to predict and filter background points, not defining any instances.
Our network prediction consists of a forward pass, fully implemented with convolutions and trained end-to-end with the combined, multi-task loss defined as $\mathcal{L} := \mathcal{L}_\text{offset} + \mathcal{L}_\text{size} + \mathcal{L}_\text{score} + \mathcal{L}_\text{sem}$.

\subsection{Implementation and Training Details}
We train our network end-to-end and from scratch with the Adam optimizer,
using an initial learning rate of 0.001, a batch size of 8 entire scenes,
and train for 500 epochs on a single NVIDIA Quadro RTX 8000.
For data augmentation, scenes are randomly rotated around height, flipped, and scaled in $\text{Uniform[0.9, 1.1]}$.
Our backbone is a 6-layer sparse-convolutional encoder-decoder including skip connections based on \cite{Chen21ICCV}.
The MLP heads are implemented using 3 layers with 96 hidden units.
Similar to other current segmentation methods\cite{Liang21CVPR,Han20CVPR,Nekrasov213DV}, we perform point over-segmentation \cite{Felzenszwalb04IJCV,Karpathy04IJCV} on ScanNet and ARKitScenes, and similarly employ \cite{Landrieu18CVPR, Landrieu19CVPR} on S3DIS.
This reduces the number of votes by averaging over segments before clustering, alleviating the computational load.
Empirically, we set the NMC clustering threshold $\tau$\,$=$\,$0.3$.
More details are in the appendix.

%% file: figures/method_overview.tex
\begin{figure}[t]
\centering
\includegraphics[width=\textwidth]{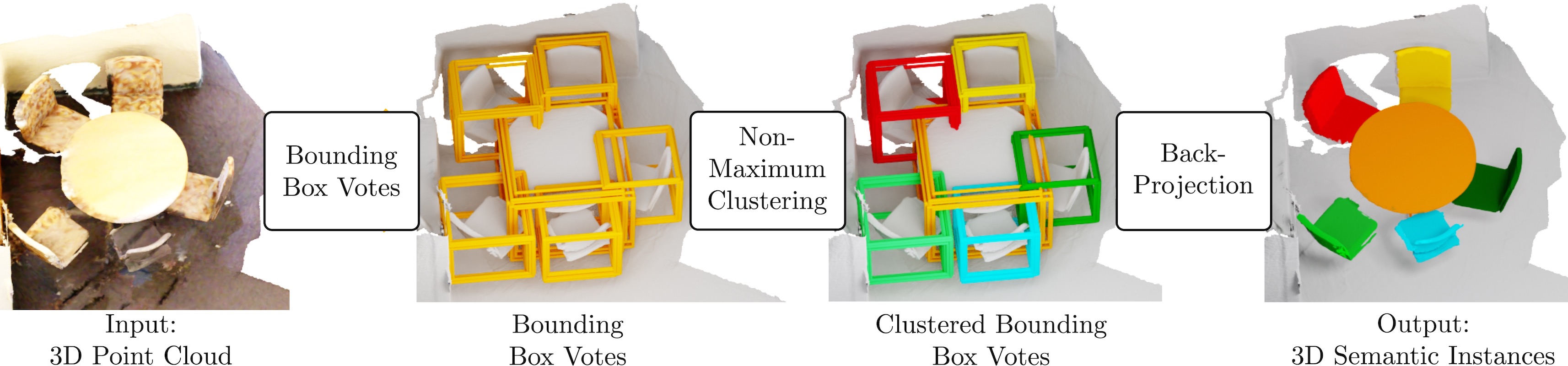}
\vspace*{-5px}
\caption{
\textbf{\name{} Overview.}
Input to Box2Mask is a colored 3D point cloud of a scene. 
\textbf{Bounding Box Voting:} For each point in the input scene, our model predicts the points instance, parameterized as 6-DoF bounding box. 
The key contribution is the training procedure with only coarse bounding box labels (requiring no per-point labels) by associating points to bounding box labels. 
\textbf{Non-Maximum Clustering:} Votes are clustered using our Non-Maximum Clustering (NMC) that is specifically tailored to the bounding box representation. 
\textbf{Back-Projection:} A point is associated with the cluster of the box it predicted. Doing this for each point yields the final instance masks.
}
\vspace{-15px}
\label{fig:method_overview}
\end{figure}


%% file: figures/distancemetric/iou_clustering.tex
\begin{figure}[t]
    \centering
    \begin{overpic}[ width=1\textwidth]{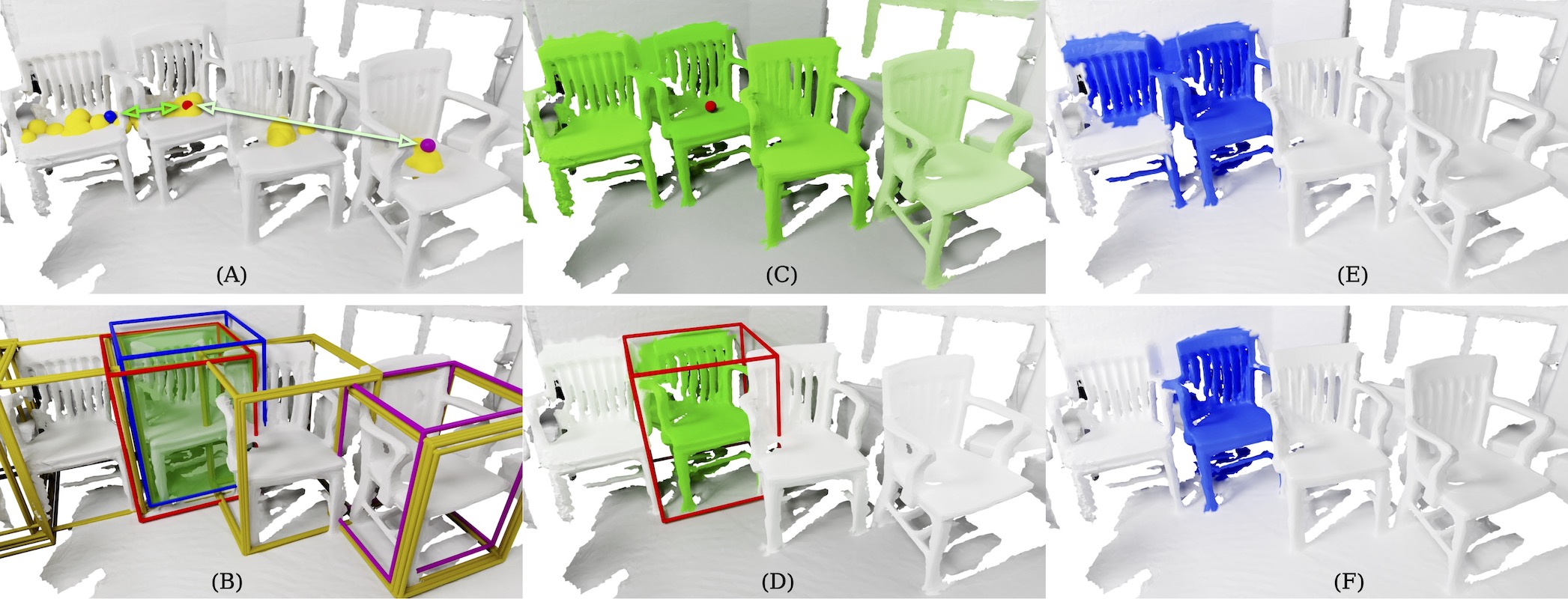}

    \end{overpic}
    
        \vspace{-2mm}
    \caption{\textbf{Center Clustering (top row) \vs{} Our Box Clustering (bottom row).}
    \textbf{(A)} Scene points predict their instance centers (yellow). For clustering, the Euclidean metric between votes is used (arrows between highlighted votes). \textbf{(B)} In contrast, we propose to use IoU (Eq. \ref{eq:vote_space_similarity}) on bounding box votes. Intersection of blue and red votes visualized in green. Since boxes define the extend of objects, this metric can discriminate distinctively when overlap is not present, as with the violet box vote. This is key for obtaining sharp similarity decay of the scene points \textbf{(D)}, instead of smooth decay with distance \textbf{(C)}. The latter is sensitive to errors in binarization thresholds (handcrafted and dataset specific) when converting to instance masks \textbf{(E)}. In contrast, our method naturally encodes this threshold via box sizes, \ie, converting from \textbf{(D)} to \textbf{(F)} is robust. 
    }
    \label{fig:nmc}
    \vspace{-9mm}
\end{figure}

%% file: sections/4_experiments.tex
\newpage
\section{Experiments}

\subsection{Comparing with State-of-the-art Methods.}
\label{sec:comparison_sota}
\vspace{-5px}
\parag{Datasets.} {S3DIS}\cite{Armeni16CVPR} consists of 272 scans of 6 large-scale indoor areas collected from three different buildings.
Scans are represented as point clouds and points are annotated with instance- and semantic-labels out of 13 object classes.
To obtain bounding box annotations from masks, we use the standard approach of \cite{Engelmann20CVPR, He21CVPR, Hou19CVPR, Hou21CVPR, Qi19ICCV, Xie20CVPR, Yang19NIPS, Zhang20ECCV}, \ie, we obtain axis-aligned bounding boxes from the instance point annotations.
We report scores on Area-5 and 6-fold cross-validation.

{ScanNet}\cite{Dai17CVPR} is a richly annotated dataset of 3D reconstructed indoor scenes represented as meshes.
Similar to S3DIS, each scene is annotated with semantic- and instance-segmentations of
18 object categories.
It consists of 1201 training scenes, 312 validation scenes, and 100 hidden test scenes.
Bounding box annotations are obtained the same way as for S3DIS.

{ARKitScenes}\cite{Baruch21NIPS} is the largest of these datasets with 4499 training scenes and 550 validation scenes.
The scenes are represented as reconstructed meshes and are recorded in real-world homes.
The dataset is annotated with oriented object bounding boxes across 17 semantic classes.
Importantly, per-point labels are not available.
Nevertheless, our approach is able to leverage the bounding box annotations as weak supervision signal, which is an \textit{immense practical advantage} over existing 3D instance segmentation methods which require dense per-point annotations and can therefore not be trained on this dataset.

\parag{Methods in Comparison.}
We compare to both fully-supervised and weakly-supervised SOTA prior methods.
Fully-supervision methods are the majority: we compare top-down segmentation methods 3D-BoNet\cite{Yang19NIPS}, 3D-SIS\cite{Hou19CVPR} and bottom-up methods MTML\cite{Lahoud19ICCV},
PointGroup\cite{Jiang20CVPR}, 3D-MPA\cite{Engelmann20CVPR}, OccuSeg\cite{Han20CVPR}, HAIS\cite{Chen21ICCV} and
SSTNet\cite{Liang21CVPR}. See Sec. \ref{sec:related} for more details.

Weakly-supervised methods are much less, and only recently explored.
PointContrast\cite{Xie20ECCV} and CSC\cite{Hou21CVPR} both make use of unsupervised pre-training via contrastive-learning. Compared to PointContrast, CSC follows a more sophisticated approach by taking spatial scene context into account.
For 3D instance segmentation, the pre-trained models are supervised with a limited number of sparsely annotated points (20, 50, 100 or 200 Points), for which the ground truth object centers and semantic classes are known during training.

\input{tables/scannet_comparison}

\parag{Results} on S3DIS and ScanNet are summarized in \reftab{comparision_scannet_s3dis}.
Our approach improves upon prior (point-based) weakly-supervised methods \cite{Xie20ECCV, Hou21CVPR} by more than \textbf{10~mAP}.
While sparse point labels and bounding box labels might not be directly comparable,
it is noteworthy that this improvement is achieved without pre-training as used by \cite{Xie20ECCV, Hou21CVPR}.
Compared to fully-supervised approaches, our weakly-supervised method achieves {92\%} and {94\%} of the performance of leading methods on ScanNet (SSTNet, val, mAP$_{50}$) and S3DIS (HAIS, A5, mPrec) respectively.
This is extremely encouraging, as it indicates that densely labeled points might not be entirely necessary.
Qualitative ScanNet results are shown in \reffig{qualitative_scannet}.
Our method predicts clear masks in heavily cluttered environments and accurately segments even very large objects like tables.
The difference between weak- and full-supervision is marginal, however, bounding boxes need only be annotated on object-level in contrast to per-point annotations.
Additional qualitative results and analysis, including S3DIS, are in the appendix.

Quantitative results on ARKitScenes are shown in \reftab{arkitscenes_scores}, visual results in \reffig{qualitative_arkitscenes}.
As per-point instance labels are not available, we cannot report segmentation scores.
Instead, as a proxy, we compare to recent object detection methods\cite{Qi19ICCV, Zhang20ECCV, Xie20CVPR} by fitting oriented bounding boxes to our predicted masks.
This indirectly measures mask quality.
However, high detection scores are only obtained if the predicted point masks are accurate. 
Therefore, the correctness of position and size of the masks are measured.
Our approach achieves leading performance among all methods (+\textbf{4\,mAP}) suggesting good quality masks.

\parag{Limited Annotations 3D Semantic Instance Benchmark.}
On this benchmark, the ground truth labels are given for only a limited number of annotated points per scene.
We compare to the baseline methods introduced in \cite{Hou21CVPR}.
These methods perform instance segmentation by predicting centers, which means that they rely on annotated centers (see Fig. \ref{fig:annotationtypes}, middle).
Instead, our approach relies on bounding box annotations.
We believe that bounding boxes are more realistic and easier to annotate than 3D object centers, which are usually located somewhere in empty space and can be hard for an annotator to precisely locate.
Results are shown in \reftab{data_efficient_benchmark}.
Our approach consistently outperforms prior work with a large margin, 
even without relying on any pre-training.

\newpage
\input{tables/arkitscenes_object_detection}
\input{figures_tex/quali_arkitscenes}

\newpage
\begin{table}[]
    \begin{center}
\setlength{\tabcolsep}{6pt}
\begin{footnotesize}
\resizebox{\textwidth}{!}{
    \begin{tabular}{l|llll}
    \toprule
      mAP$_{50}$                          & 200 points & 100 Points & 50 Points & 20 Points \\
    \midrule
        CSC trained from scratch \cite{Hou21CVPR}                        & 46.4  & 41.8  & 31.1 & 20.0 \\
        PointContrast$^*$\cite{Xie20ECCV}   & 47.1  & 45.6  & 40.0 & 25.9 \\
        CSC$^*$\cite{Hou21CVPR}             & 49.4  & 46.0  & 41.4 & 28.9 \\
        Ours                            &
        \textbf{59.2} {\color{darkgreen}\scriptsize (+9.8)} &
        \textbf{56.5} {\color{darkgreen}\scriptsize (+10.5)} &
        \textbf{49.8} {\color{darkgreen}\scriptsize (+8.4)} &
        \textbf{46.5} {\color{darkgreen}\scriptsize (+17.6)} \\
        \bottomrule
    \end{tabular}
}
    \vspace{-6px}
\end{footnotesize}
    \end{center}
    \caption{\textbf{ScanNet Data Efficient Benchmark Test.}
    Instance segmentation on limited annotations (LA).
    Scores as in \cite{Hou21CVPR}.
    Star $(^*)$ indicates usage of pre-training.}
    \label{tab:data_efficient_benchmark}
        \vspace{-16mm}
\end{table}

\input{figures_tex/quali_scannet}

\newpage
\subsection{Analysis}
\label{sec:analysis}
\vspace{-10px}
\parag{Boxes or Centers?} An important \\
\vspace{-11px}
\begin{wrapfigure}{r}{0.50\textwidth}
\begin{minipage}{0.50\textwidth}
\vspace{-15px}
    \input{tables/centers_vs_bbs}
\end{minipage}
\end{wrapfigure}
baseline to the proposed \emph{bounding-box}
representation is the popular \emph{center} representation \cite{Qi16ECCV, Jiang20CVPR, Hou21CVPR, Chen21ICCV, Liang21CVPR}.
We also analyse different techniques for clustering the voting space and
compare to the proposed \emph{non-maximum clustering} (NMC).
Spatial clustering (SC), such as breadth-first search as in \cite{Jiang20CVPR} or DBScan as in \cite{Engelmann20CVPR}, groups votes based on their pairwise Euclidean distance.
Further, it is common practice to cluster votes separately \emph{per semantic class},
which ensures that points of disagreeing semantics are in different instances.
\reftab{inst_rep} shows that clustering conditioned on the semantic class is beneficial only for centers.
This indicates that box votes already encode sufficient semantics (via the size) increasing robustness to wrongly predicted semantics. 
More importantly, the proposed bounding boxes consistently outperform centers,
suggesting that object size is important for vote clustering.
The largest improvement is observed by NMC over SC.
While SC treats all dimensions in the voting space equally,
NMS is tailored to bounding boxes, using the actual geometric meaning of each feature dimension in the voting space.

\parag{Weak Supervision Analysis.} We \\
\vspace{-11px}
\begin{wrapfigure}{r}{0.5\textwidth}
\begin{minipage}{0.5\textwidth}
\vspace{-15px}
    \input{tables/association_strategies}
\end{minipage}
\end{wrapfigure}
introduced \emph{undecided points} as
points inside multiple ground truth bounding boxes.
\emph{Decided} points are either supervised as background (if they are in no box) or with the single box they are in (\emph{c.f.} Eq.2).
For all others, the undecided points, we compare multiple heuristics, as summarized in Tab. \ref{tab:association_strategies}.
The simplest baseline (1) does not supervise undecided points at all, which results in 56 mAP$_{50}$.
This is already 87\% of the performance of the fully-supervised state-of-the-art SSTNet\cite{Liang21CVPR} (64.3 mAP$_{50}$).
We then compare two additional heuristics:
points that are in multiple ground truth bounding boxes are supervised with the closest bounding box in terms of distance to the center (2), and the smallest bounding box in terms of volume (3).
The additional supervision improves scores by +3.7 mAP while the smallest box performs a bit better than the closest. Importantly, using these associations, 
our weakly supervised model obtains {97\%} of the performance of a comparable fully supervised model which shows that coarse bounding box annotations are surprisingly strong supervision signal compared to dense per-point annotations.

\parag{Effect of Noisy Box Labels.}
Since training bounding boxes on ScanNet are obtained from point masks, they are perfectly aligned to the points -- an accuracy a human annotation might not achieve.
This motivates an experiment on the robustness towards more incorrect labels.
We trained separate models on annotation with missing labels (levels 0 to 10\%) and inaccurate placement

{\color{white}.}\\
\begin{minipage}[b]{0.49\textwidth}
\vspace{-36px}
\begin{figure}[H]
\vspace{-30px}
    \includegraphics[width=0.49\textwidth]{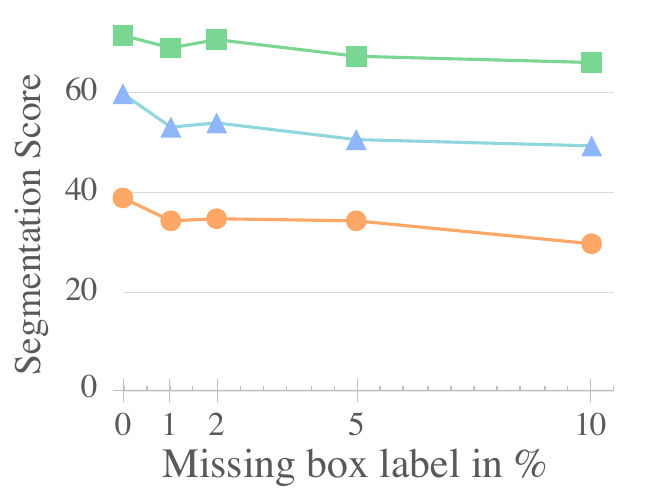}
    \includegraphics[width=0.49\textwidth]{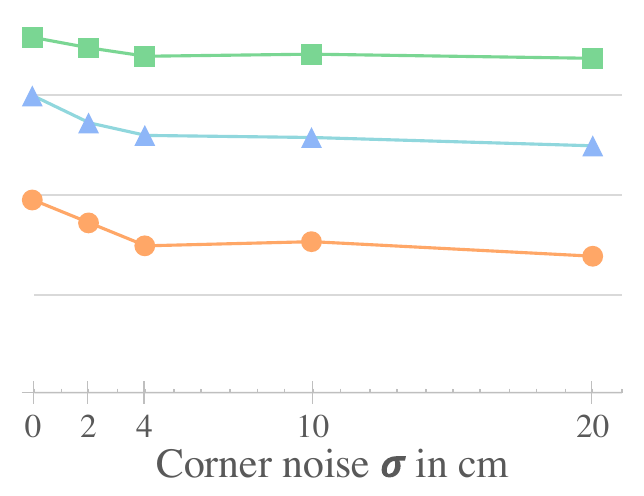}
    \vspace{-20px}
    \caption{\small
\textbf{Reduced Annotation Quality.} Semantic instance segmentation scores on ScanNet val. trained with missing box labels \emph{(left)} and noisy box labels \emph{(right)}.
}
    \label{fig:noisy_boxes}
\end{figure}
\end{minipage}
\raisebox{48px}{
\begin{minipage}{0.49\textwidth}
\vspace{-36px}
\input{tables/fully_supervised_comparison}

\end{minipage}
}
(0 to 20 cm error in box corners).
See \reffig{noisy_boxes} for the results.
We observe good robustness, with only around 4 mAP differences.

\vspace{-4px}
\parag{Fully Supervised Setting.}
Our model can also be adapted to the fully supervised setting, where dense per-point labels are available. The association function $a$ returns the corresponding ground truth point label. Our model compares favorably to recent state-of-the-art approaches, as summarized in \reftab{fully_supervised_comparison}.

\vspace{-4px}
\parag{Is a Detection Model Enough?}
As  a simple baseline, instead of using the box annotations for directly training instance segmentation, we train a detection model that predicts one box per object.
We obtain an instance mask via post-processing with the best performing box-to-point association strategy (Tab. \ref{tab:association_strategies}), which was also used for weak supervision of our model. 
Our proposed approach largely outperforms this baseline quantitatively (+11.8 mAP$_{50}$ on ScanNet) as well as qualitatively, see appendix (Sec.\,A) for details.
This suggests that our model generalizes beyond the weak point associations, to complete object priors.

\label{exp_association_strategies}

%% file: tables/scannet_comparison.tex
\begin{table}[t]
\setlength{\tabcolsep}{3px}%
\begin{center}
\resizebox{\textwidth}{!}{
\begin{tabular}{l l c c c c c c c c c c c c}
\toprule
& \multirow{2}{*}{\hspace{3mm} } &  & \multicolumn{2}{c}{ScanNet} & & \multicolumn{2}{c}{ScanNet}
& & \multicolumn{2}{c}{S3DIS} & & \multicolumn{2}{c}{S3DIS}\\
& \multirow{2}{*}{\hspace{3mm} } &  & \multicolumn{2}{c}{Validation} & & \multicolumn{2}{c}{Hidden Test}
& & \multicolumn{2}{c}{Area\,5} & & \multicolumn{2}{c}{6-fold CV}\\
\cmidrule(r){4-5} \cmidrule(r){7-8} \cmidrule(r){10-11} \cmidrule(r){13-14}
&  &  & {\footnotesize mAP} & {\footnotesize mAP} & & {\footnotesize mAP} & {\footnotesize mAP}\\
& Method & Supervision & {\footnotesize @50\%} & {\footnotesize @25\%} & & {\footnotesize @50\%} & {\footnotesize @25\%} & & mPrec & mRec & & mPrec & mRec \\
\midrule
\multirow{9}{*}{\rotatebox{90}{Weakly Supervised}}
& CSC\cite{Hou21CVPR}  & 20 Points  & 26.3 & -- & & 28.9 & 49.6 & & & & & &  \\
& PointContrast\cite{Xie20ECCV}   & 20 Points  & -- & -- & & 25.9 & 47.4 & &  &  &  &   &  \\
& CSC\cite{Hou21CVPR}  & 50 Points  & 32.6 & -- & & 41.4 & 62.0 & &  &  &  & &   \\
& PointContrast\cite{Xie20ECCV}   & 50 Points  & --   & -- & & 40.0 & 60.3 & &  &  &  & & \\
& CSC\cite{Hou21CVPR}  & 100 Points & 39.9 & -- & & 46.0 & 65.4 & & & & & & \\
& PointContrast\cite{Xie20ECCV}   & 100 Points & --   & -- & & 45.6 & 63.7 & & & & & & \\
& CSC\cite{Hou21CVPR}  & 200 Points & 48.9 & -- & & 49.4 & 70.2 & & & & & &  \\
& PointContrast\cite{Xie20ECCV}   & 200 Points & 44.5 & -- & & 47.1 & 66.2 & & & & & &  \\
& \name{} (Ours)        & Boxes      & \textbf{59.7} & \textbf{71.8} & & \textbf{67.7} & \textbf{80.3} & & 66.7 & 65.5 &  & 72.2  & 70.5  \\
& \footnotesize{Relative to SOTA}        &       & \emph{\scriptsize $92.8$\%} & \emph{ \scriptsize $95.0$\%} & & \emph{\scriptsize $96.9$\%} & \emph{ \scriptsize $100$\%} & & \emph{ \scriptsize $93.8$\%} & \emph{ \scriptsize $99.8$\%} &  & \emph{ \scriptsize $98.2$\%}  & \emph{ \scriptsize $96.0$\%}  \\
\midrule
\multirow{8}{*}{\rotatebox{90}{Fully Supervised}}
& 3D-SIS\cite{Hou19CVPR}       & All Points & 18.7 & 35.7 & & 38.2 & 55.8 & &  &  &  &   &   \\
& 3D-BoNet\cite{Yang19NIPS}    & All Points & -- 	& --   & & 48.8 & 68.7 & & -- & -- &  & 65.6 & 47.6  \\
& MTML\cite{Lahoud19ICCV}      & All Points & 40.2 & 55.4 & & 54.9 & 73.1 & &  &  &  &   &   \\
& PointGroup\cite{Jiang20CVPR} & All Points & 56.9 & 71.3 & & 63.6 & 77.8 & & 61.9 & 62.1 &  & 69.6  & 69.2  \\
& 3D-MPA\cite{Engelmann20CVPR} & All Points & 59.1 & 72.4 & & 61.1 & 73.7 & & 46.7 & \textbf{65.6} &  & 66.7 & 64.1  \\
& OccuSeg\cite{Han20CVPR}     & All Points & 60.7 & 71.9 & & 63.4 & 73.9 & & --   & --   &  & 72.8 & 60.3  \\
& HAIS\cite{Chen21ICCV}        & All Points & 64.1 & \textbf{75.6} & & \textbf{69.9} & \textbf{80.3} & & \textbf{71.1} & 65.0 &  & 73.2 & 69.4  \\
& SSTNet\cite{Liang21CVPR}     & All Points & \textbf{64.3} & 74.0 & & 69.8 & 78.9 & & 65.5 & 64.2 &  & \textbf{73.5} & \textbf{73.4} \\
\bottomrule
\end{tabular}
}
\end{center}
\vspace{-5px}
\caption{
\small
\textbf{State-of-the-art 3D Semantic Instance Segmentation}.
We show fully-supervised methods (dense point annotations) and weakly-supervised methods (sparse points and bounding boxes) on ScanNet\cite{Dai17CVPR} and S3DIS\cite{Armeni16CVPR}.\,\cite{Xie20ECCV} is as reported in \cite{Hou21CVPR}.}
\label{tab:comparision_scannet_s3dis}
\vspace{-20px}
\end{table}

%% file: tables/arkitscenes_object_detection.tex
\begin{table}[!ht]
\begin{center}
\resizebox{\textwidth}{!}{
\footnotesize
\setlength{\tabcolsep}{4pt}
\begin{tabular}{l|ccccccccc}
\toprule
                           & Cabinet & Refrig.      & Shelf         & Stove         & Bed           & Sink          & Washer        & Toilet        & Bathtub \\
\midrule
VoteNet\cite{Qi19ICCV}   & 37.1 & 62.7            & 12.4          & 0.3           & 85.0          & 31.1          & 45.3          & 75.5          & 93.3 \\
H3DNet\cite{Zhang20ECCV} & 40.2 & 59.4            & 10.0          & 1.6           & \textbf{88.2} & 40.1          & 49.0          & 83.8          & 93.0 \\
MLCVNet\cite{Xie20CVPR}  & 45.1 & \textbf{70.0}   & 16.9          &  2.4          & 88.0          & \textbf{40.2} & 51.5          & 85.9          & \textbf{94.1} \\
\midrule
\name{}(ours)            & \textbf{45.9} & 62.6   & \textbf{28.0} & \textbf{5.2}  & 87.1          & 30.6          & \textbf{53.8} & \textbf{89.4} & 92.9\\
\midrule
                            & Oven          & Dishw.        & Fireplace     & Stool         & Chair         & Table         & Monitor       & Sofa & \cellcolor{blue!5}\textbf{mAP} \\
\midrule
VoteNet\cite{Qi19ICCV}    & 18.3          & 2.9           & 22.1          & 3.0           & 20.1          & 31.0          & 0.6           & 68.3 &\cellcolor{blue!5}35.8 \\
H3DNet\cite{Zhang20ECCV}  & 24.1          & \textbf{3.9}  & 19.5          & 8.8           & 25.2          & 32.2          & 1.5           & 70.4 &\cellcolor{blue!5}38.3 \\
MLCVNet\cite{Xie20CVPR}   & 24.2          & 3.0           & 38.5          & 8.0           & 31.5          & 36.6          & 4.1           & 71.9 &\cellcolor{blue!5}41.9 \\
\midrule
\name{}\,(ours)             & \textbf{28.1} & 3.8           & \textbf{59.9} & \textbf{20.8} & \textbf{35.2} & \textbf{60.3} & \textbf{7.3}  & \textbf{82.8} &\cellcolor{blue!5}\textbf{46.7}\\
\bottomrule
\end{tabular}
}
\end{center}
\vspace{-5px}
\caption{
{\small
\textbf{Whole-scene 3D Object Detection Scores on ARKitScenes\cite{Baruch21NIPS}.}
The ground truth includes only oriented bounding box annotations, no point-level instance masks.
Therefore, we cannot directly compute instance segmentation scores.
Instead, as a proxy, we compare to recent object detection methods by fitting an oriented bounding box containing our predicted masks.
We report the average precision on the validation set with an IoU threshold of 50\% as in \cite{Qi19ICCV}.
All other scores are as reported in \cite{Baruch21NIPS}.}}
\label{tab:arkitscenes_scores}
\vspace{-7px}
\end{table}

%% file: figures_tex/quali_arkitscenes.tex
\definecolor{cabinet}{rgb}{0.1215,0.4666,0.7058}
\definecolor{bed}{rgb}{1.0,0.7333,0.4705}
\definecolor{chair}{rgb}{0.7372,0.7411,0.1333}
\definecolor{sofa}{rgb}{0.5490,0.3372,0.2941}
\definecolor{table}{rgb}{1.0,0.5960,0.5882}
\definecolor{shelf}{rgb}{0.2588,0.7372,0.4}
\definecolor{stove}{rgb}{0.7921,0.7254,0.2039}
\definecolor{washer}{rgb}{0.2,0.6901,0.7960}
\definecolor{oven}{rgb}{0.7843,0.2117,0.5137}
\definecolor{dishwasher}{rgb}{0.3607,0.7568,0.2392}
\definecolor{fireplace}{rgb}{0.3058,0.2784,0.7176}
\definecolor{stool}{rgb}{0.6745,0.4470,0.3215}
\definecolor{refrigerator}{rgb}{1.0,0.4980,0.0549}
\definecolor{tv_monitor}{rgb}{0.3568,0.6392,0.5411}
\definecolor{toilet}{rgb}{0.1725,0.6274,0.1725}
\definecolor{sink}{rgb}{0.4392,0.5019,0.5647}
\definecolor{bathtub}{rgb}{0.8901,0.4666,0.7607}

\begin{figure}[!hb]
    \vspace{-40px}
    \centering
    
    \scriptsize

    {\includegraphics[width=0.24\textwidth]{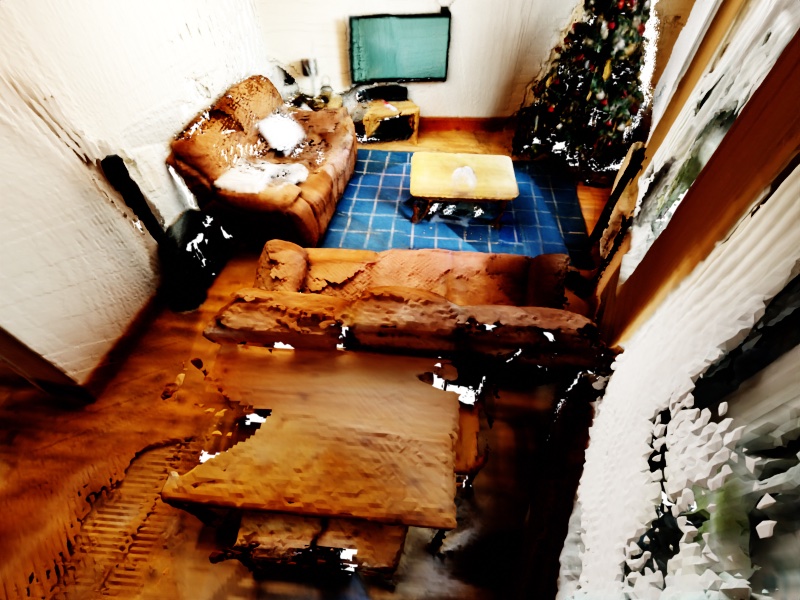}}%
    {\includegraphics[width=0.24\textwidth]{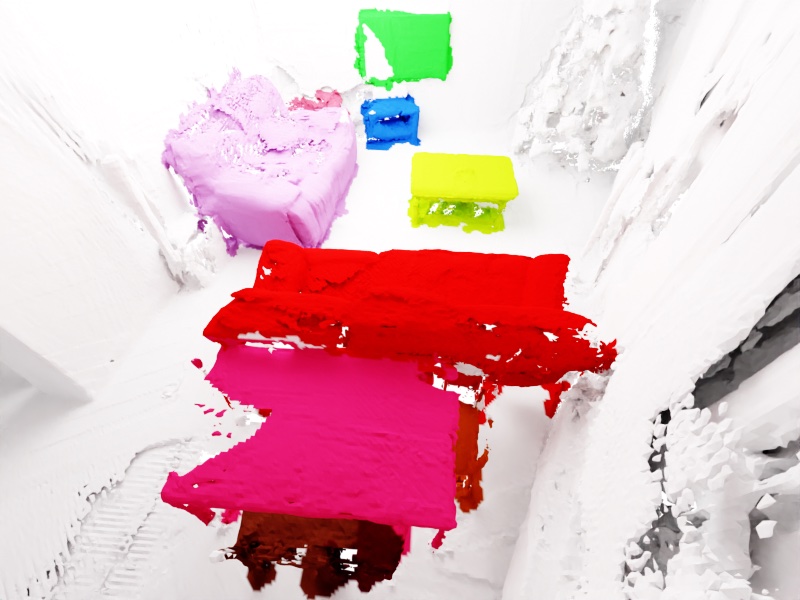}}%
    {\includegraphics[width=0.24\textwidth]{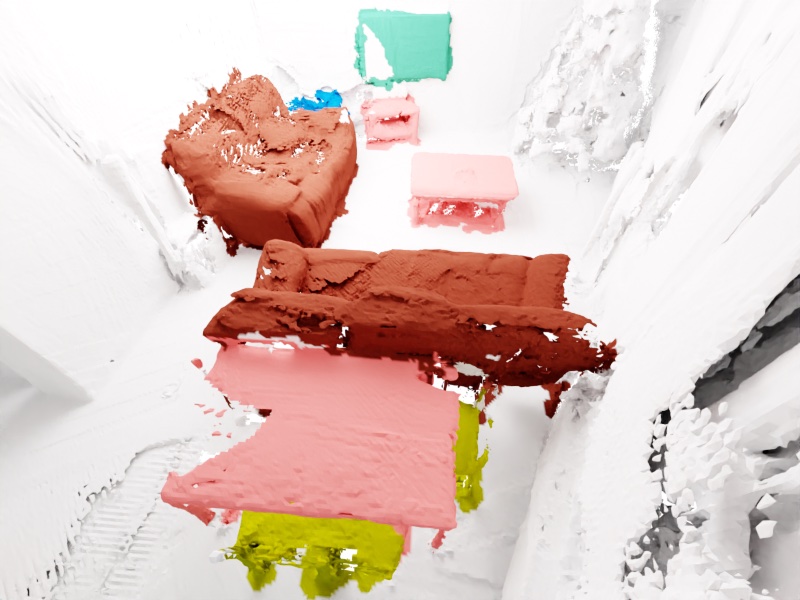}}%
    {\includegraphics[width=0.24\textwidth]{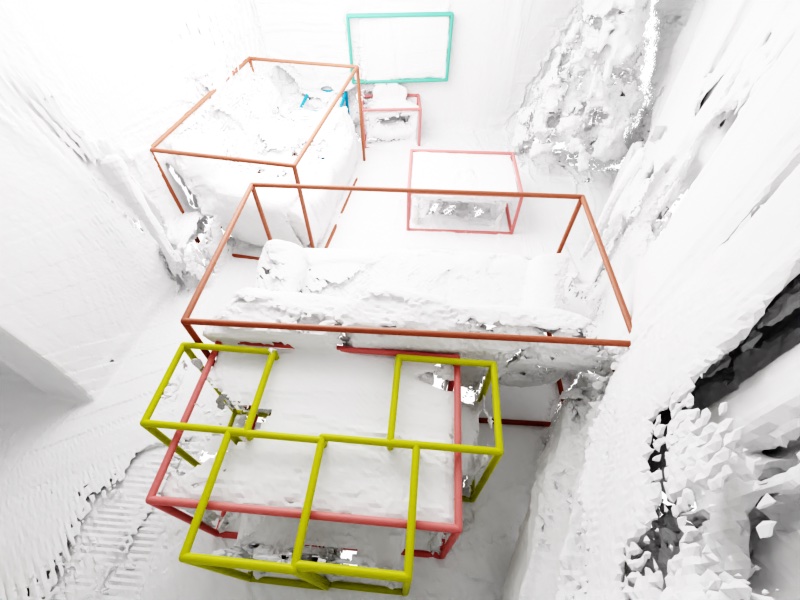}}%
\vspace{1px}
    {\includegraphics[width=0.24\textwidth]{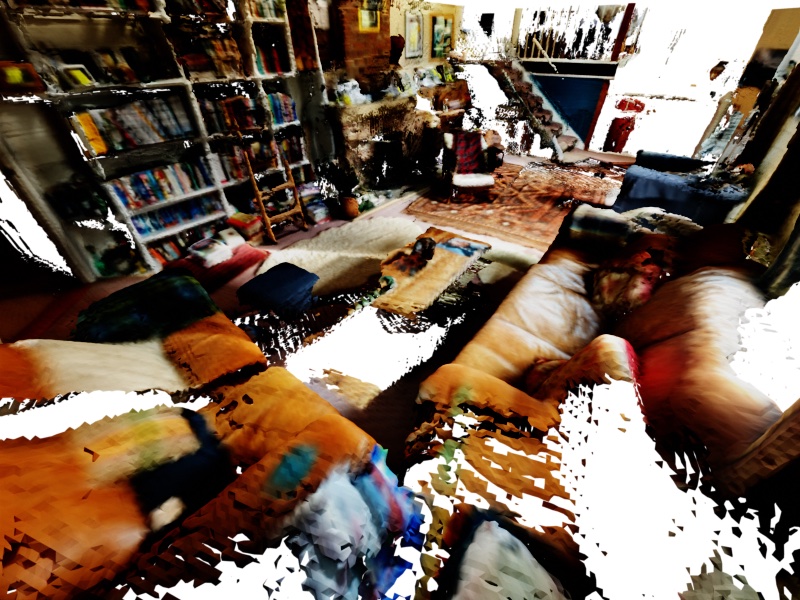}}%
    {\includegraphics[width=0.24\textwidth]{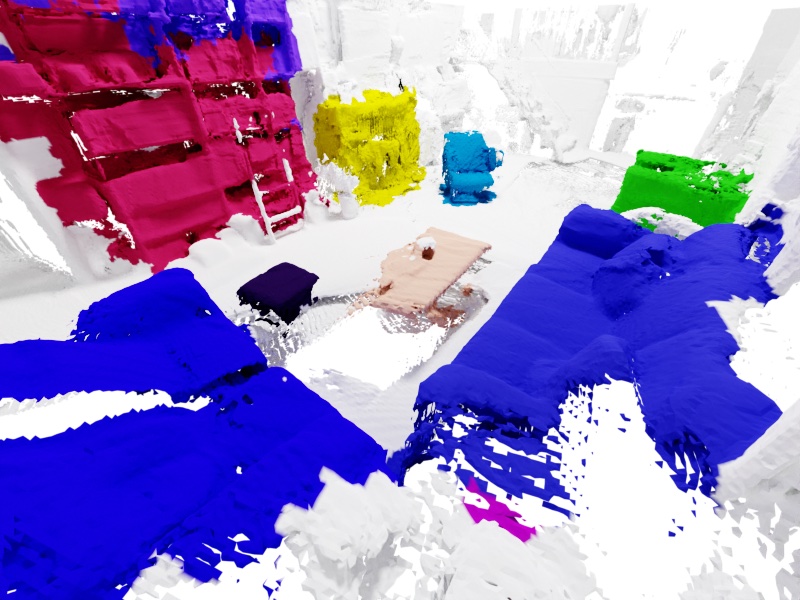}}%
    {\includegraphics[width=0.24\textwidth]{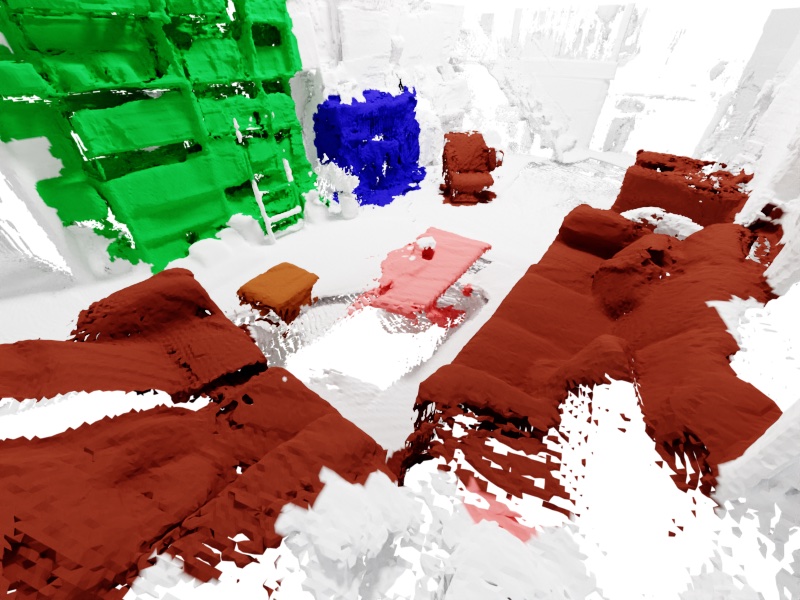}}%
    {\includegraphics[width=0.24\textwidth]{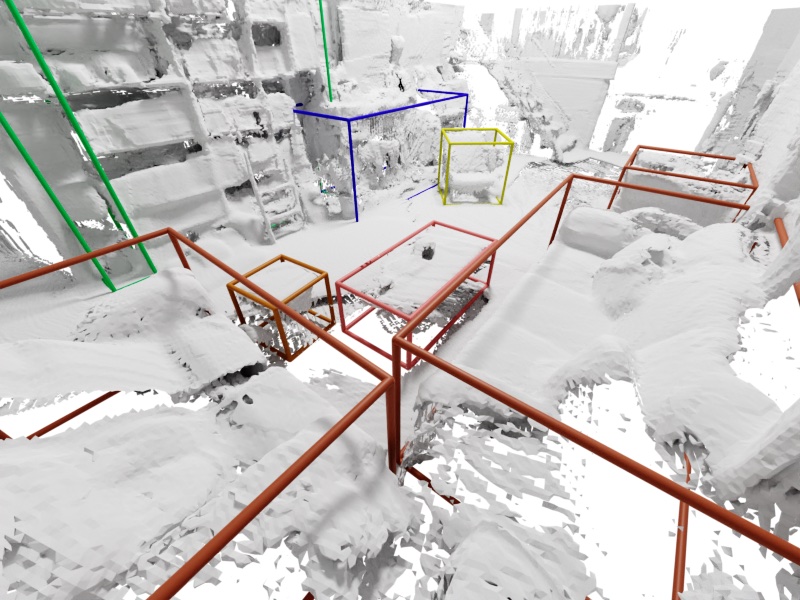}}%
\vspace{1px}
    {\includegraphics[width=0.24\textwidth]{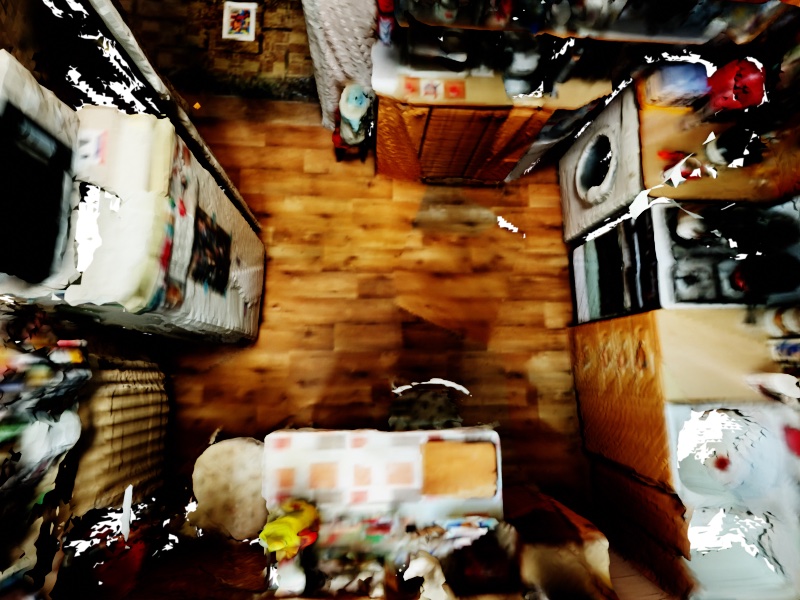}}%
    {\includegraphics[width=0.24\textwidth]{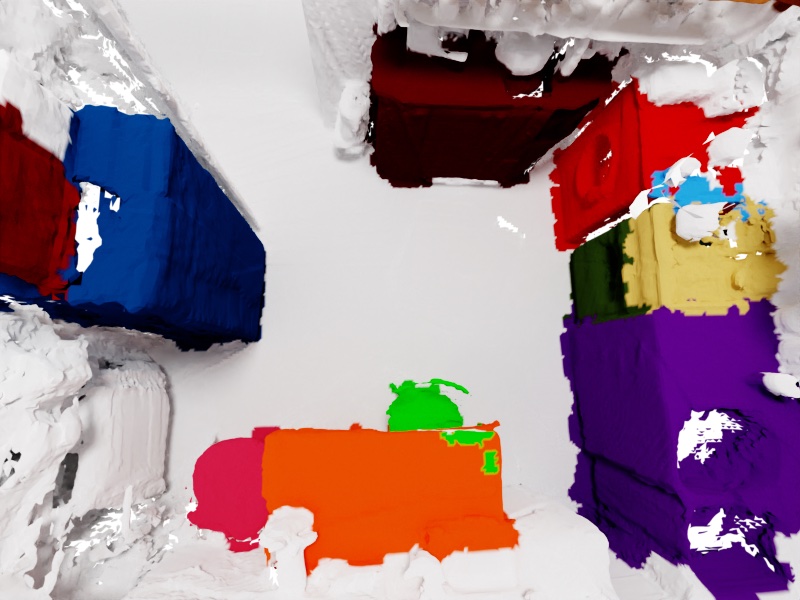}}%
    {\includegraphics[width=0.24\textwidth]{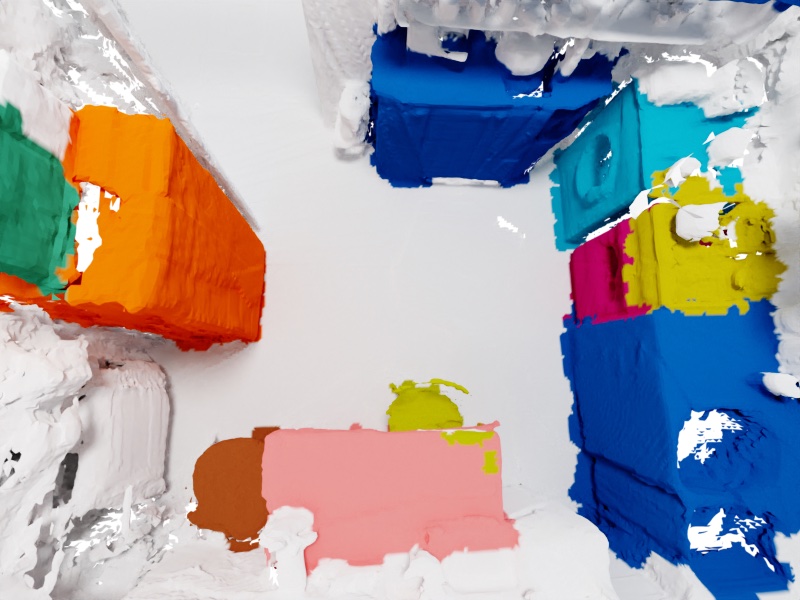}}%
    {\includegraphics[width=0.24\textwidth]{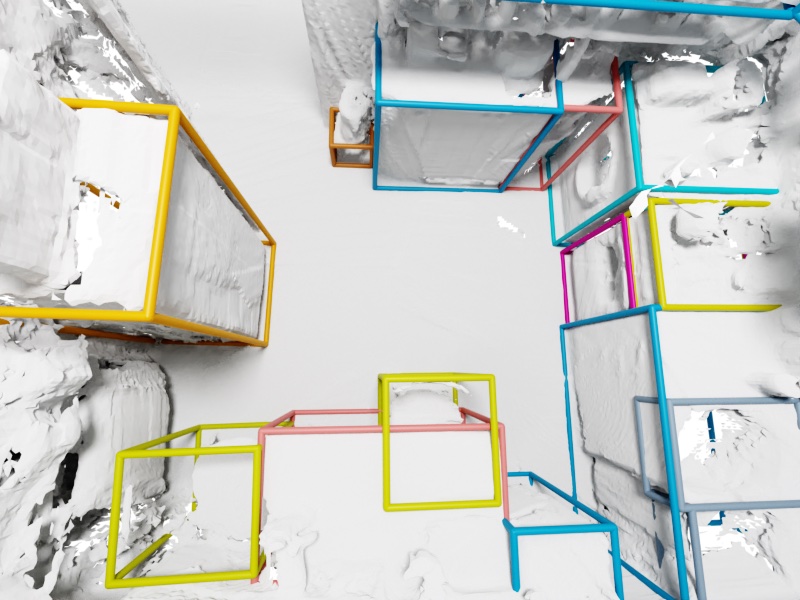}}%
\vspace{1px}
    {\includegraphics[width=0.24\textwidth]{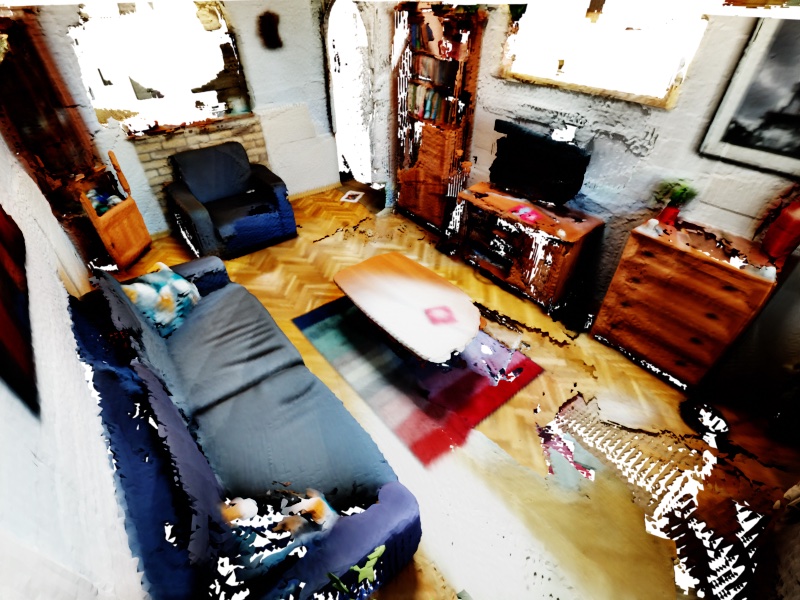}}%
    {\includegraphics[width=0.24\textwidth]{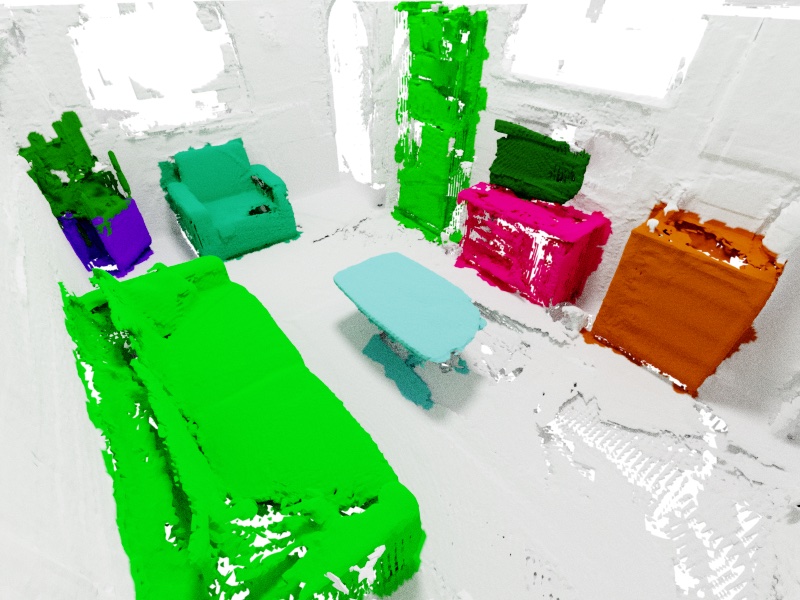}}%
    {\includegraphics[width=0.24\textwidth]{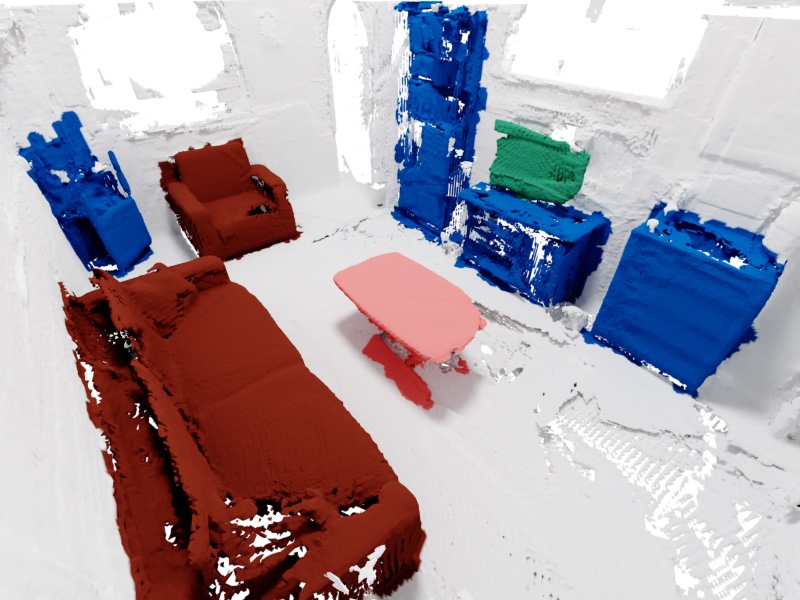}}%
    {\includegraphics[width=0.24\textwidth]{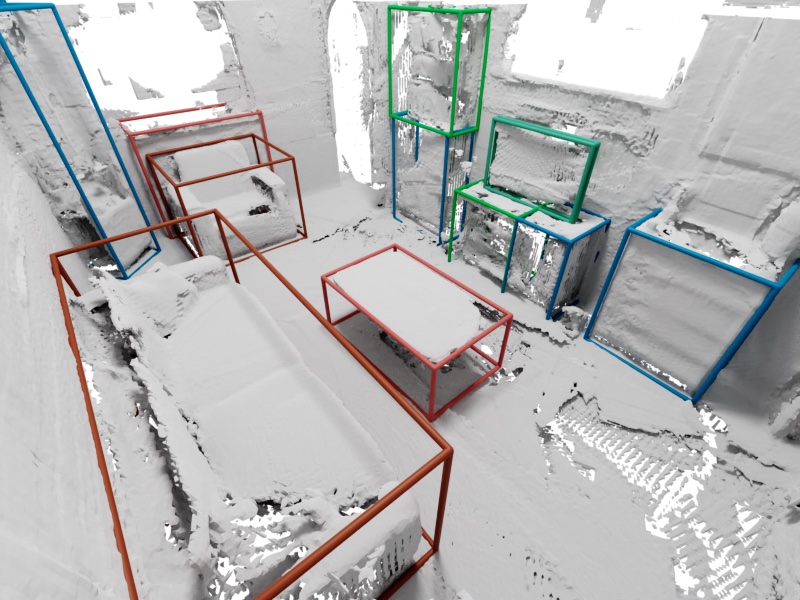}}%

\resizebox{\textwidth}{!}{
        \begin{tabular}{cccc}
         Input 3D Scene & Predicted Instance Masks & Predicted Semantic Classes & GT Bounding Boxes \\
         \hspace{95px} &
         \hspace{95px} &
         \hspace{95px} &
         \hspace{95px} \vspace{-2px}
 \\
         \multicolumn{4}{c}{
\textcolor{cabinet}{\ColorMapCircle}\,Cabinet \hspace{3px}
\textcolor{bed}{\ColorMapCircle}\,Bed \hspace{3px}
\textcolor{chair}{\ColorMapCircle}\,Chair \hspace{3px}
\textcolor{sofa}{\ColorMapCircle}\,Sofa \hspace{3px}
\textcolor{table}{\ColorMapCircle}\,Table \hspace{3px}
\textcolor{shelf}{\ColorMapCircle}\,Shelf \hspace{3px}
\textcolor{stove}{\ColorMapCircle}\,Stove \hspace{3px}
\textcolor{washer}{\ColorMapCircle}\,Washer \hspace{3px}
\textcolor{oven}{\ColorMapCircle}\,Oven \hspace{3px}
} \\
\multicolumn{4}{c}{
\textcolor{dishwasher}{\ColorMapCircle}\,Dishwasher \hspace{3px}
\textcolor{fireplace}{\ColorMapCircle}\,Fireplace \hspace{3px}
\textcolor{stool}{\ColorMapCircle}\,Stool \hspace{3px}
\textcolor{refrigerator}{\ColorMapCircle}\,Refrigerator \hspace{3px}
\textcolor{tv_monitor}{\ColorMapCircle}\,Monitor \hspace{3px}
\textcolor{toilet}{\ColorMapCircle}\,Toilet \hspace{3px}
\textcolor{sink}{\ColorMapCircle}\,Sink \hspace{3px}
\textcolor{bathtub}{\ColorMapCircle}\,Bathtub
         }
    \end{tabular}
    }
    \caption{
    \textbf{Qualitative Instance Segmentation Results on ARKitScenes\cite{Baruch21NIPS}.}
    Individual instance masks are colored randomly.
    Semantic classes are colored as indicated. 
    Ground truth boxes are shown for reference only and are not used during inference.}
    \label{fig:qualitative_arkitscenes}
\end{figure}

%% file: figures_tex/quali_scannet.tex
\begin{figure}[!hb]
    \centering
    
    \scriptsize

\resizebox{\textwidth}{!}{
        \begin{tabular}{cccc}
         Input 	& 	Fully-supervised & Box-Supervised & Dense Ground Truth \\
         3D Scene & 	Instance Segmentation & Instance Segmentation & (Not used in Box-Supervision) \\
         \hspace{95px} &
         \hspace{95px} &
         \hspace{95px} &
         \hspace{95px} \vspace{-2px}
    \end{tabular}
}

    {\includegraphics[width=0.24\textwidth, trim=290 250 180 120, clip]{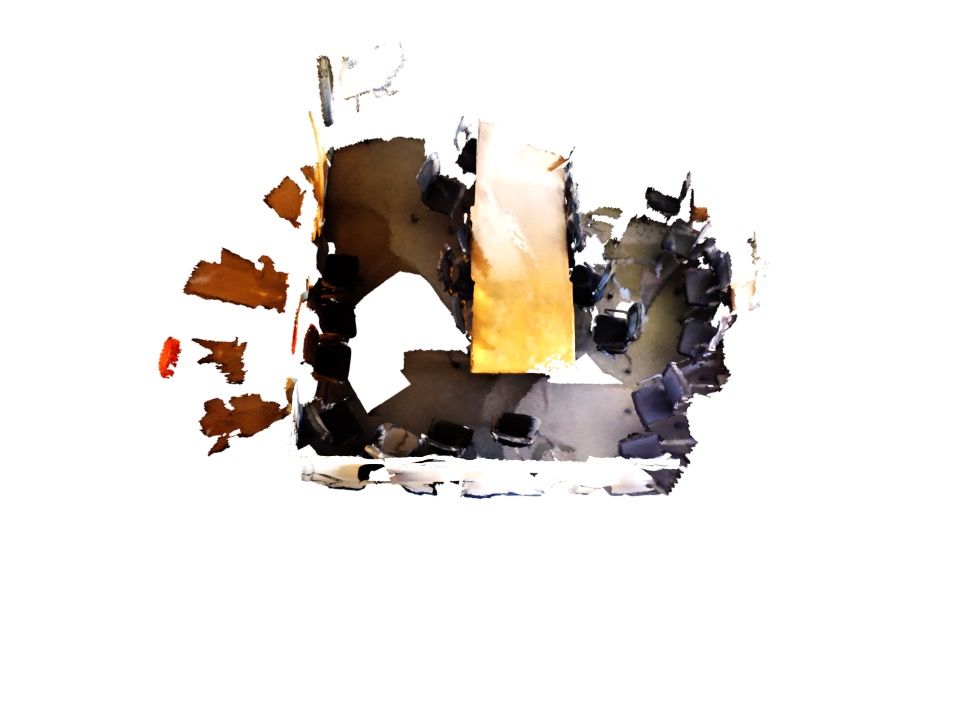}}%
    {\includegraphics[width=0.24\textwidth, trim=290 250 180 120, clip]{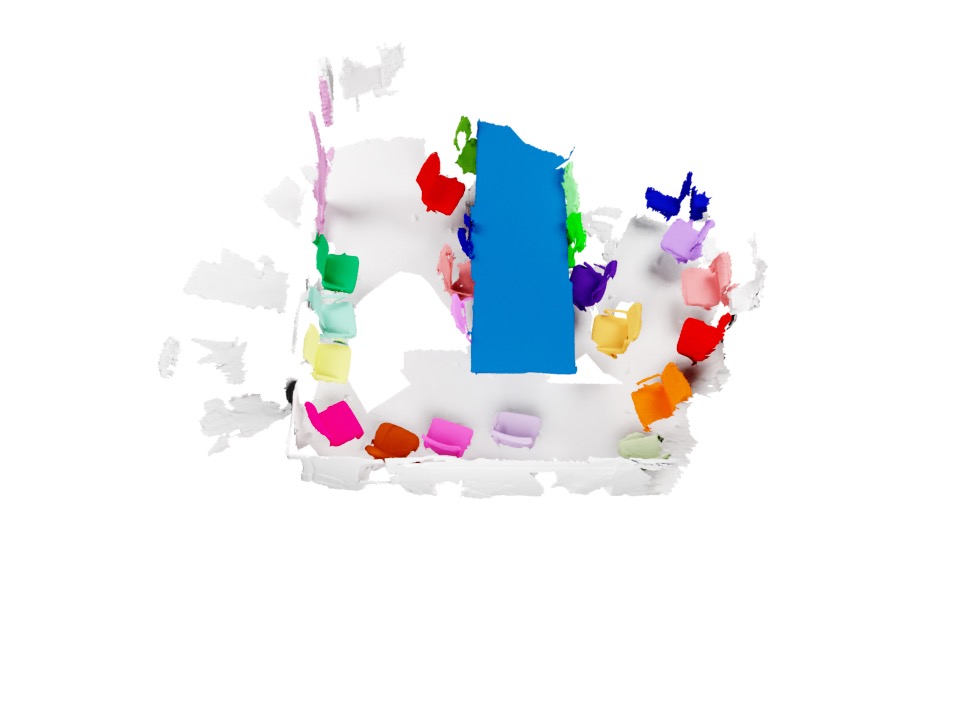}}%
    {\includegraphics[width=0.24\textwidth, trim=290 250 180 120, clip]{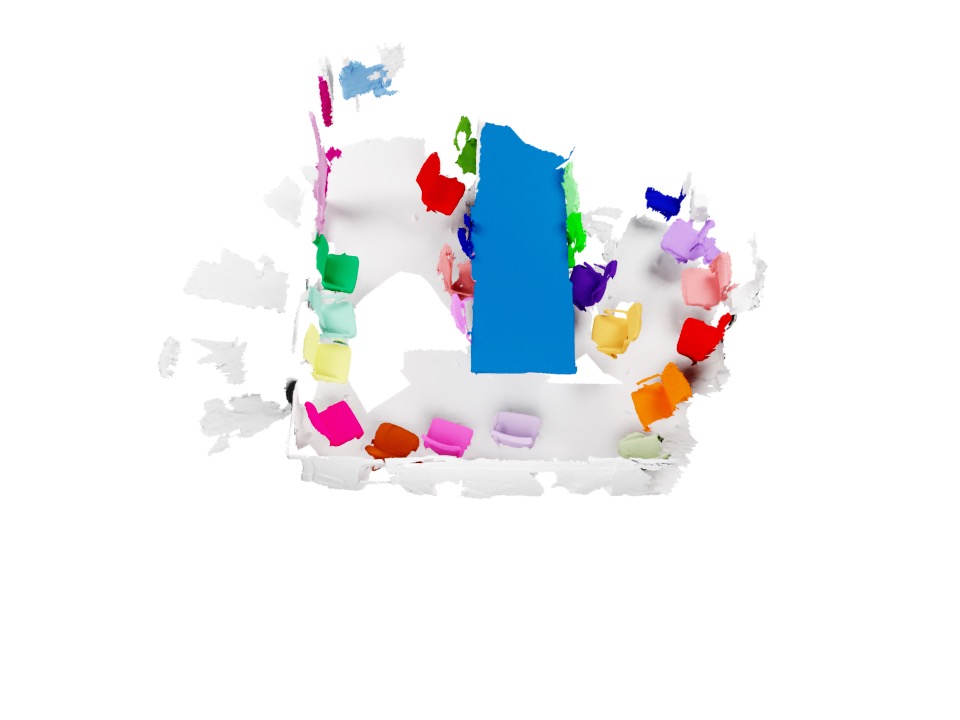}}%
    {\includegraphics[width=0.24\textwidth, trim=290 250 180 120, clip]{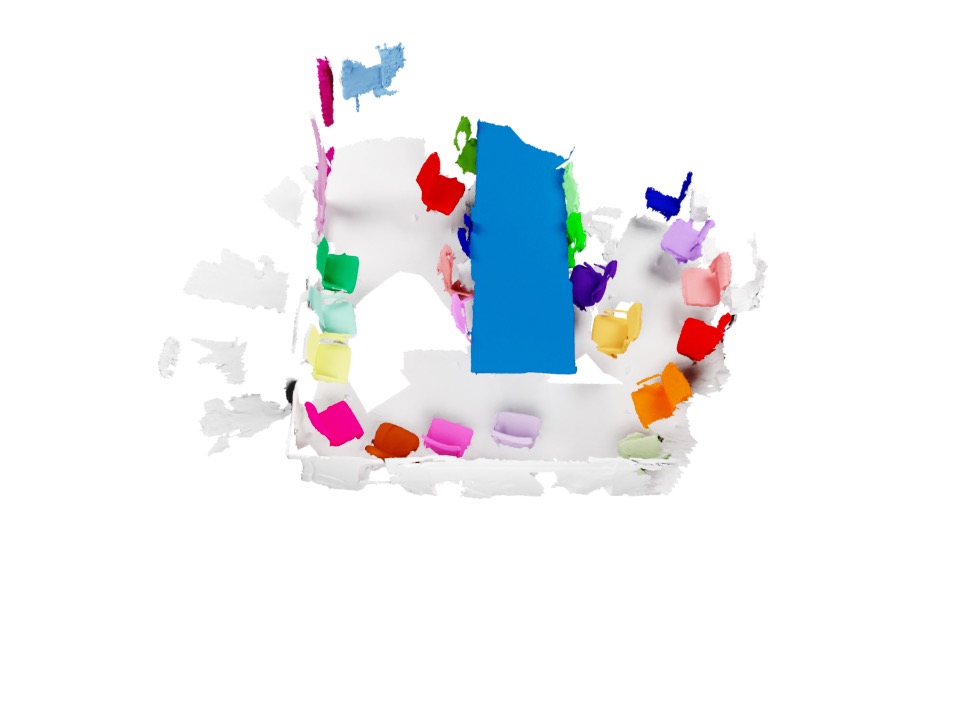}}%
\vspace{1px}
    {\includegraphics[width=0.24\textwidth, trim=150 230 180 80, clip]{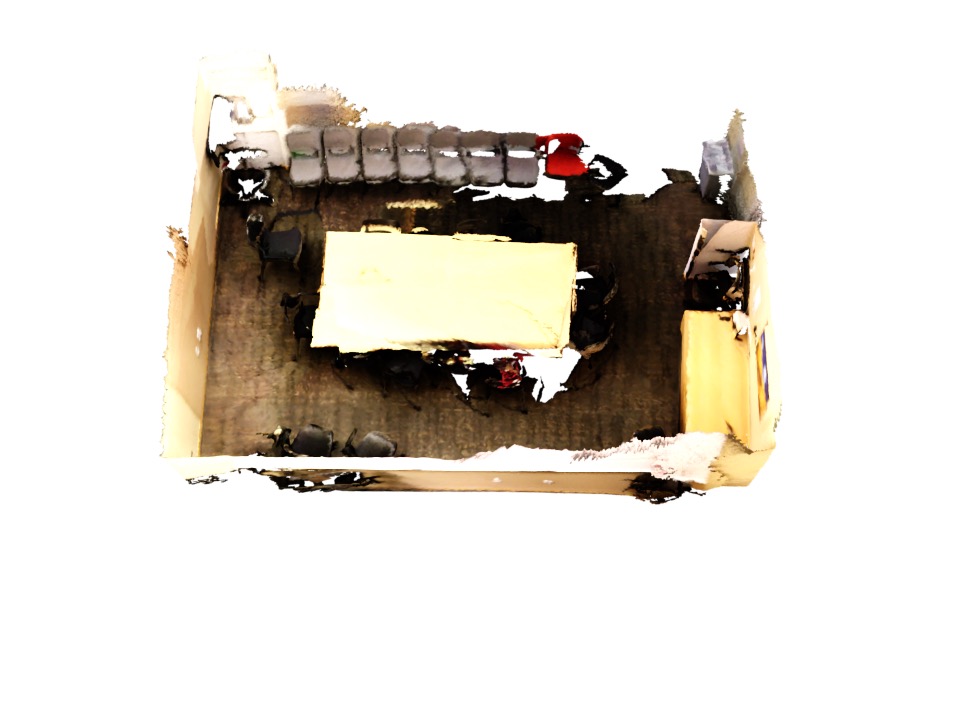}}%
    {\includegraphics[width=0.24\textwidth, trim=150 230 180 80, clip]{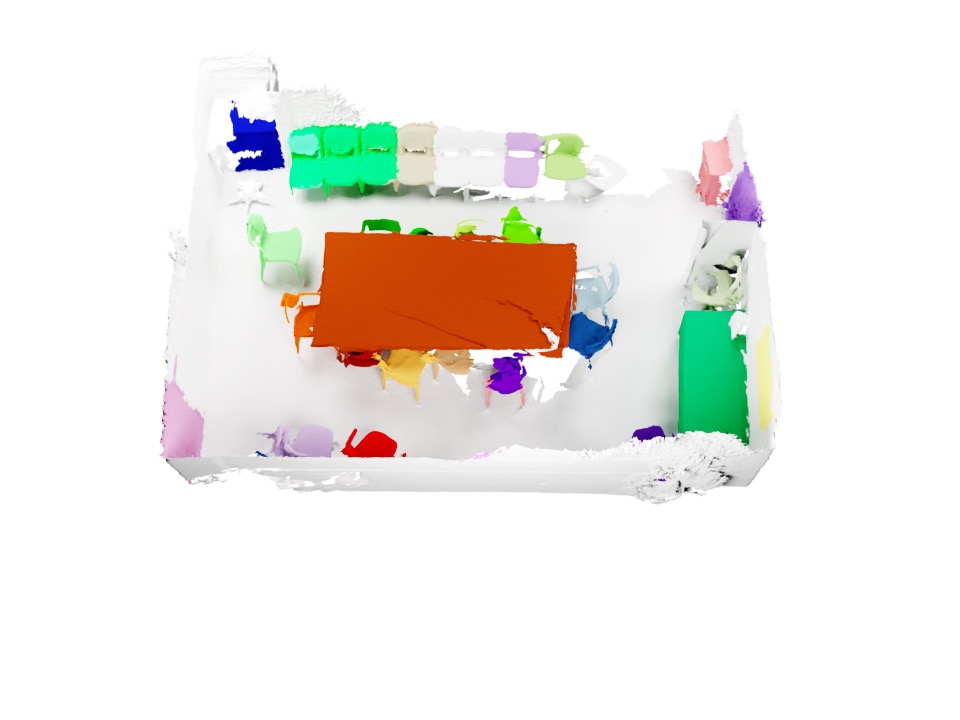}}%
    {\includegraphics[width=0.24\textwidth, trim=150 230 180 80, clip]{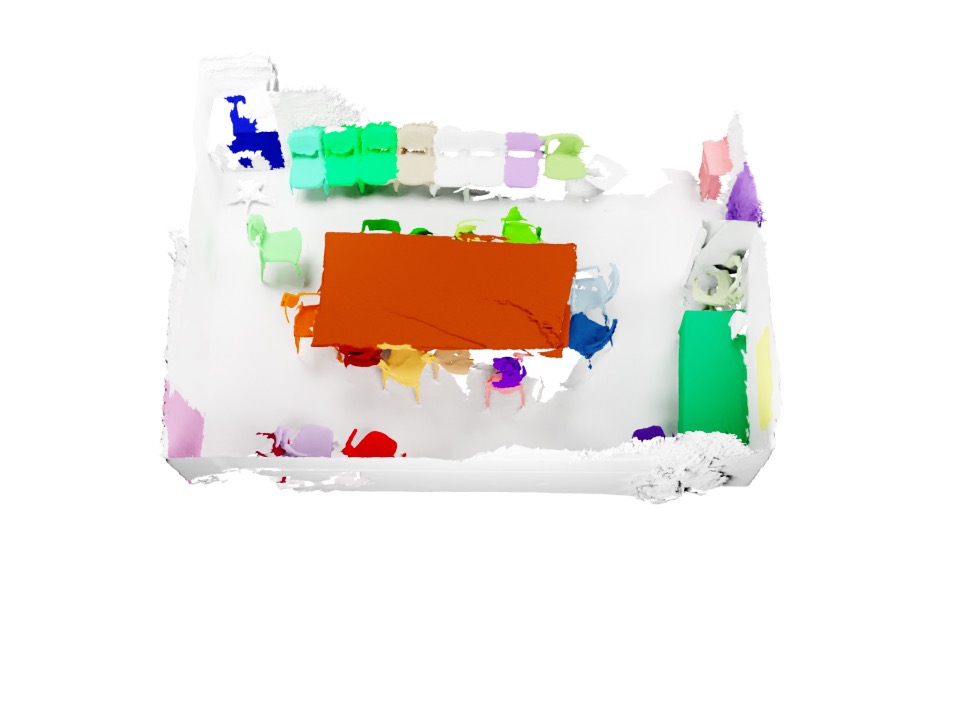}}%
    {\includegraphics[width=0.24\textwidth, trim=150 230 180 80, clip]{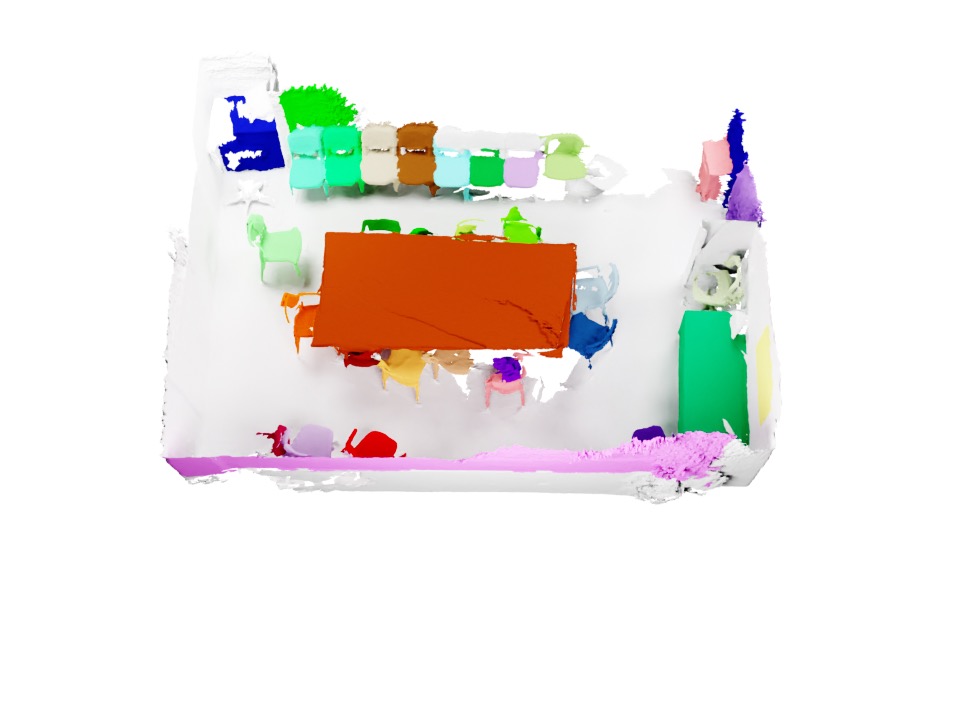}}%
\vspace{1px}

    {\includegraphics[width=0.24\textwidth, trim=160 70 140 0, clip]{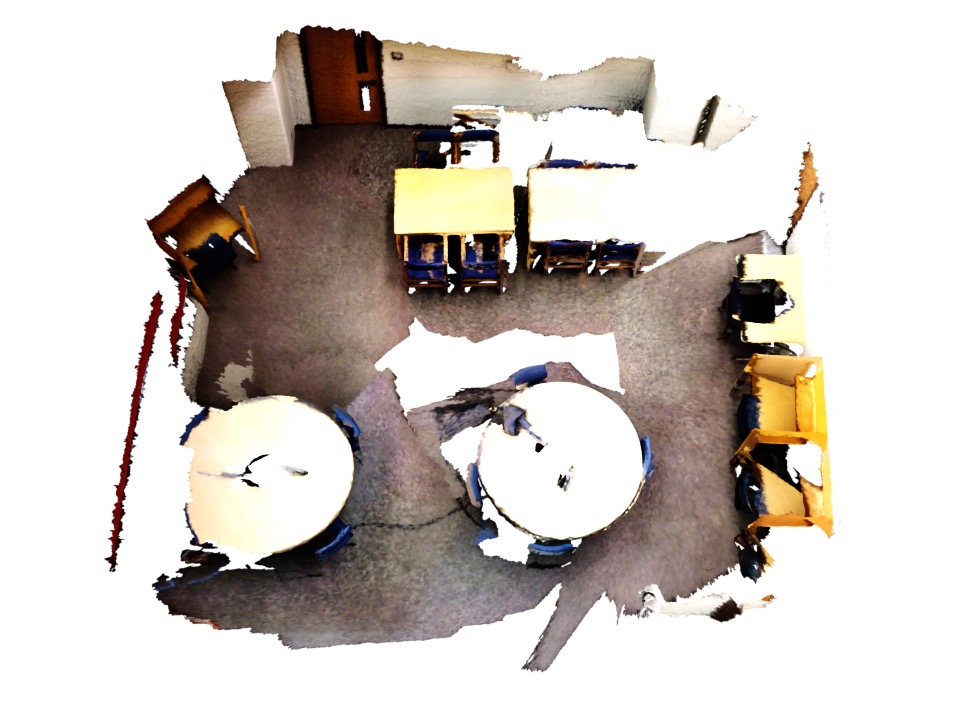}}%
    {\includegraphics[width=0.24\textwidth, trim=160 70 130 0, clip]{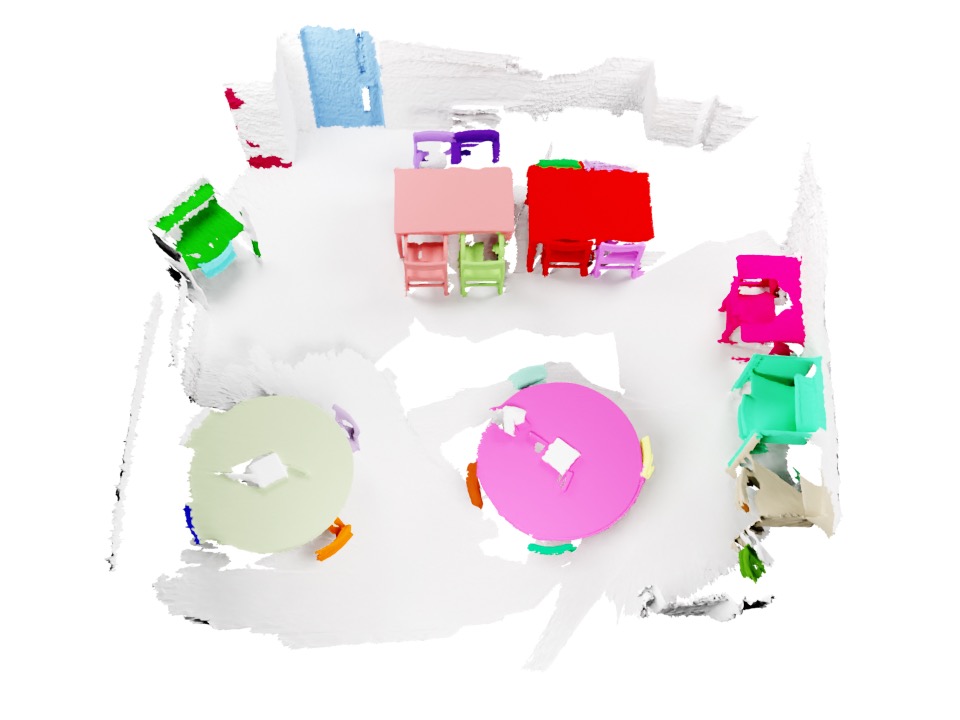}}%
    {\includegraphics[width=0.24\textwidth, trim=160 70 120 0, clip]{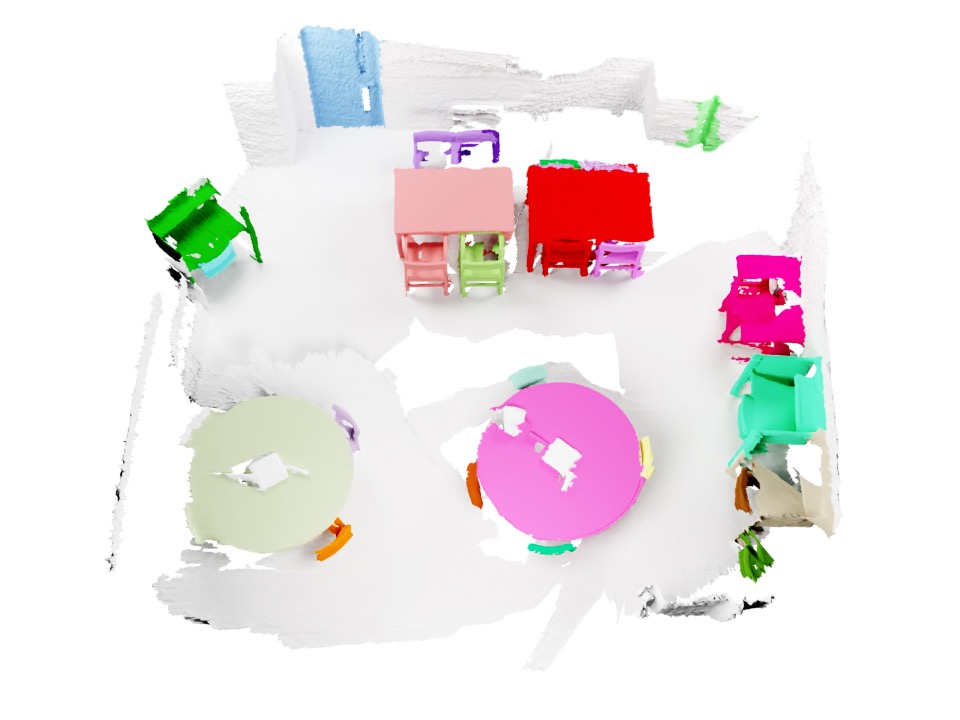}}%
    {\includegraphics[width=0.24\textwidth, trim=160 70 110 0, clip]{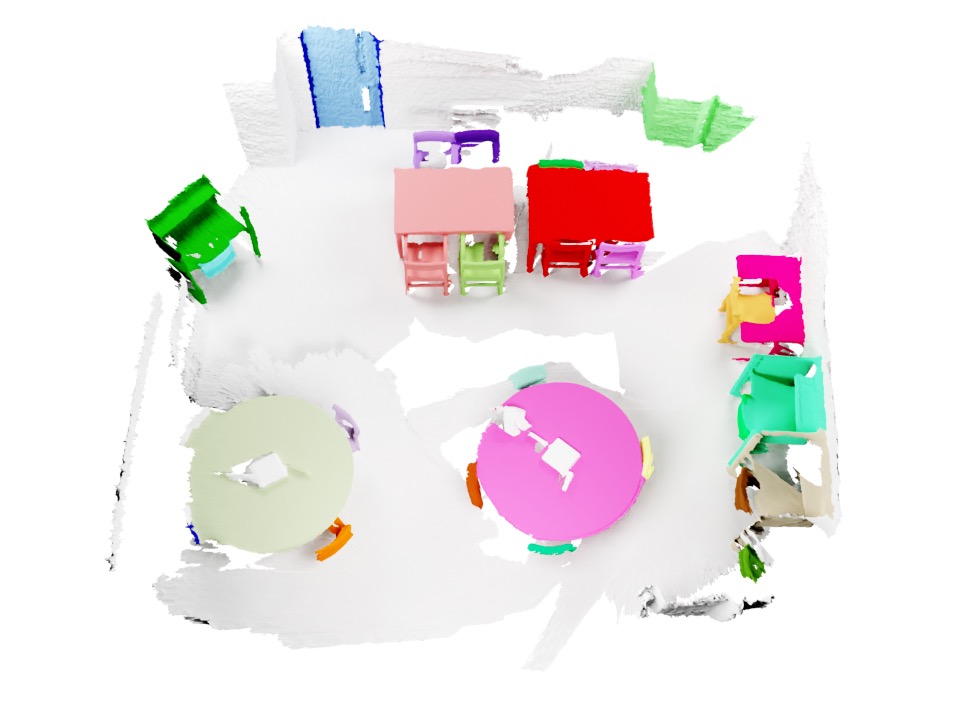}}%
\vspace{1px}
    {\includegraphics[width=0.24\textwidth, trim=170 180 50 0, clip]{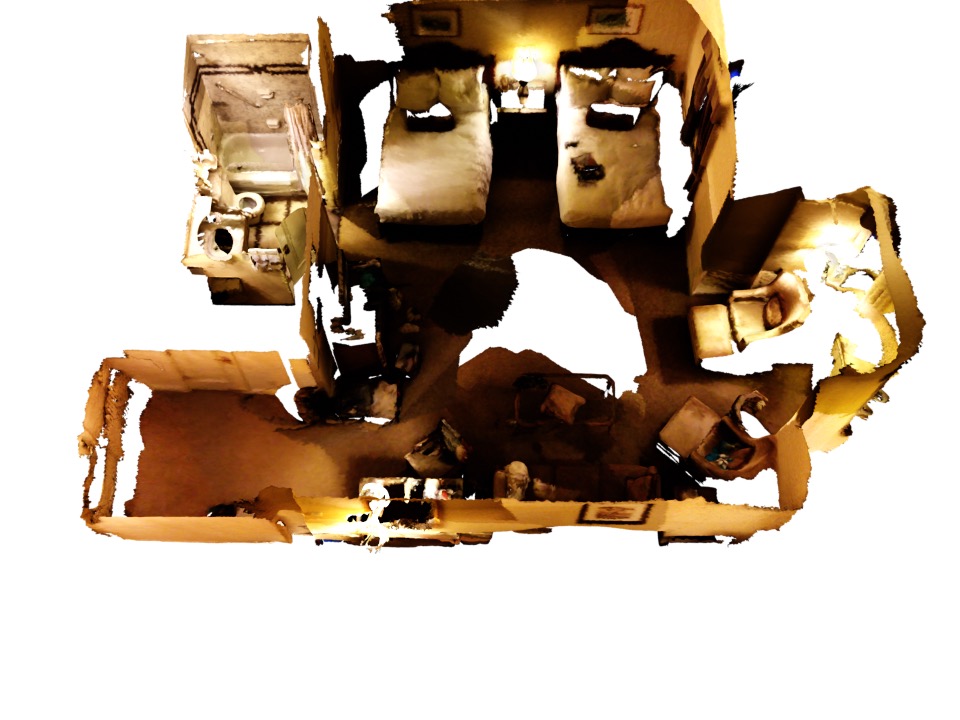}}%
    {\includegraphics[width=0.24\textwidth, trim=170 180 50 0, clip]{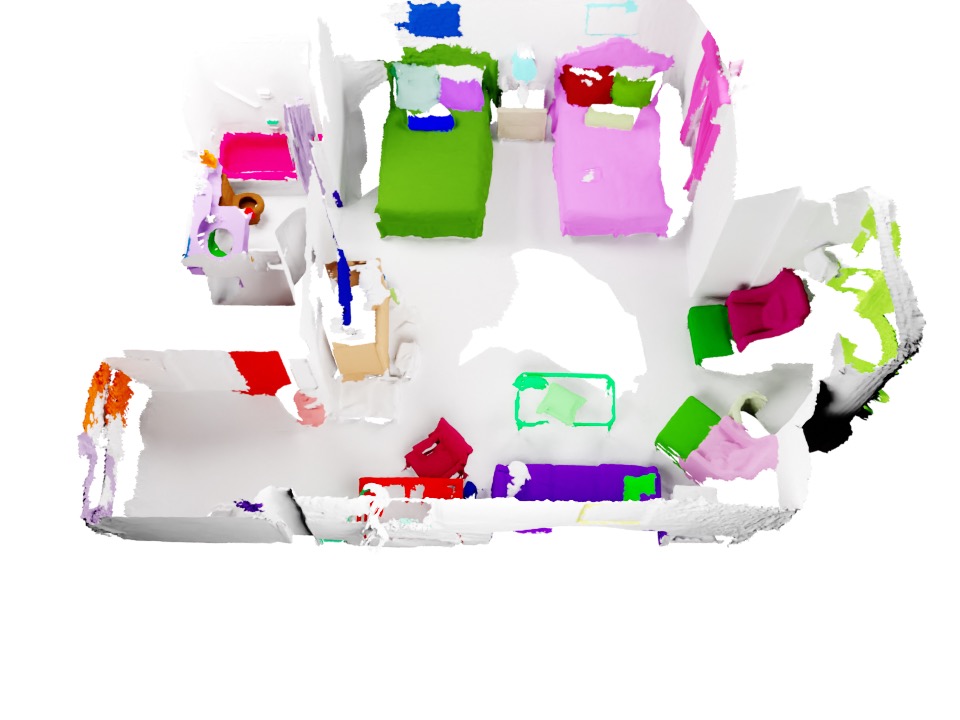}}%
    {\includegraphics[width=0.24\textwidth, trim=170 180 50 0, clip]{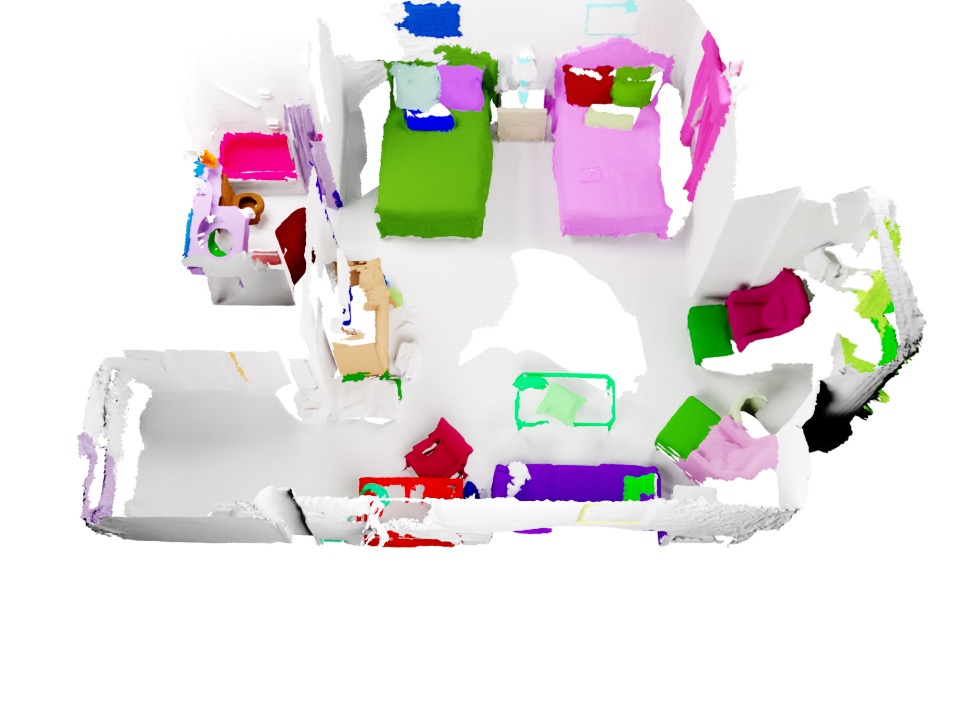}}%
    {\includegraphics[width=0.24\textwidth, trim=170 180 50 0, clip]{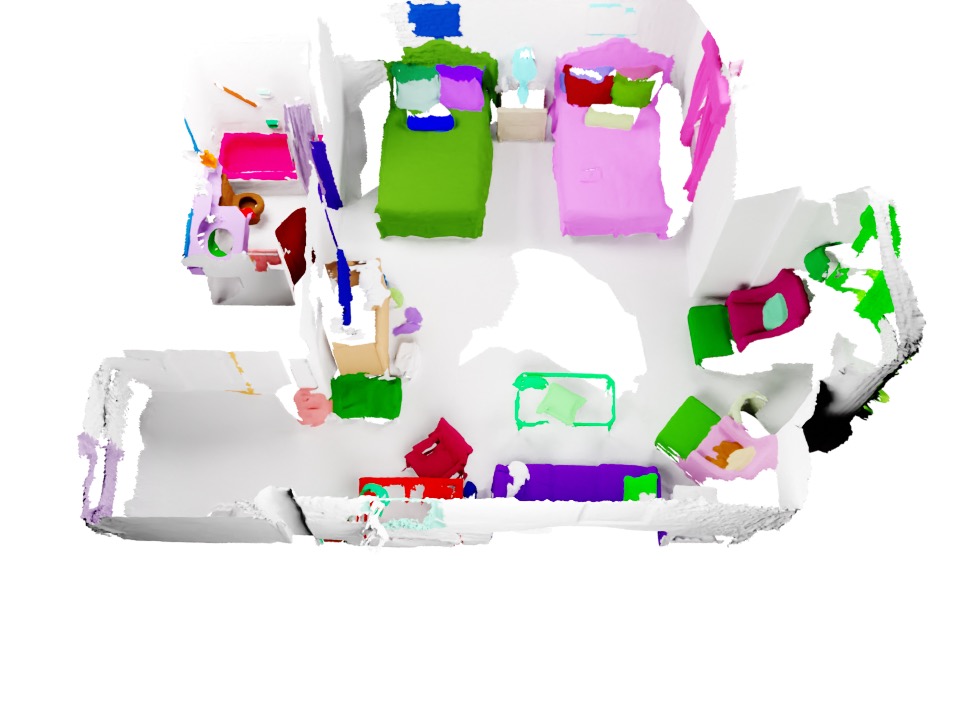}}%
    
    \vspace{1px}
    
    {\includegraphics[width=0.24\textwidth, trim=200 200 350 80, clip]{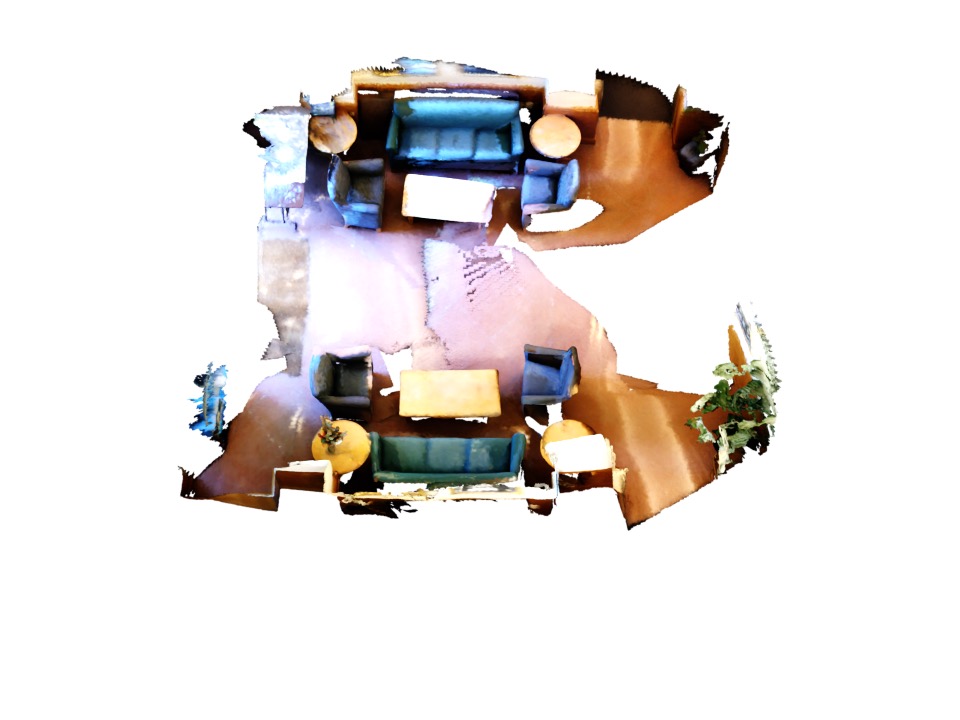}}%
    {\includegraphics[width=0.24\textwidth, trim=200 200 350 80, clip]{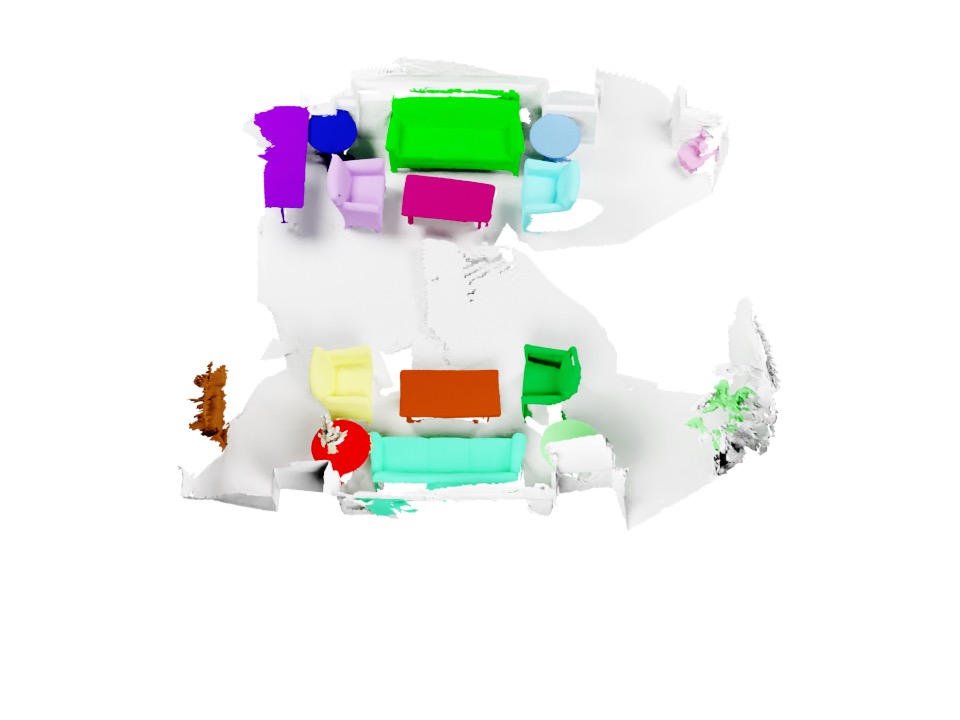}}%
    {\includegraphics[width=0.24\textwidth, trim=200 200 350 80, clip]{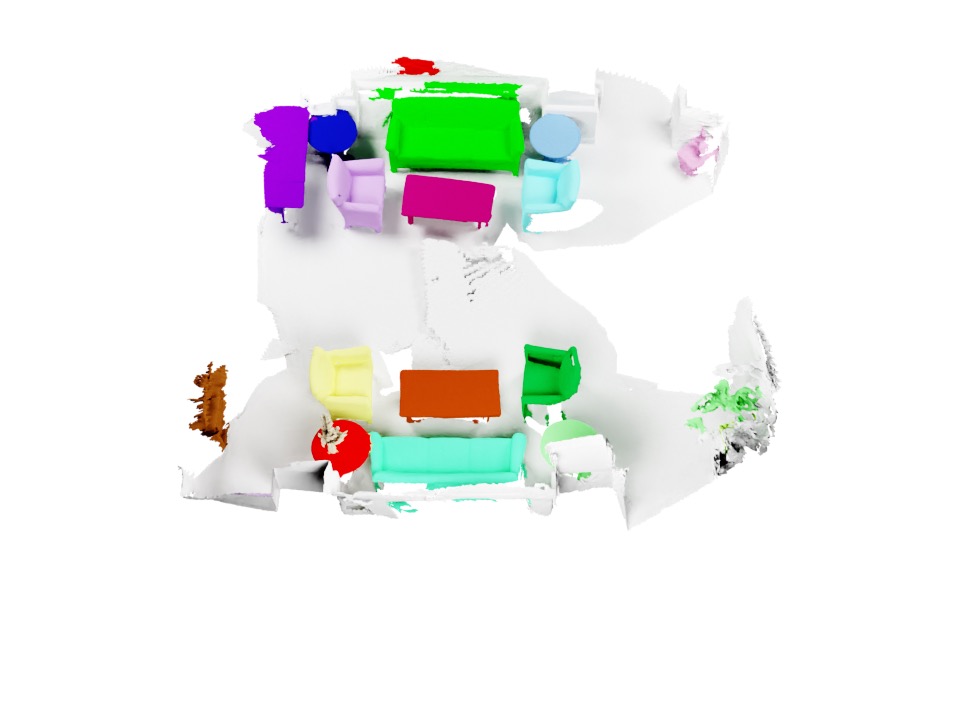}}%
    {\includegraphics[width=0.24\textwidth, trim=200 200 350 80, clip]{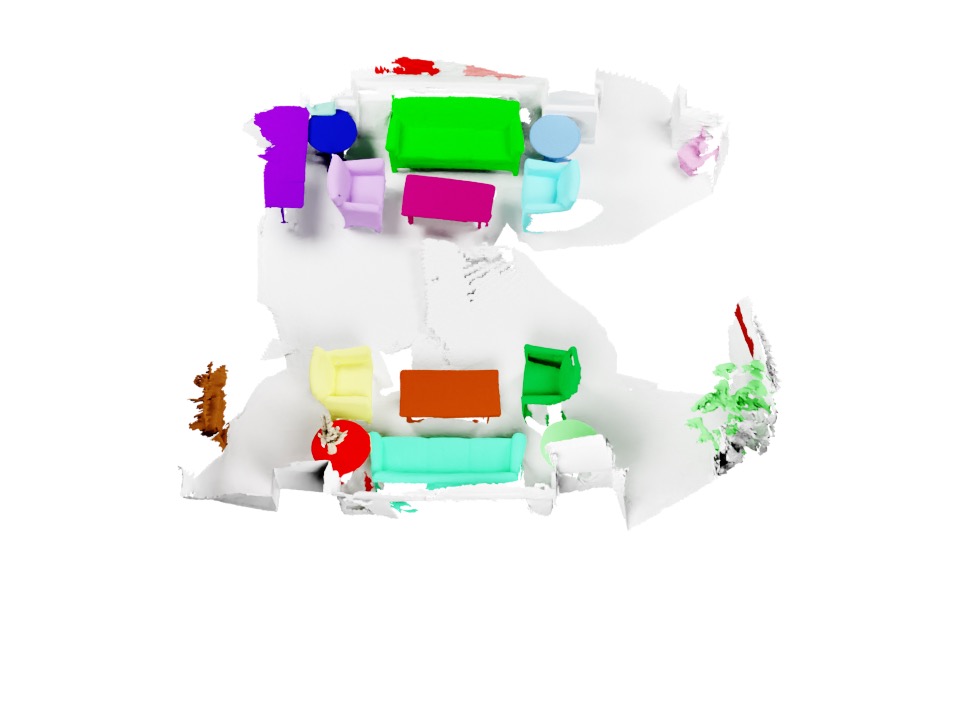}}
    \\
    \vspace{-5px}
    \caption{
    \textbf{Qualitative Instance Segmentation Results (validation) on  ScanNet\cite{Dai17CVPR}.}
    Results trained only on bounding boxes well resemble the fully supervised model, and both are close to densely annotated ground truth. 
    Instance masks have the same random color as the corresponding ground truth mask.
    }
    \label{fig:qualitative_scannet}
\end{figure}

%% file: tables/centers_vs_bbs.tex
\vspace{-45px}
\begin{table}[H]
\setlength{\tabcolsep}{3px}%
\begin{center}
\resizebox{\textwidth}{!}{
        \begin{tabular}{l | c c c}
        \toprule
               	     & mAP$_{50}$ &mAP$_{25}$   \\
        \midrule
            Centers + SC                &   51.4    &   63.8 \\
            Centers + SC (per Sem.) & 52.0 & 67.7 \\
            Boxes + SC (per Sem.) & 53.1 & 67.9 \\
            Boxes + NMC (per Sem.) & 55.1 & 68.3 \\
            Boxes + SC & 53.5 & 65.2 \\
            Boxes + NMC     & \textbf{59.7} &     \textbf{71.8} \\
        \bottomrule
    \end{tabular}
}
\end{center}
\vspace{-5px}
\caption{
{\footnotesize
Non-maximum clustering (NMC) and spatial clustering (SC) on center- and bounding box-votes.}
}
\label{tab:inst_rep}
\vspace{-35px}
\end{table}

%% file: tables/association_strategies.tex
\begin{table}[H]
\vspace{-43px}
\setlength{\tabcolsep}{3px}%
\begin{center}
\resizebox{\textwidth}{!}{
    \begin{tabular}{l | c c c}
    \toprule
            Supervision   	           & mAP$_{50}$     &mAP$_{25}$   \\
        \midrule
            (1) Decided Only                 & $56.0$ & 70.8\\
            (2) Decided + Closest Box	    & $58.7$ & 71.7\\
            (3) Decided + Smallest Box    & \textbf{59.7} & \textbf{71.8}\\
    \bottomrule
    \end{tabular}
}
\end{center}
\vspace{-5px}
\caption{
\footnotesize
{Analysis of association strategies}
}
\label{tab:association_strategies}
\vspace{-30px}
\end{table}

%% file: tables/fully_supervised_comparison.tex
\begin{table}[H]
\setlength{\tabcolsep}{1px}%
    \begin{center}
\scriptsize
\resizebox{1\textwidth}{!}{
        \begin{tabular}{l  ccccc}
        \midrule
    \multirow{3}{*}{Fully Supervised\hspace{3mm}} &
    \multicolumn{2}{c}{ScanNet} & &
    \multicolumn{2}{c}{S3DIS}\\
          \cmidrule(r){2-3} \cmidrule(r){5-6}
            & mAP$_{50}$ & mAP$_{25}$  & \hspace{5px} & mPrec & mRec \\
         \midrule
         PointGroup\cite{Jiang20CVPR} & 56.9 & 71.3 & & 69.6 & 69.2\\
         3D-MPA\cite{Engelmann20CVPR} & 59.1 & 72.4 & & 66.7 & 64.1\\
         OccuSeg\cite{Han20CVPR}      & -- & -- & & 72.8 & 60.3\\
         HAIS\cite{Chen21ICCV}        & 64.1 & \textbf{75.6} & & 73.2 & 69.4\\
         SSTNet\cite{Liang21CVPR}     & 64.3 & 74.0 & & 73.5 & \textbf{73.4}\\
         \name{} (Ours) & \textbf{64.7} & 74.5 & & \textbf{75.4} & 69.3\\
         \midrule
    \end{tabular}
}
\end{center}
\vspace{-9px}
\caption{
\small
\textbf{Fully Supervised Setting.}
Our model achieves competitive instance segmentation scores on ScanNet validation and S3DIS 6-fold cross validation.}
\label{tab:fully_supervised_comparison}
\end{table}

%% file: sections/5_conclusion.tex
%
\vspace{-9px}
\section{Conclusion}
\vspace{-7px}
In this work, we show that 3D bounding box annotations serve surprisingly well as weak supervision for training dense instance segmentation models. 
Prior works either use dense supervision on all points (which is costly to label), or weak supervision from only a few annotated points (which performs less well).
Bounding boxes provide an attractive alternative: the annotation effort is drastically reduced compared to dense point labeling, and they perform notably better than prior sparse labels and are even close to fully-supervised methods.
We demonstrate the effectiveness of our instance segmentation approach on several benchmarks, and in particular on the recent, largest scene dataset, ARKitScenes. 
Although annotated with 3D bounding boxes only, we obtain for the first time compelling 3D instance segmentation results. 
This unlocks a large body of 3D detection datasets to be viable for learning instance segmentation.

%% file: sections/6_acknowledgements.tex
\vspace{-4px}
\footnotesize \parag{Acknowledgements.}
 We thank Alexey Nekrasov and Jonas Schult for their feedback.
This work is funded by the Carl Zeiss Foundation.
This work is also funded by the Deutsche Forschungsgemeinschaft (DFG, German Research Foundation) - 409792180 (Emmy Noether Programme, project: Real Virtual Humans) and  the German Federal Ministry of Education and Research (BMBF): Tübingen AI Center, FKZ: 01IS18039A.
 G. Pons-Moll is a member of the Machine Learning Cluster of Excellence, EXC number 2064/1, Project number 390727645.
 J. Chibane is a fellow of the Meta Research PhD Fellowship Program - area: AR/VR Human Understanding.
 F. Engelmann is a post-doctoral research fellow at the ETH Zürich AI Center.
 

%% file: sections/7_supplementary.tex
\section{Baseline: Object Detector followed by Segmentation}
\label{sec:detection_baseline}
\begin{figure}[h]
\centering
\footnotesize
\resizebox{\textwidth}{!}{
        \begin{tabular}{cc}
         \textbf{Baseline Method (Detector+Segmentation)} & \textbf{Our Weakly-Supervised \name{} }\\
         \hspace{230px} &
         \hspace{230px} \vspace{-10px}
    \end{tabular}
    }
{\includegraphics[width=0.48\textwidth]{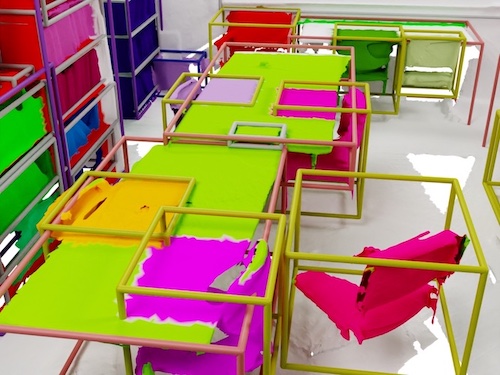}}%
{\includegraphics[width=0.48\textwidth]{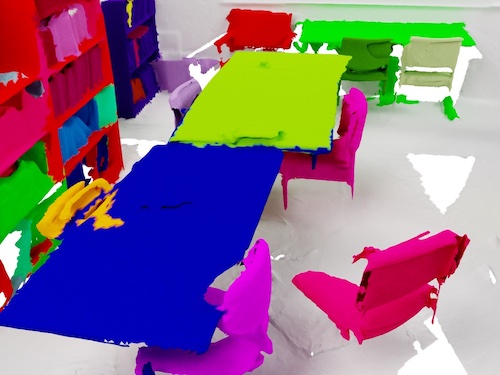}}%
\\
\caption{Qualitative comparison of the baseline \emph{(left)} and our approach \emph{(right)}.
For the baseline, the outputs of the object detector and the subsequent foreground background segmentations are shown.
The baseline fails whenever two object bounding boxes are intersecting (tabletop).
While our \name{} is supervised with comparable labels during training,
it learns to generalize beyond these weak labels and infers the correct instance masks for objects with intersecting bounding boxes (see chairs and table).}
\label{fig:baseline}
\end{figure}

In the main paper, we address how per-point instance masks can be learned from bounding box annotations only.
To show that this is a non-trivial task, and that our proposed method generalizes beyond the weak supervision signal, we present an additional baseline experiment.
This baseline is an object detector predicting bounding boxes and is trained on the given ground truth bounding box annotations.
Then, the instance masks are obtained by segmenting the points inside each predicted bounding box into foreground and background.
The baseline implementation closely follows the implementation of our main model: using a sparse convolutional network\cite{Choy19CVPR} we obtain deep learned features for each point in the input point cloud.
The learned point features then vote for object bounding box proposals.
These steps are identical to the first part of our main model shown in Fig.\,\ref{fig:annotationtypes} of the main paper.
We then perform non-maximum-suppression (NMS) to obtain object detection bounding boxes from the proposals.
The final instance masks are obtained from the predicted bounding boxes, which are segmented into foreground and background based on the number of bounding boxes each point is contained in.
This is the same mechanism as described in the main paper to obtain per-point supervision signals (Sec.~\ref{sec:bb_associations}, Eq.~\ref{eq:associate_func} in the main paper).
By doing so, it is guaranteed that the baseline is directly comparable with the proposed weakly-supervised approach.
Visual results, including the object detections, are shown in \reffig{baseline}.
Scores are shown in \reftab{baseline}.
Our proposed approach largely outperforms this baseline (+11.8\,mAP$_{50}$).
In particular, this experiment shows that learning instance masks from bounding box annotations alone is non-trivial, and that our trained model is able to generalize beyond the weak training signal obtained from the bounding box annotations. 

\definecolor{my_green}{rgb}{0.0, 0.5, 0.0}
\begin{table}[]
\setlength{\tabcolsep}{10pt}
\begin{center}
\begin{tabular}{l | l l l}
\toprule
                    & mAP   & mAP$_{50}$  & mAP$_{25}$\\
\midrule
Baseline (ours)    & 26.5  & 47.9      & 64.8 \\
\name{} (ours) & \textbf{39.1} {\color{my_green}\scriptsize (+12.6)} &  \textbf{59.7} {\color{my_green}\scriptsize (+11.8)} &    \textbf{71.8} {\color{my_green}\scriptsize (+7.0)}   \\
\bottomrule
\end{tabular}
\end{center}
\caption{Comparison of our approach to the baseline (object detector followed by segmentation) on ScanNet validation set, trained with bounding box supervision only.
The results indicate that obtaining instance masks from bounding boxes is non-trivial and that our training technique efficiently leverages weak bounding box annotations to predict dense and accurate instance masks.
This is further visualized in \reffig{baseline}.}
\label{tab:baseline}
\end{table}

\section{Per-Category Results}
In this section, we show per-category results on the ScanNet validation and test splits, and on S3DIS 6-fold cross validation, as summarised in Tab.~\ref{tab:scannet_ins1},~\ref{tab:scannet_ins2},~\ref{tab:s3dis_ins1} and ~\ref{tab:s3dis_ins2}.
On ScanNet validation and S3DIS, we show also per-category scores for the fully-supervised model trained with per-point instance labels.

\begin{table}[H]
\begin{center}
\setlength{\tabcolsep}{4pt}
     \resizebox{\textwidth}{!}{
     \begin{tabular}{l|cccccccccccccccccc|c}
     \toprule
                & cab  & bed  & chair& sofa & tabl & door & wind & bkshf& pic  & cntr & desk & curt & fridg& showr & toil & sink & bath & ofurn & \textbf{avg.}\\
\midrule
Ours {\scriptsize (mAP)}      & 28.3 & 46.3 & 62.5 & 71.1 & 30.9 & 26.0 & 27.2 & 43.3 & 32.2 & 10.3 & 12.5 & 29.8 & 47.2 & 70.1 & 88.2 & 36.3 & 74.1 & 42.4 & 43.3\\
Ours {\scriptsize (mAP@50\%)} & 50.9 & 84.7 & 81.6 & 85.2 & 57.8 & 56.2 & 48.8 & 77.1 & 44.8 & 27.7 & 48.2 & 55.8 & 79.0 & 100 & 99.7 & 66.6 & 100 & 64.0 & 67.7\\
Ours {\scriptsize (mAP@25\%)} & 70.7 & 96.2 & 88.7 & 90.2 & 75.3 & 71.5 & 63.7 & 87.4 & 46.9 & 68.6 & 96.1 & 59.8 & 70.0 & 100 & 99.7 & 91.2 & 100 & 69.4 & 80.3\\
     \bottomrule
    \end{tabular}
    }
\end{center}
\vspace{-5px}
    \caption{\textbf{Instance Segmentation on ScanNetV2~\cite{Dai17CVPR} Test Set}.
    Trained only on \emph{bounding boxes} on training and validation splits, no per-point annotations used.}
    \label{tab:scannet_ins1}
\vspace{-40px}
\end{table}

\begin{table}[H]
\begin{center}
\setlength{\tabcolsep}{4pt}
     \resizebox{\textwidth}{!}{
     \begin{tabular}{l|cccccccccccccccccc|c}
     \toprule
                & cab  & bed  & chair& sofa & tabl & door & wind & bkshf& pic  & cntr & desk & curt & fridg& showr & toil & sink & bath & ofurn & \textbf{avg.}\\
\midrule
Ours {\scriptsize (mAP)}      & 27.6 & 40.2 & 74.0 & 52.5 & 33.2 & 25.9 & 24.2 & 25.9 & 27.8 & 8.4 & 16.9 & 34.5 & 32.9 & 42.9 & 80.5 & 42.9 & 70.0 & 43.6 & 39.1 \\
Ours {\scriptsize (mAP@50\%)} & 48.0 & 72.0 & 91.8 & 77.5 & 62.9 & 48.6 & 43.3 & 49.9 & 40.9 & 27.9 & 44.3 & 51.8 & 43.4 & 56.8 & 96.9 & 72.7 & 87.1 & 59.6 & 59.7 \\
Ours {\scriptsize (mAP@25\%)} & 59.5 & 83.8 & 94.5 & 87.0 & 75.5 & 59.8 & 61.4 & 68.2 & 45.6 & 58.5 & 78.6 & 65.1 & 46.9 & 77.4 & 96.9 & 79.5 & 87.1 & 67.1 & 71.8\\
     \bottomrule
    \end{tabular}
    }
\vspace{-5px}\end{center}
    \caption{\textbf{Instance Segmentation on ScanNetV2~\cite{Dai17CVPR} Validation Set}.
    Trained only on \emph{bounding boxes} on the training split, no per-point annotations used during training.}
    \label{tab:scannet_ins2}
\vspace{-40px}
\end{table}
\begin{table}[H]
\begin{center}
\setlength{\tabcolsep}{4pt}
\resizebox{\textwidth}{!}{
\begin{tabular}{l|ccccccccccccc|c}
\toprule
                & ceiling & floor & wall & beam & column & window & door & table & chair & sofa & bookshelf & board & clutter & \textbf{avg.}\\
\midrule
Ours {\scriptsize (mPrec)} & 97.1 & 99.6 & 77.1 & 43.4 & 65.9 & 82.9 & 76.5 & 65.9 & 88.3 & 80.7 & 65.3 & 73.4 & 64.5 & 75.4\\
Ours {\scriptsize (mRec)} & 68.3 & 95.6 & 64.1 & 63.2 & 66.6 & 83.9 & 88.4 & 55.5 & 69.7 & 68.6 & 50.6 & 69.1 & 58.0 & 69.4\\
\bottomrule
\end{tabular}
}
\end{center}
\vspace{-5px}
\caption{\textbf{Instance Segmentation on S3DIS~\cite{Armeni16CVPR} 6-fold cross validation}.
Models are trained \emph{fully supervised} with per-point semantic instance annotations.}
\label{tab:s3dis_ins1}
\vspace{-40px}
\end{table}

\begin{table}[H]
\begin{center}
\setlength{\tabcolsep}{4pt}
     \resizebox{\textwidth}{!}{
     \begin{tabular}{l|ccccccccccccc|c}
     \toprule
                & ceiling & floor & wall & beam & column & window & door & table & chair & sofa & bookshelf & board & clutter & \textbf{avg.}\\
\midrule
Ours {\scriptsize (mPrec)} & 96.8 & 99.2 & 76.4 & 46.9 & 54.1 & 68.1 & 72.9 & 59.9 & 87.6 & 76.8 & 67.5 & 70.4 & 62.7 & 72.3\\
Ours {\scriptsize (mRec)} & 68.1 & 95.3 & 64.0 & 67.5 & 63.8 & 77.0 & 90.7 & 60.0 & 70.4 & 68.9 & 53.4 & 79.9 & 57.7 & 70.5\\
     \bottomrule
    \end{tabular}
    }
\vspace{-5px}
\end{center}
    \caption{\textbf{Instance Segmentation on S3DIS~\cite{Armeni16CVPR} 6-fold cross validation}.
    Models are trained with only \emph{bounding box supervision}, no per-point annotations used to train.}
    \label{tab:s3dis_ins2}
\end{table}

\section{Non-Maximum-Clustering (NMC) Algorithm}
\begin{figure}
    \centering
    \includegraphics[width=0.5\textwidth]{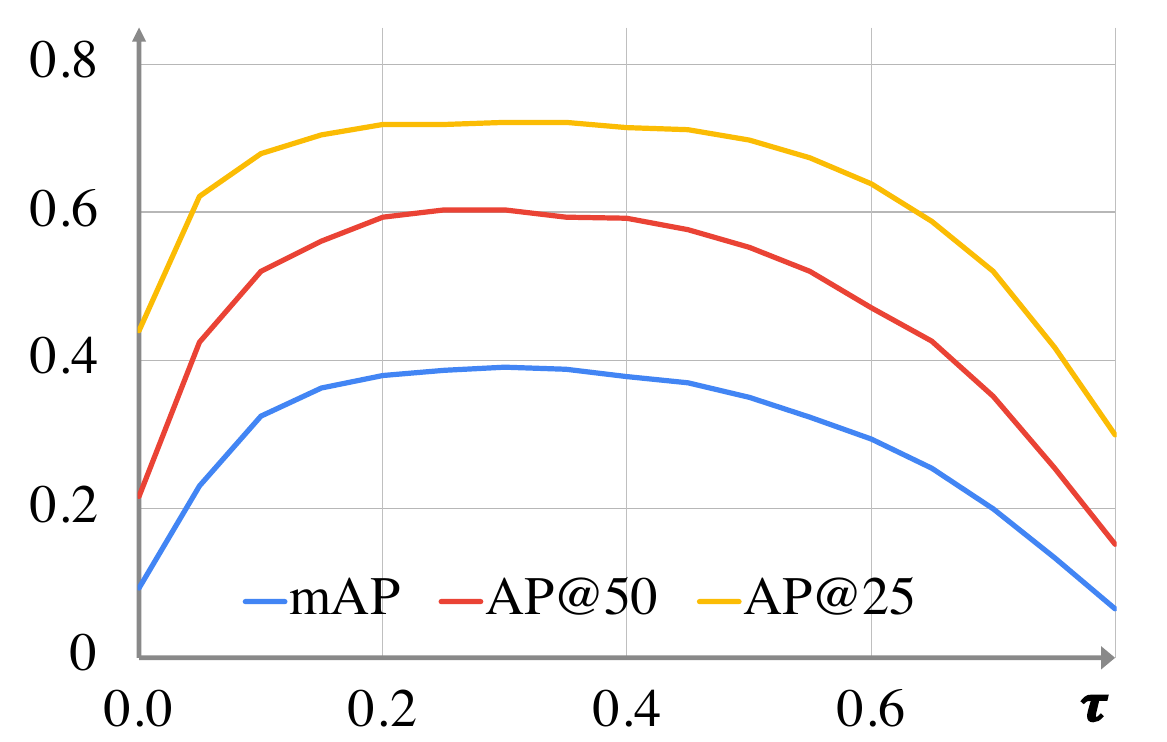}
    \caption{Effect of parameter $\tau$.}
    \label{fig:tau_eval}
\end{figure}

In \refsec{clustering} of the main paper, we introduced a clustering algorithm tailored specifically towards bounding box votes.
The pseudo-code is below.
Further, we analyse the effect of the threshold parameter $\tau$,
which can be between 0 (all boxes in single cluster) and 1 (each box is a separate cluster).
In Fig.\,\ref{fig:tau_eval}, we report mask prediction scores on ScanNet validation, and find that $\tau \approx 0.3$ performs best.

\begin{algorithm}[]
\scriptsize
\SetAlgoLined
\SetKwInOut{Input}{input}
\Input {
    $P = {(B, score)}$ \hfill// Set of bounding box votes and corresponding scores \\
}
\textbf{output:} Clustered bounding box votes.\\
$P_{candidates} \leftarrow$ P.sort (score) \hfill// Sort bounding box votes based on score  \\
Results $\leftarrow \emptyset$

\While{$P_{candidates} \neq \emptyset$} {

$P_{r} \leftarrow P.pop ()$ \hfill// Pop the highest scoring proposal\\ 
$cluster \leftarrow \{p' \;|\; IoU (p_{r}.B, p'.B) > \tau \;\&\; p' \in P$ \} \hfill // Clustering with IoUs \\
Results $\leftarrow$ Results $\cup \; \{cluster\}$  \hfill // Update the list of predictions\\ 
$P_{candidates} \leftarrow P_{candidates} \setminus cluster$ \hfill // Remove the clustered votes from the \\
\hfill // list of representative candidates
}
\Return {Results}
\label{Alg1}
\caption{Non-Maximum-Clustering (NMC)}
\end{algorithm}

\vfill

\newpage
\section{Additional Qualitative Results}
\begin{figure}[H]
\centering
\scriptsize
\resizebox{\textwidth}{!}{
        \begin{tabular}{cccc}
         {Input} & {Predicted} & {Groundtruth} & {Groundtruth} \\
         {3D Scene} & {Instance Masks} & {Instance Masks} & {Bounding Boxes} \\
         \hspace{80px} &
         \hspace{80px} &
         \hspace{80px} &
         \hspace{80px} \vspace{-10px}
    \end{tabular}
    }
{\includegraphics[width=0.24\textwidth]{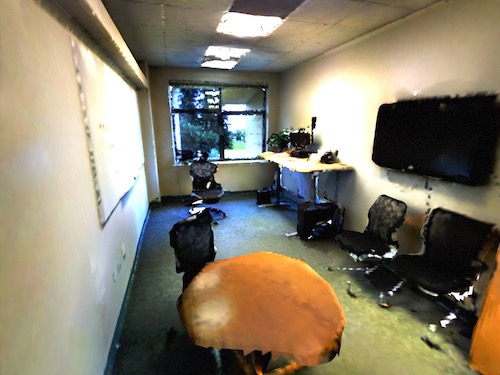}}%
{\includegraphics[width=0.24\textwidth]{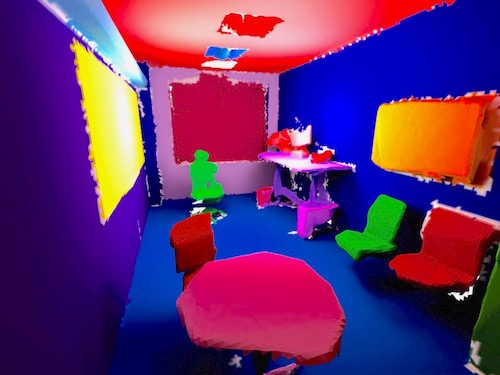}}%
{\includegraphics[width=0.24\textwidth]{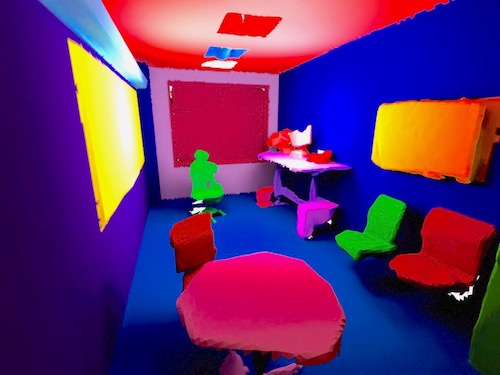}}%
{\includegraphics[width=0.24\textwidth]{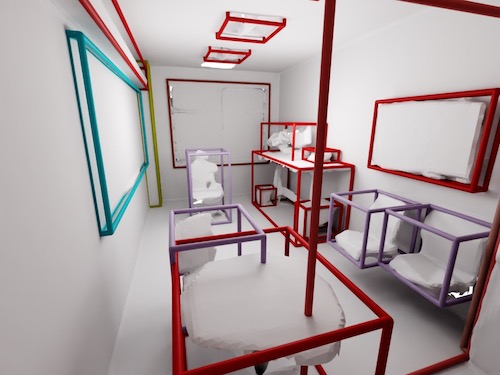}}%
\vspace{1px}
{\includegraphics[width=0.24\textwidth]{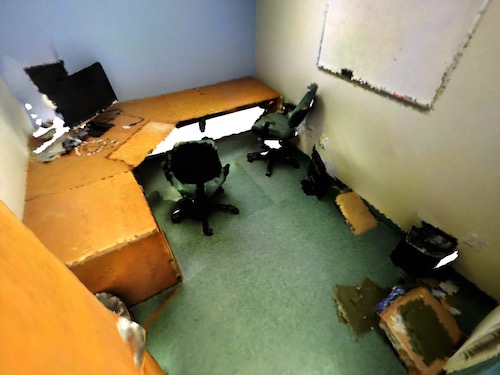}}%
{\includegraphics[width=0.24\textwidth]{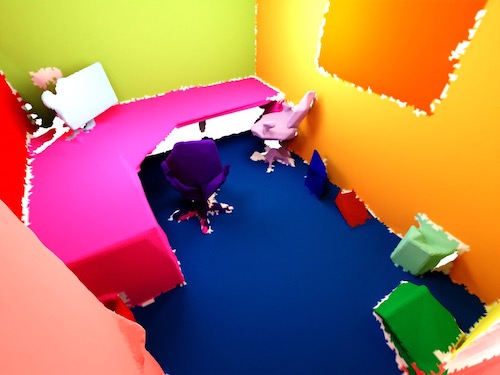}}%
{\includegraphics[width=0.24\textwidth]{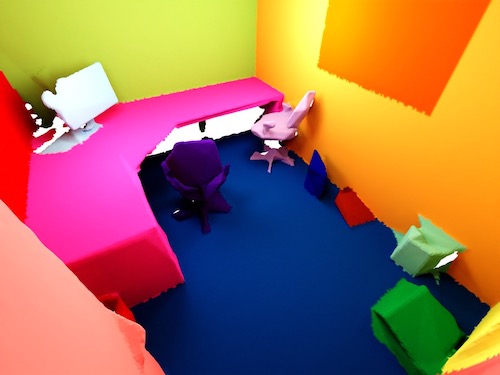}}%
{\includegraphics[width=0.24\textwidth]{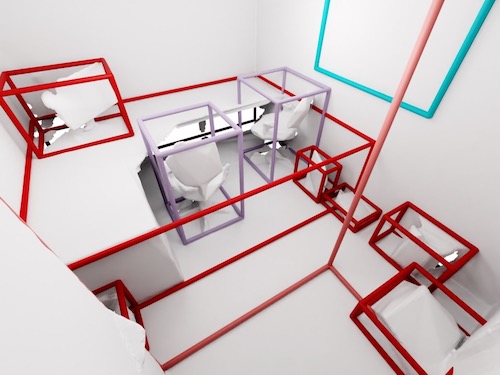}}%
\vspace{1px}
{\includegraphics[width=0.24\textwidth]{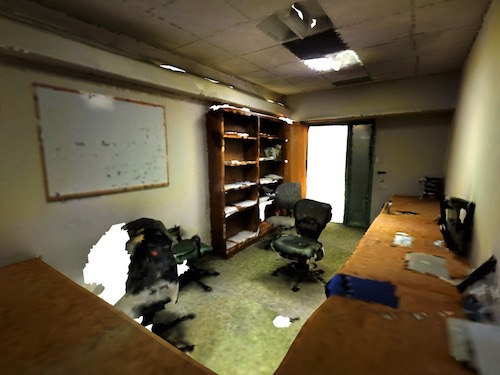}}%
{\includegraphics[width=0.24\textwidth]{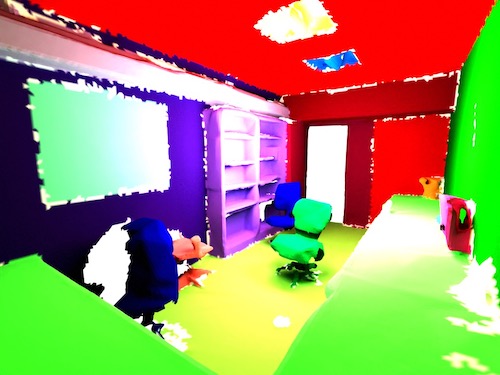}}%
{\includegraphics[width=0.24\textwidth]{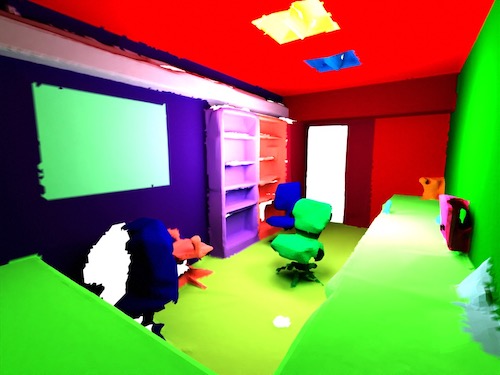}}%
{\includegraphics[width=0.24\textwidth]{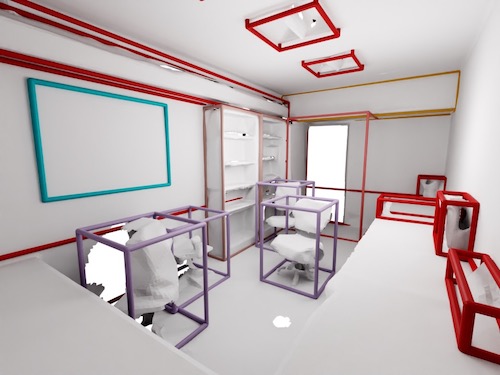}}%
\vspace{1px}
{\includegraphics[width=0.24\textwidth]{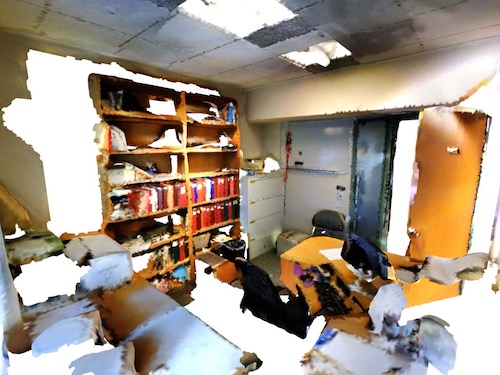}}%
{\includegraphics[width=0.24\textwidth]{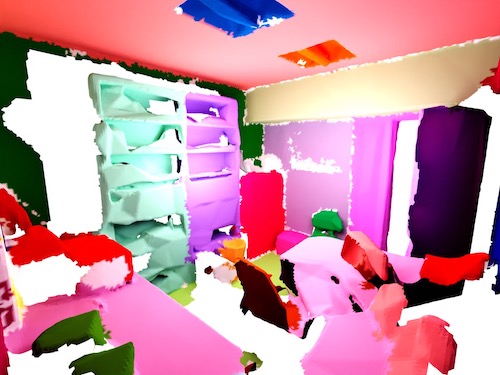}}%
{\includegraphics[width=0.24\textwidth]{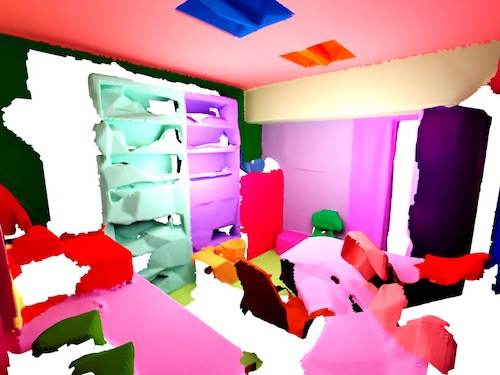}}%
{\includegraphics[width=0.24\textwidth]{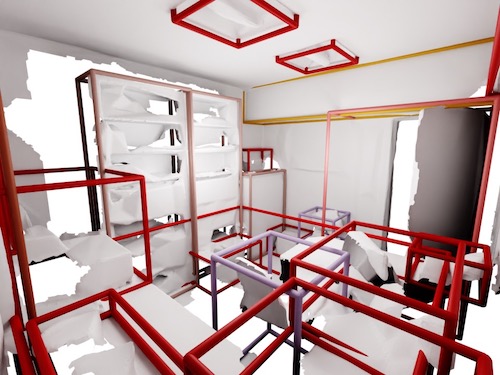}}%
\vspace{1px}
{\includegraphics[width=0.24\textwidth]{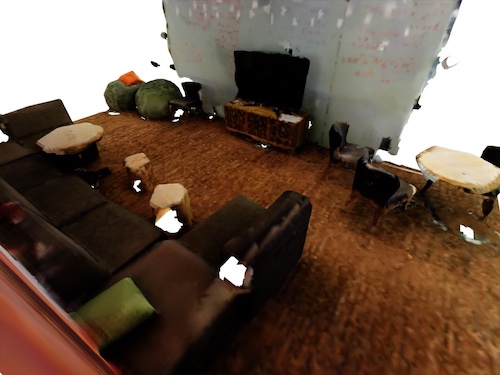}}%
{\includegraphics[width=0.24\textwidth]{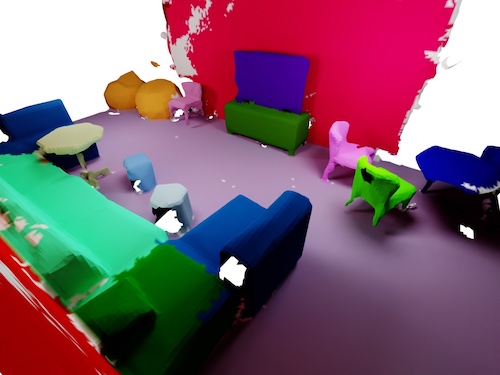}}%
{\includegraphics[width=0.24\textwidth]{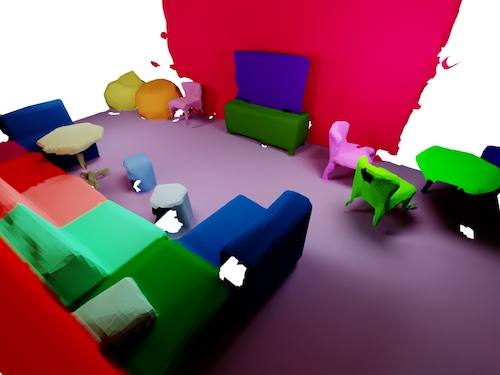}}%
{\includegraphics[width=0.24\textwidth]{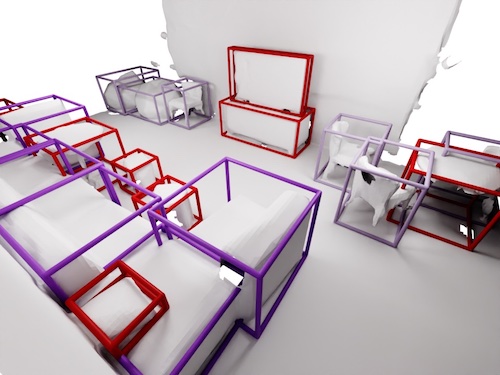}}%
\\
    \caption{
    \textbf{Qualitative Instance Segmentation Results on S3DIS\,\cite{Armeni16CVPR}}
    Individual instance masks are colored randomly and match the ground truth instance mask colors.
    During training, only bounding box annotations are used (last column), per-point instance masks (third column) are not used, and are shown here only for judging the quality of the predicted instance masks (second column).
    }
    \label{fig:qualitative_s3dis}
\end{figure}

\begin{figure}[H]
\centering
\scriptsize
{\includegraphics[width=0.44\textwidth]{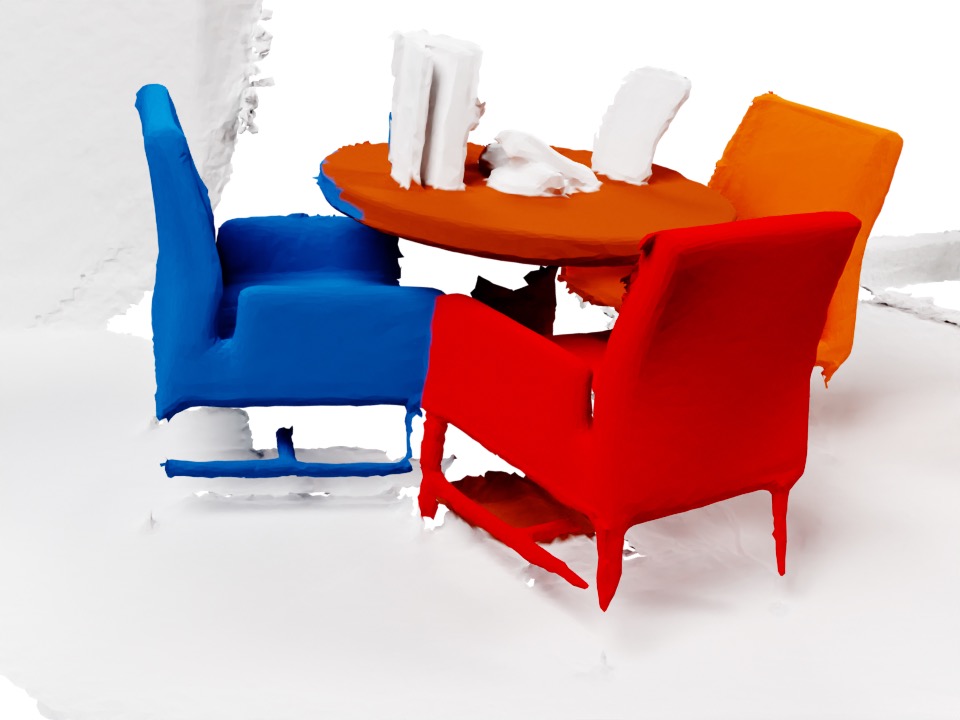}}
{\includegraphics[width=0.44\textwidth]{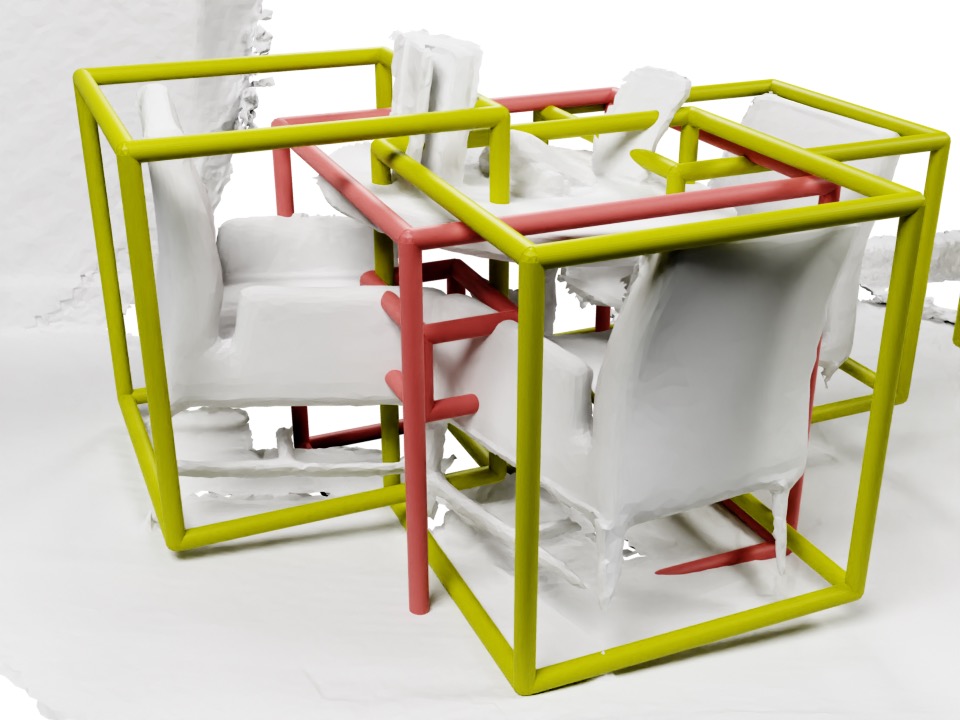}}\\
\vspace{-10px}
\resizebox{0.88\textwidth}{!}{\begin{tabular}{ll}
{Ground truth Per-Point Instance Masks} & {Ground truth Bounding Boxes}\\
\hspace{160px} & \hspace{155px} \\
& \vspace{-15px}
\end{tabular}}
{\includegraphics[width=0.44\textwidth]{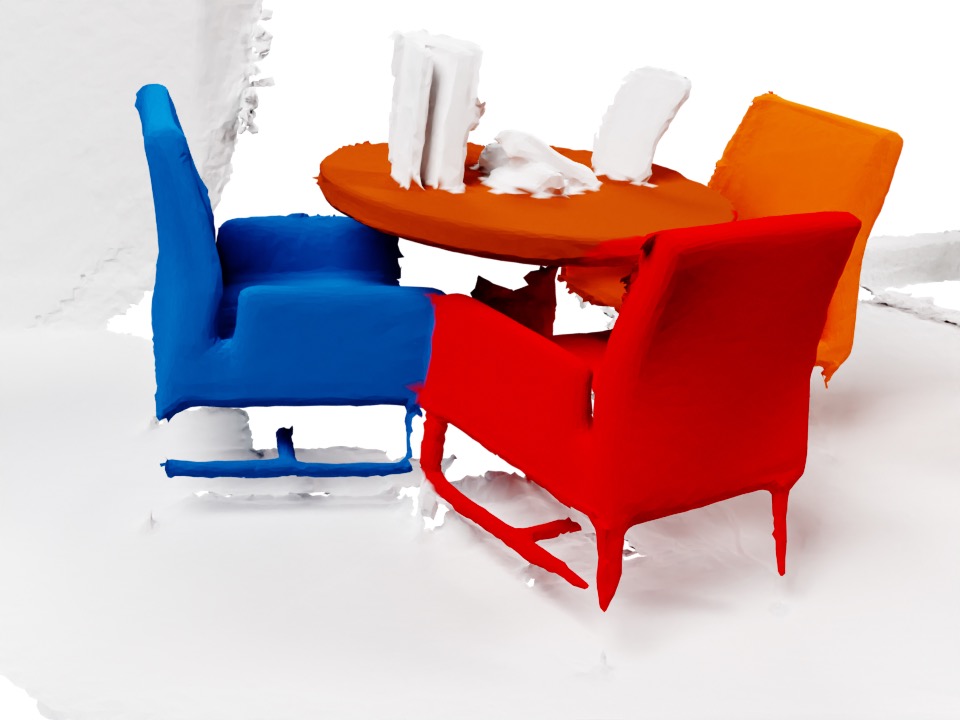}}
{\includegraphics[width=0.44\textwidth]{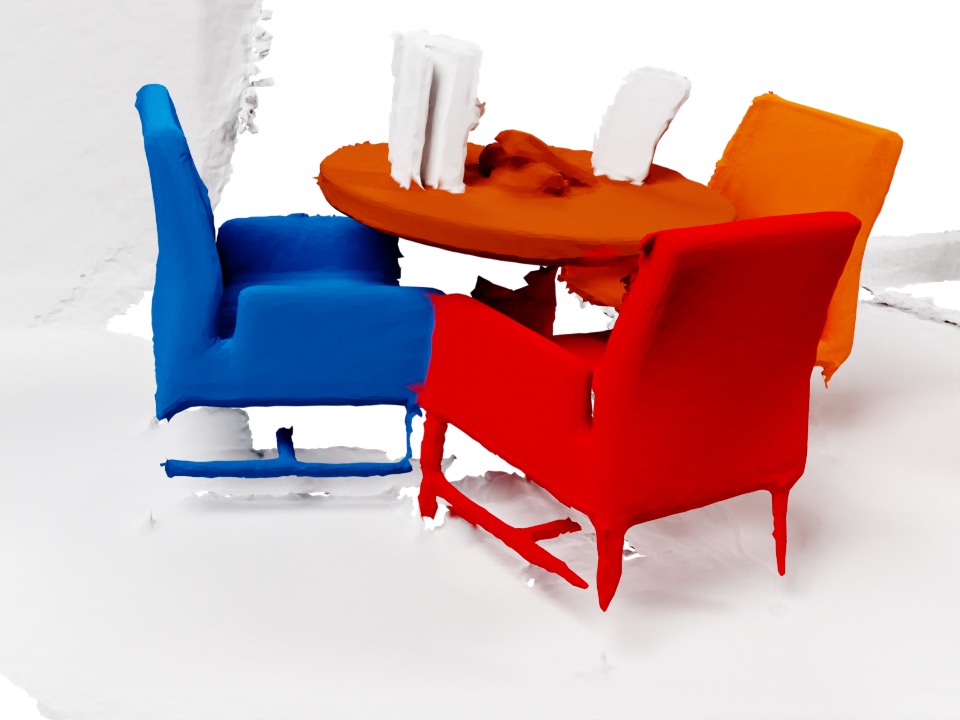}}\\
\vspace{-10px}
\resizebox{0.88\textwidth}{!}{\begin{tabular}{ll}{Predictions from Per-Point Supervision} & {Predictions from Bounding Box Supervision}\\
\hspace{160px} & \hspace{155px} \\
& \vspace{5px}
\end{tabular}}

{\includegraphics[width=0.44\textwidth]{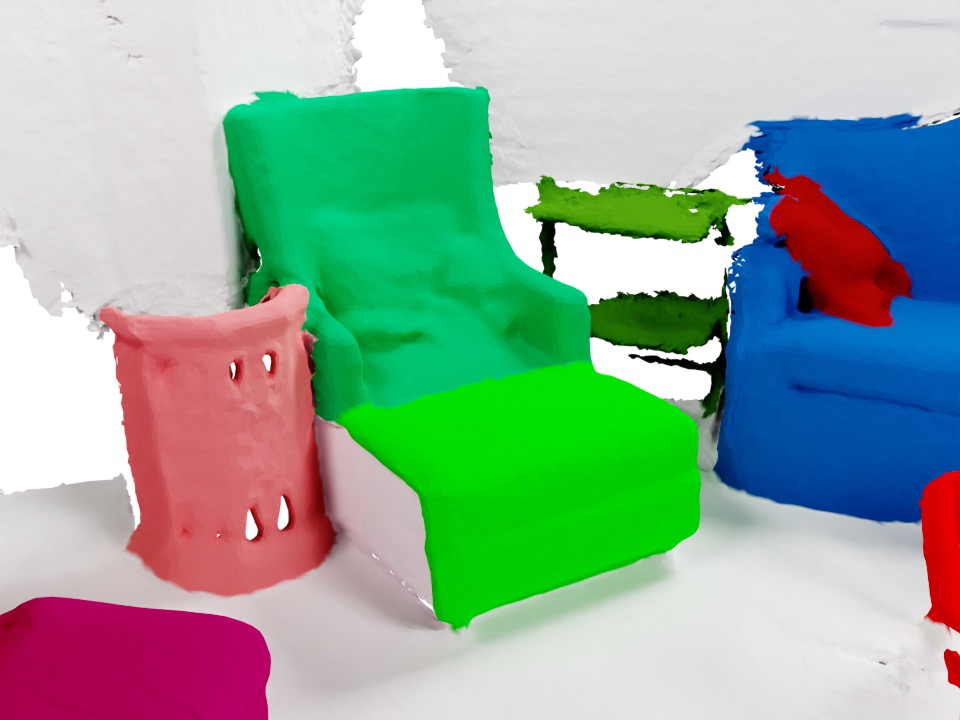}}
{\includegraphics[width=0.44\textwidth]{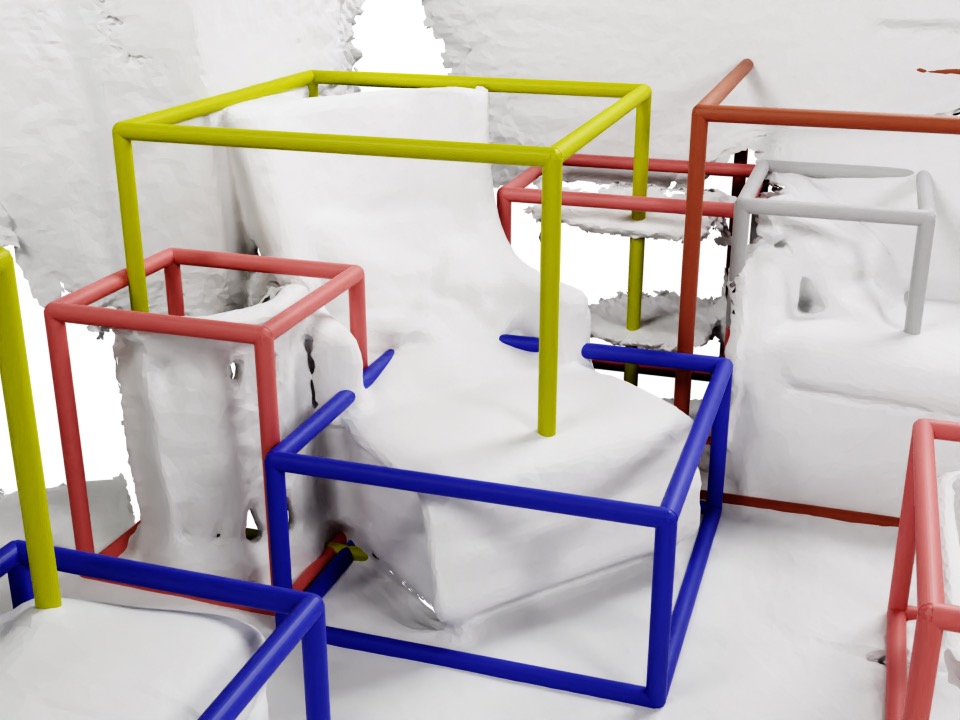}}\\
\vspace{-112px}
\resizebox{0.88\textwidth}{!}{\begin{tabular}{ll}{Ground truth Per-Point Instance Masks} & {Groundtruth Bounding Boxes}\\
\hspace{160px} & \hspace{155px} \\
& \vspace{95px}\end{tabular}}
{\includegraphics[width=0.44\textwidth]{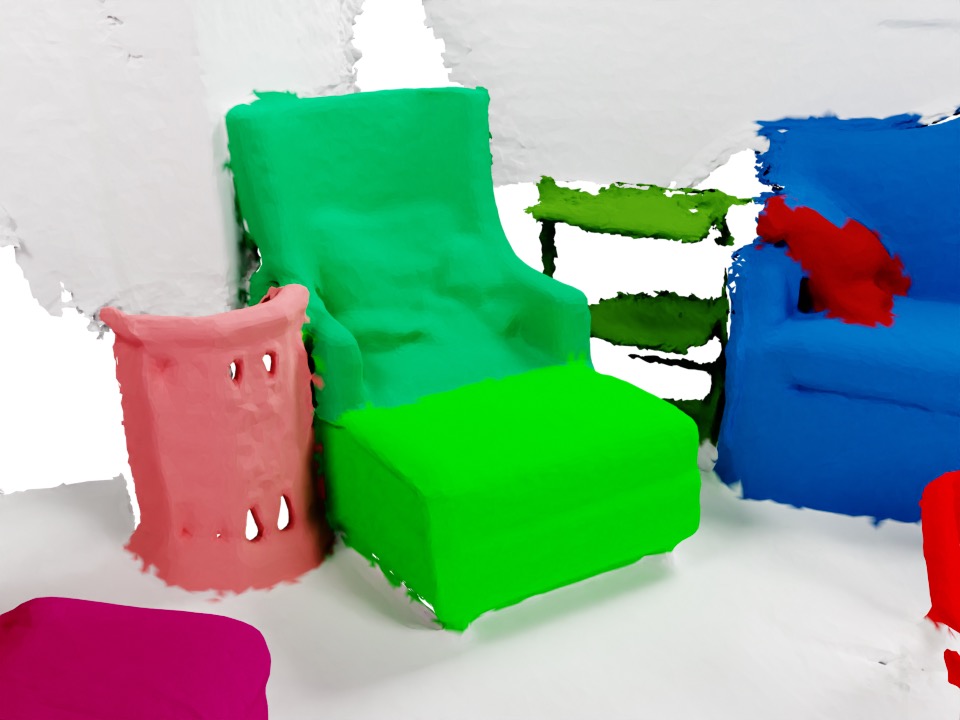}}
{\includegraphics[width=0.44\textwidth]{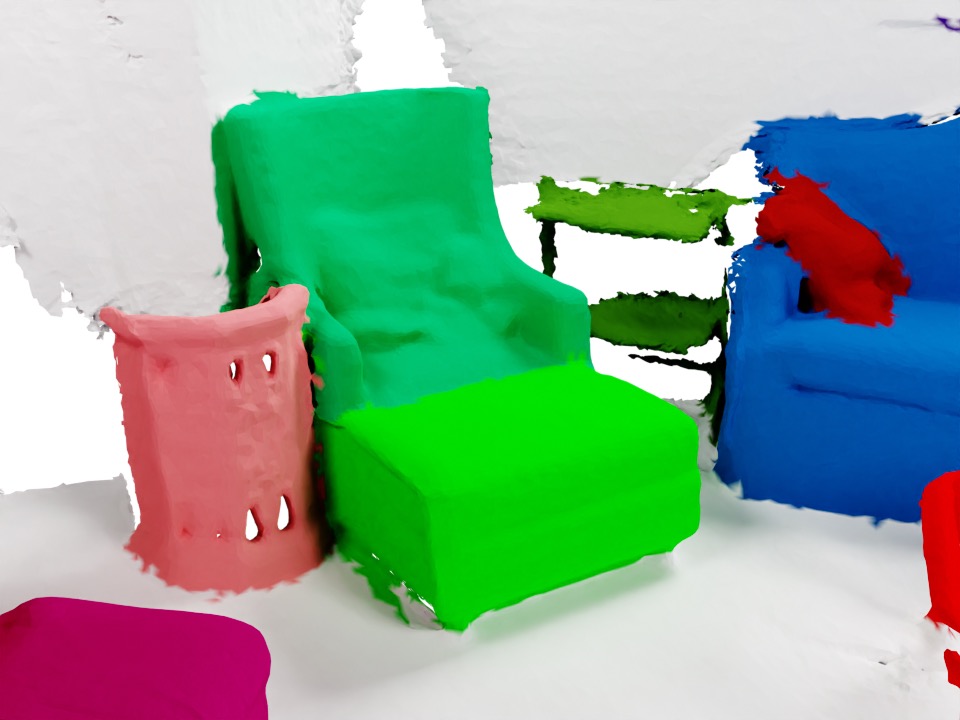}}\\
\vspace{-112px}
\resizebox{0.88\textwidth}{!}{\begin{tabular}{ll}{Predictions from Per-Point Supervision} & {Predictions from Bounding Box Supervision}\\
\hspace{160px} & \hspace{155px} \\
& \vspace{95px}
\end{tabular}}

    \caption{
    \textbf{Qualitative Instance Segmentation Results on ScanNet\,\cite{Dai17CVPR}}
    Individual instance masks are colored randomly and match the ground truth instance mask colors.
    Left: results from full per-point supervision. Right: weak bounding-box supervision.
    }
    \label{fig:qualitative_scannet_1}
\end{figure}

\begin{figure}
\centering
\scriptsize
{\includegraphics[width=0.44\textwidth]{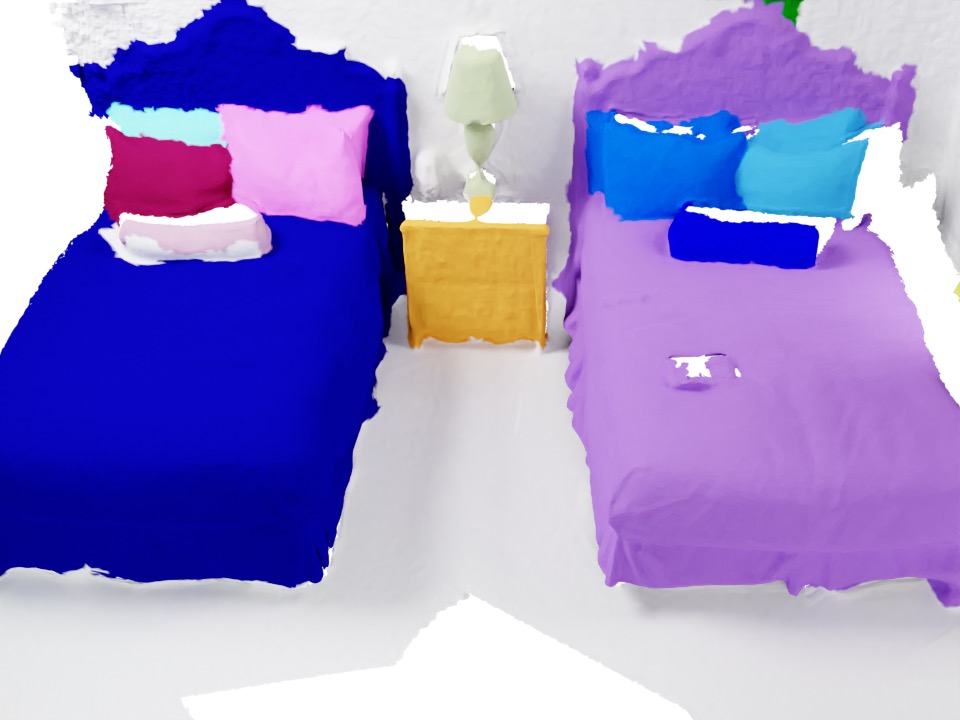}}
{\includegraphics[width=0.44\textwidth]{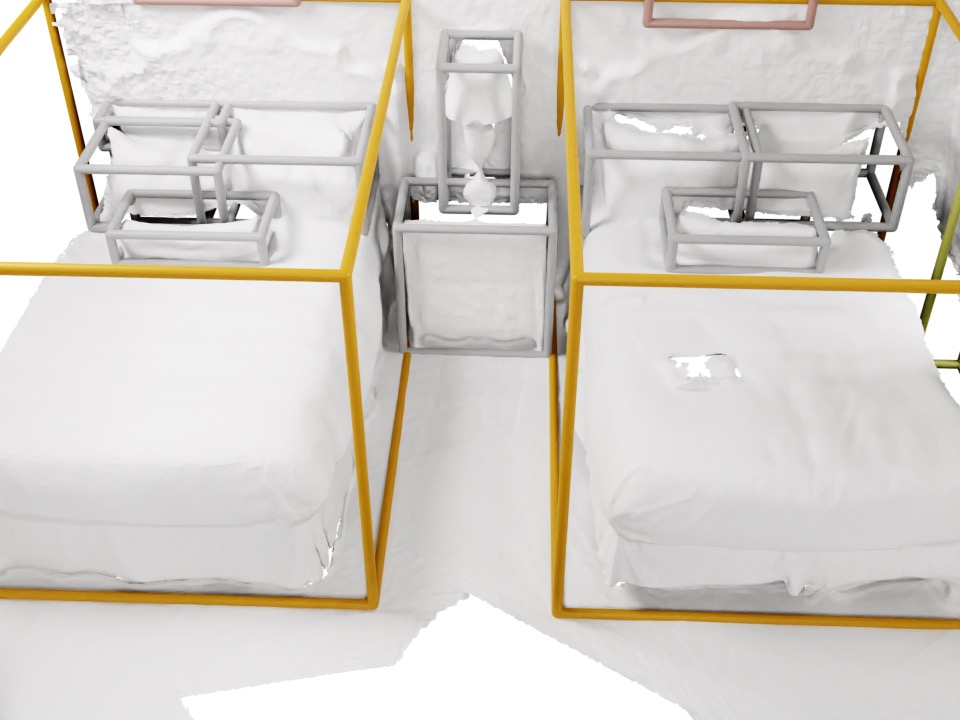}}\\
\vspace{-10px}
\resizebox{0.88\textwidth}{!}{\begin{tabular}{ll}{Ground truth Per-Point Instance Masks} & {Ground truth Bounding Boxes}\\
\hspace{160px} & \hspace{155px} \\
& \vspace{-15px}\end{tabular}}
{\includegraphics[width=0.44\textwidth]{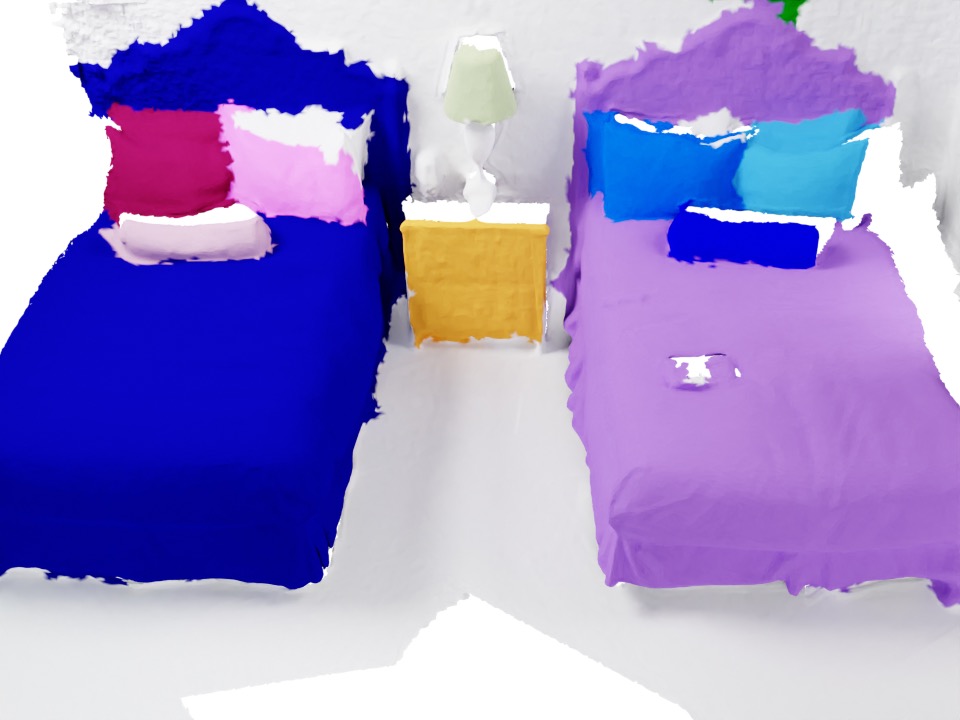}}
{\includegraphics[width=0.44\textwidth]{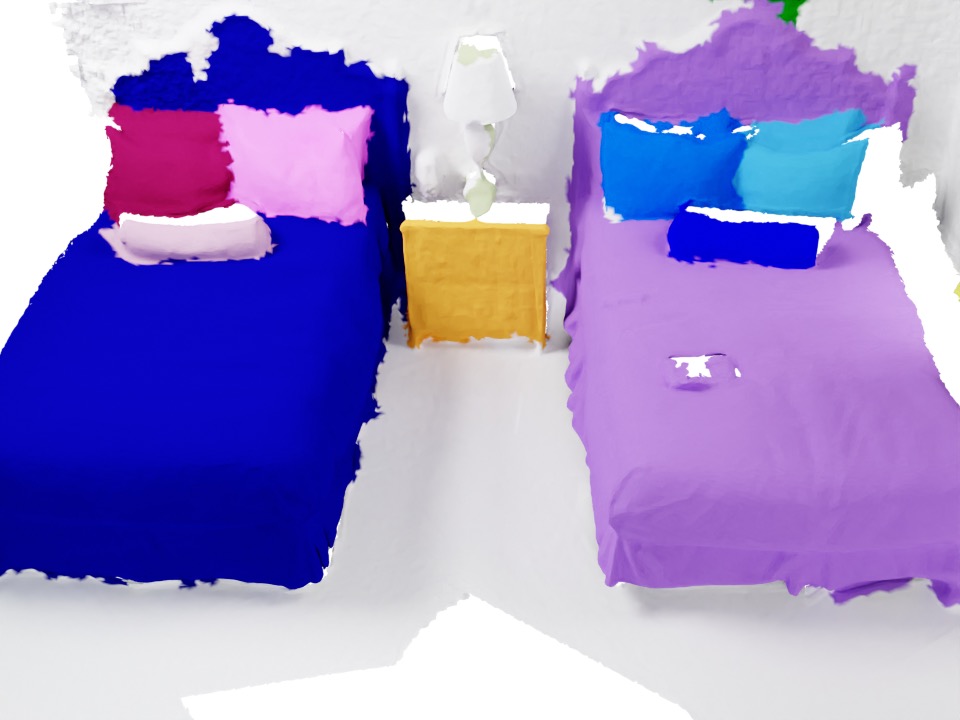}}\\
\vspace{-10px}
\resizebox{0.88\textwidth}{!}{\begin{tabular}{ll}{Predictions from Per-Point Supervision} & {Predictions from Bounding Box Supervision}\\
\hspace{160px} & \hspace{155px} \\
& \vspace{5px}
\end{tabular}}

{\includegraphics[width=0.44\textwidth]{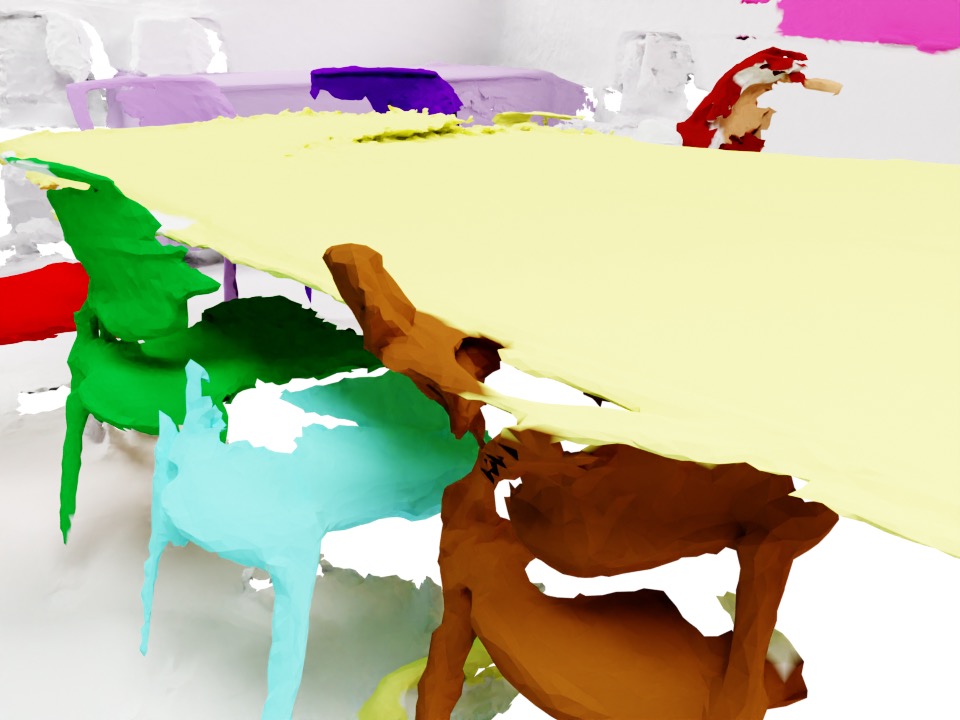}}
{\includegraphics[width=0.44\textwidth]{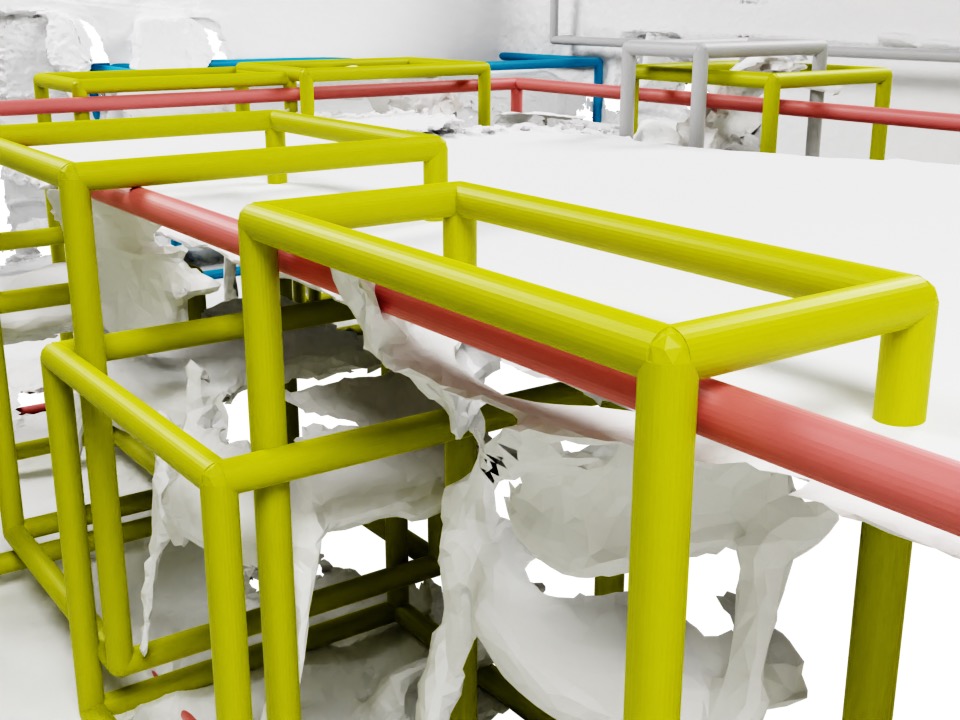}}\\
\vspace{-112px}
\resizebox{0.88\textwidth}{!}{\begin{tabular}{ll}{Ground truth Per-Point Instance Masks} & {Ground truth Bounding Boxes}\\
\hspace{160px} & \hspace{155px} \\
& \vspace{95px}\end{tabular}}
{\includegraphics[width=0.44\textwidth]{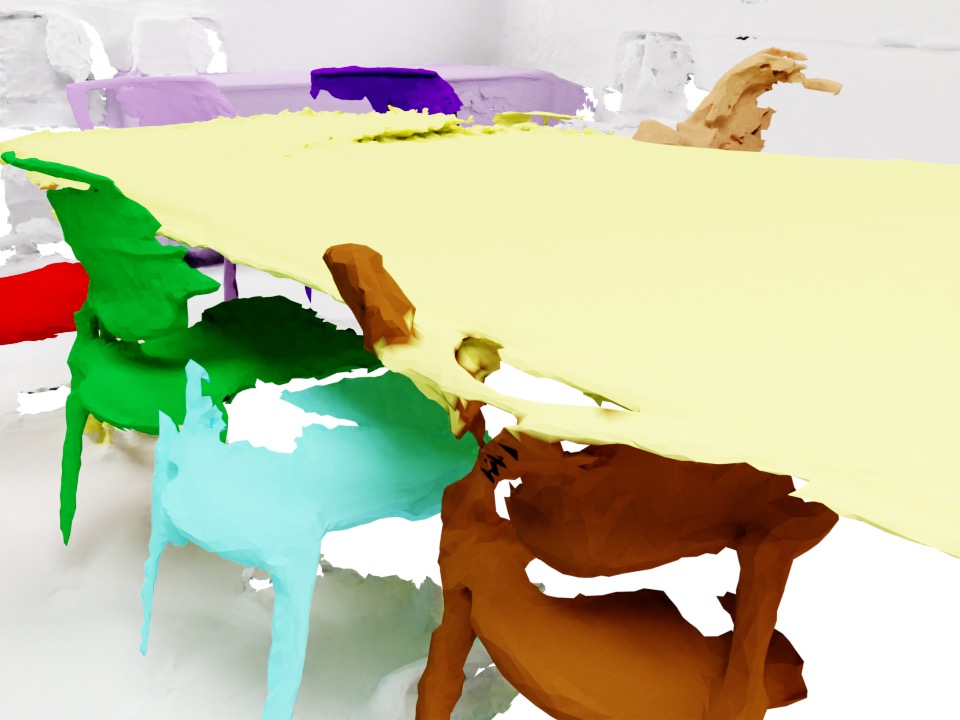}}
{\includegraphics[width=0.44\textwidth]{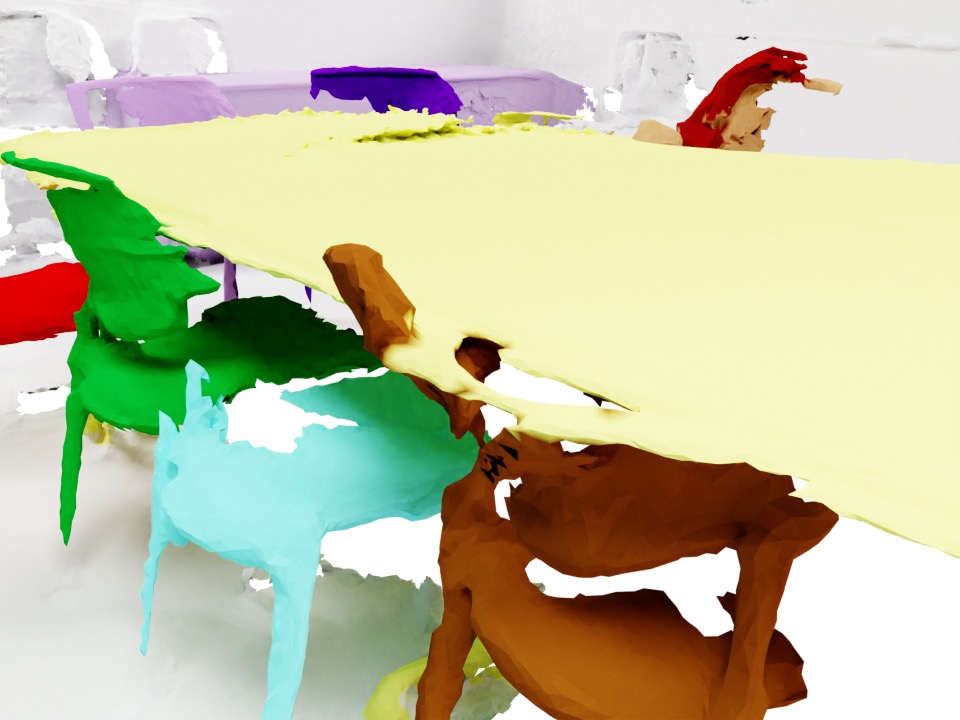}}\\
\vspace{-112px}
\resizebox{0.88\textwidth}{!}{\begin{tabular}{ll}{Predictions from Per-Point Supervision} & {Predictions from Bounding Box Supervision}\\
\hspace{160px} & \hspace{155px} \\
& \vspace{95px}
\end{tabular}}

    \caption{
    \textbf{Qualitative Instance Segmentation Results on ScanNet\,\cite{Dai17CVPR}}
    Individual instance masks are colored randomly and match the ground truth instance mask colors.
    Left: results from full per-point supervision. Right: weak bounding-box supervision.
    }
    \label{fig:qualitative_scannet_2}
\end{figure}

\vspace{-10px}
In \reffig{qualitative_s3dis}, we show exemplary qualitative results of our method on the S3DIS dataset\cite{Armeni16CVPR}.
We show the 3D input scene, our predicted instance masks learned from weak bounding box annotations and the ground truth instance masks as well as the ground truth bounding box annotations for comparison.
In \reffig{qualitative_scannet_1} and \reffig{qualitative_scannet_2},
we show additional close-up qualitative results on the ScanNet dataset\cite{Dai17CVPR}.
Besides results of our weakly-supervised model, we also show results of the same model fully-supervised with dense per-point labels. Notably, the predicted instance segmentation masks of the two models hardly differ, indicating that bounding box annotations are appropriate to train dense segmentation models.

\newpage

\clearpage
\section{Bounding Box labels \vs{} Full Point Labels}

%

%
In this section, we analyse the question: ``Is our initial point-to-box association strategy (Eq. \ref{eq:associate_func}, main paper) enough to obtain the good performance of our model?''

\begin{figure}[H]
    \centering
    \includegraphics[width=0.7\textwidth]{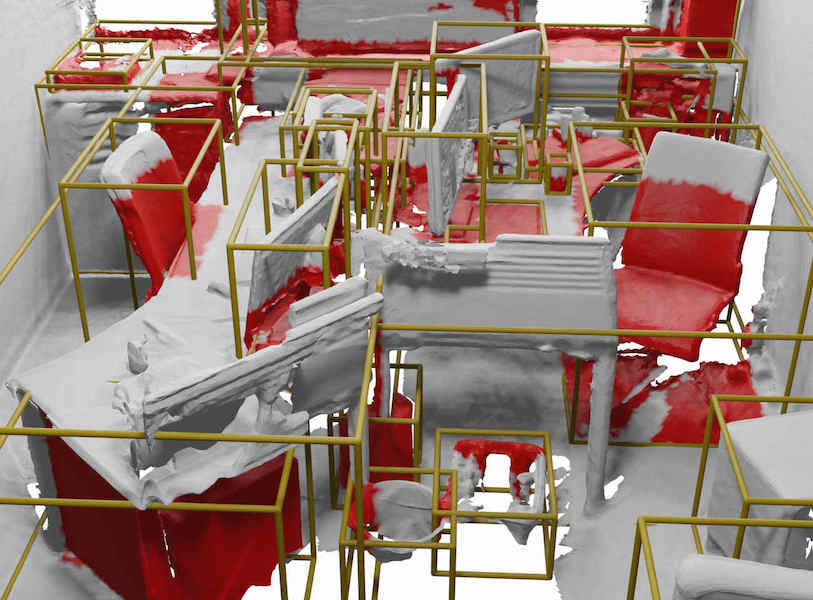}
    \caption{\textcolor{red}{\ding{108}}: Undecided Points}
    \label{fig:undecided}
\end{figure}

It is indeed correct, that this simple strategy can give good results (87\% of current fully-supervised state-of-the-art models).
It would, however, be wrong to assume that the differences between point and box labels are insignificant.
To clearly investigate this aspect, we quantitatively compare the quality of the bounding box labels to the full point labels.
Our bounding box labels achieve \textbf{$\mathbf{70.4}$\,mAP} (measured on ScanNet scenes) when evaluated against the full per-point labels (which naturally define  $100$\,mAP).
This is a performance gap of 30\%.
The reason for this difference are the ``undecided'' points that fall into multiple bounding boxes, Fig. \ref{fig:undecided}.
They are generally between two neighboring instances and make up \textbf{$\mathbf{13.5}$\%} of all points.
It is specifically these points, that are crucial for learning accurate and sharp masks of adjacent instances.

Then how is it possible that our method still achieves close to fully-supervised scores?
The reason is twofold: \textbf{1)} We observe generalization beyond the weak bounding box labels which enable precise masks on full instances (Fig.\,\ref{fig:baseline}).
During training, the model sees a large variety of scenes where the correctly supervised regions of objects outweigh the noisy ones.
This likely allows our model to build specific priors of full instance masks such that the model learns to generalize beyond the weaker box labels.\\
\textbf{2)} 
Our novel algorithm for voting and clustering based on bounding boxes can fully leverage the weak supervision.
This is shown in Tab. \ref{tab:inst_rep} (main paper) where our proposed bounding box approach largely outperforms prior center-based approaches (+8\,mAP). This is the main factor enabling almost fully-supervised performance.
\vspace{-5px}
